\documentclass[10pt,twocolumn,letterpaper]{article}

\newif\ifarxiv
\arxivtrue

\ifarxiv
\usepackage[pagenumbers]{cvpr} %
\else
\usepackage{cvpr}              %
\fi

\def\MYTITLE{Geometric-Photometric Event-based 3D Gaussian Ray Tracing}

\definecolor{cvprblue}{rgb}{0.21,0.49,0.74}
\ifarxiv
\usepackage[breaklinks,colorlinks,allcolors=cvprblue,bookmarks=true,bookmarksnumbered=true]{hyperref}
\hypersetup{
  pdftitle={\MYTITLE},
  pdfsubject={Computer Vision, Robotics},
  pdfauthor={Kai Koyama, Yoshimitsu Aoki, Guillermo Gallego, Shintaro Shiba},
  pdfkeywords={Event Cameras, Gaussian Splatting, Ray Tracing}
}
\usepackage[absolute]{textpos}
\else
\usepackage[breaklinks,colorlinks,allcolors=cvprblue]{hyperref} %
\usepackage[accsupp]{axessibility} 
\fi

\usepackage{amsmath,amssymb}
\usepackage{graphicx}
\usepackage{caption}
\usepackage{subcaption}
\usepackage{makecell}
\usepackage{booktabs}
\usepackage{siunitx}
\usepackage{adjustbox} %
\usepackage{array}
\usepackage{multirow}
\usepackage{rotating}
\usepackage{etoolbox}
\robustify\bfseries

\usepackage{url}
\usepackage{color}

\def\Lum{L}
\def\tref{t_\text{ref}} %
\def\pol{p} %

\def\cE{\mathcal{E}} %
\def\numEvents{N_e} %
\def\numPixels{N_p} %

\def\bx{\mathbf{x}}

\def\pol{p}
\def\velflow{\mathbf{v}}
\def\linvel{\mathbf{V}} %
\def\angvel{\boldsymbol{\omega}} %

\def\cN{\mathcal{N}} %
\def\bmu{\boldsymbol{\mu}} %

\def\cD{\mathcal{D}} %

\def\IWE{I} %
\def\bzero{\mathbf{0}}

\def\Real{\mathbb{R}} %

\newcommand{\bnum}[1]{\bfseries #1}

\newcommand{\novalue}{{\textendash}}

\definecolor{light-gray}{gray}{0.6}
\newcommand\gframe[1]{{\color{light-gray}\frame{#1}}}

\usepackage[capitalize]{cleveref}
\crefname{section}{Sec.}{Secs.}
\Crefname{section}{Section}{Sections}
\Crefname{table}{Table}{Tables}
\crefname{table}{Tab.}{Tabs.}

\def\Lum{L}
\def\tref{t_\text{ref}} %
\def\pol{p} %

\def\tmid{t_\text{mid}} %

\def\cE{\mathcal{E}} %
\def\numEvents{N_e} %
\def\numPixels{N_p} %

\def\bx{\mathbf{x}}

\def\pol{p}
\def\velflow{\mathbf{v}}
\def\linvel{\mathbf{V}} %
\def\angvel{\boldsymbol{\omega}} %

\def\bmu{\boldsymbol{\mu}}

\def\IWE{\text{IWE}}
\def\depth{D} %

\def\cN{\mathcal{N}} %

\def\bzero{\mathbf{0}}

\def\Real{\mathbb{R}} %

\def\numGaussians{N_g} %
\def\gs{\mathcal{G}}  %
\def\gsmean{\boldsymbol{\mu}}
\def\gsvar{\Sigma}
\def\gscolor{\mathbf{c}}
\def\gsopacity{\alpha}
\def\gsden{\mathbb{G}}  %

\def\cD{\mathcal{D}}
\def\raycolor{\mathbf{C}}   %
\def\worldX{\mathbf{X}}
\def\pixelX{\mathbf{x}}

\def\contrast{C_\text{th}}

\def\loss{\mathcal{L}} %

\usepackage[super]{nth}
\usepackage[sectionbib]{chapterbib}

\title{\MYTITLE}

\author{
Kai Kohyama $^{1}$, Yoshimitsu Aoki $^{1}$,  Guillermo Gallego $^{2}$, Shintaro Shiba $^{1,3}$ \\
$^{1}$~Keio University,
$^{2}$~Technische Universit\"at Berlin,
Science of Intelligence Excellence Cluster,\\
Einstein Center Digital Future,
and Robotics Institute Germany,
$^{3}$~The University of Tokyo.
}

\begin{document}
\ifarxiv
\definecolor{somegray}{gray}{0.5}
\newcommand{\darkgrayed}[1]{\textcolor{somegray}{#1}}
\begin{textblock}{11.5}(2.25, 0.8)  %
\begin{center}
\darkgrayed{This paper has been accepted for publication at the\\
IEEE Conference on Computer Vision and Pattern Recognition (CVPR), Denver, 2026.
\copyright IEEE}
\end{center}
\end{textblock}
\fi

\maketitle

\begin{abstract}

Event cameras offer a high temporal resolution over traditional frame-based cameras, which makes them suitable for motion and structure estimation.
However, it has been unclear how event-based 3D Gaussian Splatting (3DGS) approaches could leverage fine-grained temporal information of sparse events.
This work proposes \emph{GPERT},
a framework to address the trade-off between accuracy and temporal resolution in event-based 3DGS.
Our key idea is to decouple the rendering into two branches: event-by-event geometry (depth) rendering and snapshot-based radiance (intensity) rendering, by using ray-tracing and the image of warped events.
The extensive evaluation shows that our method achieves state-of-the-art performance on the real-world datasets and competitive performance on the synthetic dataset. 
Also, the proposed method works without prior information (e.g., pretrained image reconstruction models) or COLMAP-based initialization, 
is more flexible in the event selection number,
and achieves sharp reconstruction on scene edges with fast training time.
We hope that this work deepens our understanding of the sparse nature of events for 3D reconstruction.
\textcolor{magenta}{\url{https://github.com/e3ai/gpert}}

\end{abstract}

\section{Introduction}
\label{sec:intro}

\begin{figure}[t]
  \centering
  {{\includegraphics[clip,trim={0.3cm 0cm 12.9cm 0},width=\linewidth]{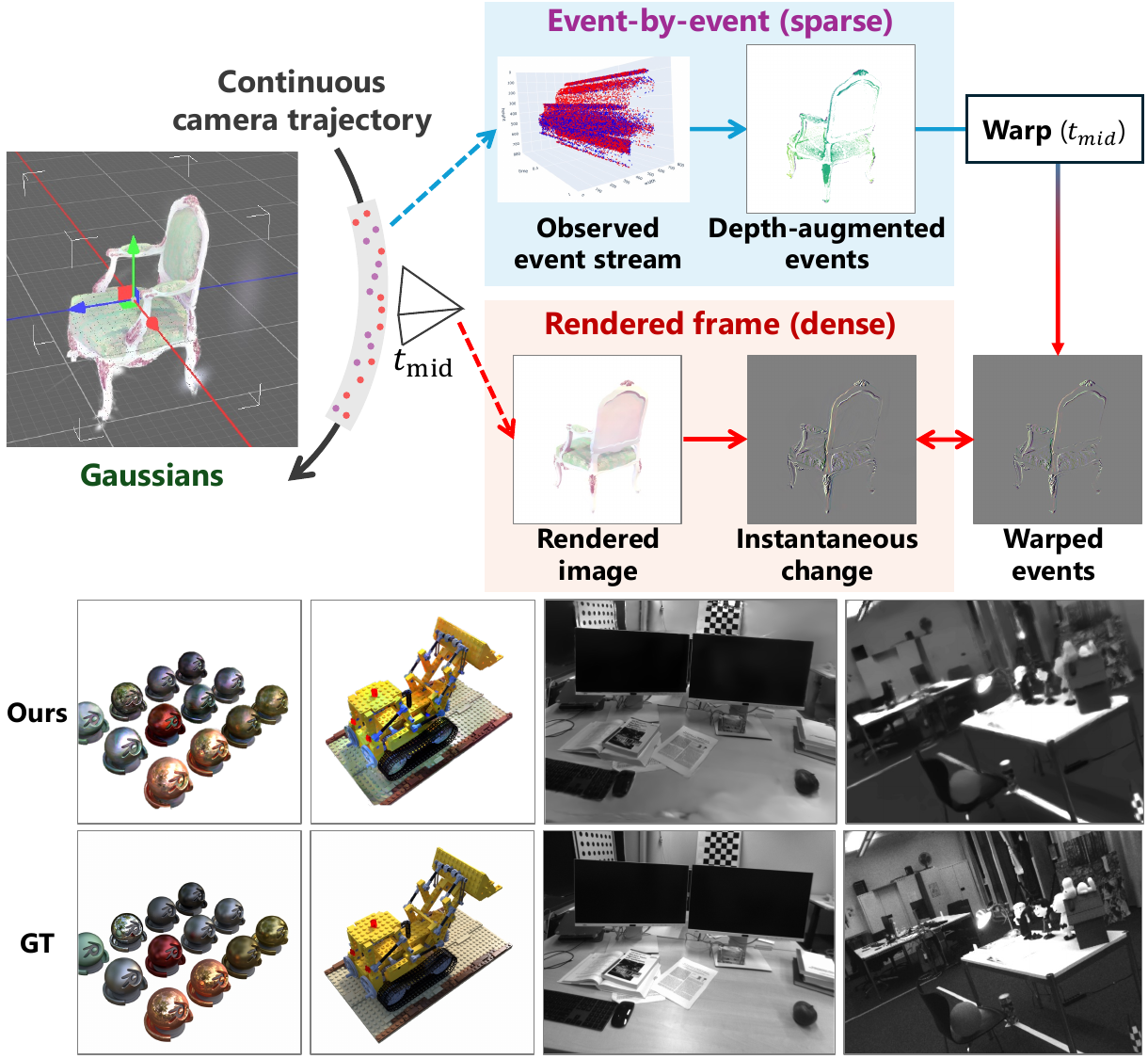}}}\\
\caption{\emph{Overview of the proposed method}, which takes raw events and poses as input.
During the optimization of the 3D Gaussians,
rendering is decoupled into two pathways: 
\emph{event-by-event} (temporally dense) depth rendering and spatially dense intensity rendering.
We use the image of warped events to connect these two pathways to compute both geometric and photometric losses. 
The color results are from a synthetic dataset, and the monochrome results are from two standard, real-world datasets.}
\label{fig:eyecatch}
\end{figure}

Event cameras have attracted increasing attention in computer vision and robotics due to their advantages and capabilities for various tasks \cite{Gallego20pami,AliAkbarpour24survey}.
Unlike conventional cameras that record synchronous frames at fixed time intervals,
event cameras respond asynchronously to per-pixel brightness changes with $\mu$s resolution \cite{Lichtsteiner08ssc,Posch14ieee,Finateu20isscc}.
This working principle makes them highly sensitive to motion, since changes are due to scene contrast and relative motion.

In parallel, Gaussian Splatting (GS) \cite{Kerbl23tog} has emerged as a state-of-the-art representation for photometric 3D reconstruction and novel view synthesis (NVS).
Since 3D structure and motion are tightly connected in the generation of event data, it is paramount to develop 3DGS algorithms that leverage the event camera's fine-grained temporal information.
Also, thanks to the event camera's minimal motion blur and high dynamic range (DR), 
event-based GS methods have the potential to overcome some of the limitations of frame-based GS, such as motion blur and low DR.

Previous approaches in event-based photometric 3D reconstruction (e.g., NeRF \cite{Mildenhall21nerf} and GS) typically perform two dense renderings per sample (e.g., \cite{Yura25cvpr}).
The difference between these two rendered images is compared with the edge-like image obtained by pixel-wise aggregation of the event data, 
resulting in a photometric loss that drives the 3D-Gaussian optimization \cite{Huang25cvpr,Yura25cvpr}.
However, as this approach requires dense (all-pixel) scene rendering twice,
it not only slows down the training, but also introduces a fundamental limitation: 
the trade-off between accuracy and temporal window selection.
A short time interval between the two renderings fails to capture subtle intensity variations that generate only a few events.
Contrarily, a large interval makes the predicted edge image blurry and discards fine-grained temporal information in the observed edge.
Similar observations have been made in \cite{Rudnev23cvpr,Xiong24corl}: a trade-off between capturing global lighting and local details.

In this work, we fundamentally address these limitations and introduce the first framework for event-based GS that renders the dense intensity (radiance) \emph{once} per sample (i.e., a batch of events),
while keeping the rendering efficient and leveraging the high temporal resolution.
Our key idea is to consider dedicated structure and appearance %
updates in the GS framework (\cref{fig:eyecatch}).
A ``structure'' pathway defines a geometric loss built on top of Contrast Maximization \cite{Gallego18cvpr} by leveraging the known camera motion and the event-by-event (i.e., sparse) rendered depth.
An ``appearance'' pathway defines a photometric loss between the instantaneous brightness change modeled by the rendered dense intensity and the change measured by the event data.

Our method showcases several advantages through extensive evaluations.
First, it does not rely on any prior knowledge (e.g., frames or pretrained models for depth and intensity) or COLMAP \cite{Schoenberger16cvpr} for initialization.
Second, it achieves state-of-the-art rendering quality performance on real-world datasets, where poses and raw events are noisy, 
with fast training time (it is faster than the latest methods that we benchmark \cite{Huang25cvpr,Low23iccv,Yura25cvpr}).
Third, it shows robustness with respect to the number of events processed per sample, without compromising accuracy.

In summary, this work presents several distinctive contributions in event-only Gaussian Splatting:
\begin{enumerate}
\item The proposed method decouples two different quantities in 3DGS rendering: 
the continuous-time spatially sparse depth, and the instantaneous dense intensity. 
It addresses the trade-off between accuracy and temporal resolution in existing event-based 3D reconstruction methods.
\item The comprehensive evaluation shows state-of-the-art results on real-world datasets and competitive results in simulation without relying on any prior knowledge, as opposed to existing event-based 3DGS methods.
\item The proposed framework connects event-by-event depth estimation and 3DGS, which is enabled by an efficient event-by-event ray-tracing implementation. 
\item The method achieves the fastest training time among other tested state-of-the-art methods (e.g., \cite{Huang25cvpr,Yura25cvpr,Low23iccv}).
\end{enumerate}
We hope that this work unblocks the potential of high temporal resolution event data in 3D reconstruction.

\section{Related Work}
\label{sec:related}
\begin{figure*}[t]
  \centering
  {\includegraphics[clip,trim={0cm 4.7cm 4.9cm 0cm},width=\linewidth]{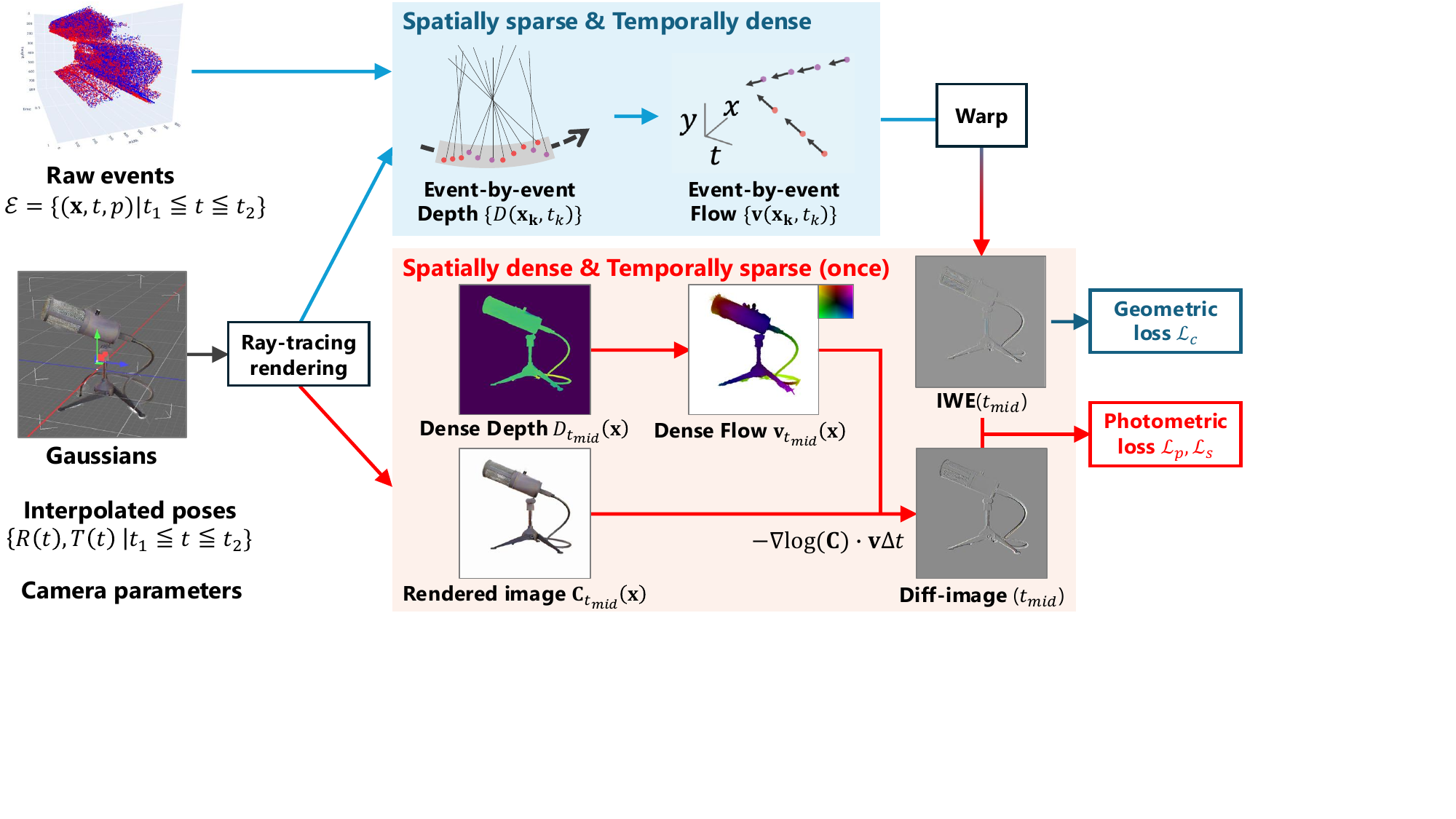}}
  \vspace{-2ex}
\caption{\emph{Method overview}. 
Using ray-tracing renderer, we estimate depth for each event and compute the flow with the interpolated poses (i.e., motion field). Performing event warping produces the image of warped events at $\tmid$ and computes the contrast loss.
We render the dense intensity (radiance) at $\tmid$ and compute the instantaneous brightness increment image, which we use for the photometric loss.
}
\label{fig:method}
\end{figure*}

3D Gaussian Splatting (3DGS) \cite{Kerbl23tog}
represents the scene as a collection of anisotropic Gaussian ellipsoids
and renders via differentiable splatting \cite{Kerbl23tog}.
These methods achieve high‐fidelity NVS with faster rendering and more scalable training in static and dynamic scenes \cite{Huang24siggraph,Wu24cvpr,Kulhanek24wildgaussians} 
than Neural Radiance Fields (NeRFs) \cite{Mildenhall21nerf}.
While most works focus on rasterization-based splatting,
recently, ray-tracing approaches have emerged, which inspire our work.
Notably, 3D Gaussian Ray Tracing (3DGRT) \cite{Moenne24tog} casts rays against volumetric Gaussian particles,
and 3DGUT \cite{Wu25cvpr} replaces Elliptical weighted-average raster splatting with an Unscented Transform projection of Gaussian particles.

Event cameras have inspired substantial work in reconstructing 3D geometry and scene appearance, leveraging motion information from their data modality of high temporal resolution.
Several approaches propose enhanced frame-based NVS aided by the complementary information in the event data, 
e.g., in the presence of fast motion (images with motion blur) or deformations \cite{Deguchi24icip,Xiong24corl,Cannici24cvpr,He25tvcg,Deng25arxiv}. 
Such approaches may still suffer from bottlenecks of the frame-based cameras in the system, such as low DR, and inaccuracies in sensor fusion calibration. 
Instead, event-only methods like our work focus on 
unlocking the potential of event cameras.

The key question of event-based NVS methods (NeRF and GS) lies in the measurement model: 
how to compare the modeled scenes, 
which typically display absolute intensity rendered at concrete viewpoints and times, 
to the acquired event data stream, which measures sparse intensity differences asynchronously at a quasi-continuous collection of times and viewpoints.
They seem opposites. 
The typical approach in the literature consists in accumulating events into an image (of intensity increments) and computing the photometric error with respect to the difference between two rendered frames at the first and last event timestamps.
While many event NeRF methods (e.g., \cite{Rudnev23cvpr,Hwang23wacv,Wang24icra,Wang25ijcnn}) and GS methods (e.g., \cite{Xiong24corl,Zhang25icra,Zahid25threedv,Liu25icra}) follow such philosophy, they face the trade-off between accuracy and temporal window selection.
Instead of rendering dense images, per-event loss computation has been proposed in event-NeRF \cite{Klenk23ral,Low23iccv,Feng25aaai} based on ray-tracing, leveraging event generation models to directly compare each observed event.
However, these NeRF methods tend to suffer from considerable event noise in real-world cameras and slow rendering.

Notably, most existing works in event-based GS utilize some prior knowledge.
For example, \mbox{Elite-EvGS} \cite{Zhang25icra} utilizes pretrained event-to-video models for initialization and regularization.
Event-3DGS \cite{Han24neurips}, which jointly estimates sensor parameters for better photometric reconstruction, also uses video reconstruction for initialization. %
\mbox{IncEventGS} \cite{Huang25cvpr}, which is conceived as an incremental tracking and mapping system (and therefore does not need camera pose information), %
uses a depth-pretrained model for bootstrapping.
\mbox{E-3DGS} \cite{Yin25applied} uses an additional piece of data to better recover absolute intensity: exposure events obtained by controlling the camera's aperture. %

Our work falls into the category of event-only 3DGS methods, such as \mbox{Event-3DGS} \cite{Han24neurips} and EventSplat \cite{Yura25cvpr}, however, with several significant differences:
($i$) earlier work utilize prior knowledge (e.g., the pretrained E2VID model \cite{Rebecq19pami}) for initial intensity recovery or initial 3D Gaussians \cite{Huang25cvpr}, while ours does not or rely on any prior knowledge, and
($ii$) we explicitly incorporate geometric and photometric loss terms in the proposed render-once framework, which improve robustness with respect to the choice of the number of events processed, as opposed to using a multi-window optimization scheme in a two-rendering pipeline \cite{Yura25cvpr,Rudnev23cvpr}.
The idea of using event warping is concurrently proposed in PAEv3D \cite{Wang24icra} (NeRF) and EF-3DGS \cite{Liao25neurips} (GS using both events and frames),
however, none of them tackle event-only GS or realize per-event depth rendering.

\section{Methodology}
\label{sec:method}

The overview of our framework is shown in \Cref{fig:method}.
The scene is modeled via 3D Gaussians (\cref{sec:method:gs}) comprising structure and appearance parameters that interact with the event data through the optimization of a weighted loss function.
The loss combines a \emph{geometric} term that measures the goodness of fit between the event data and the modeled apparent motion, 
and \emph{appearance / photometric} terms that measure the goodness of fit between the events and the brightness increment predicted by the 3D Gaussian scene model.
Accordingly, we propose to decouple the processing in two branches:
an event-by-event (i.e., spatially sparse but temporally dense) scene rendering of the unknown depth for geometric loss computation (\cref{sec:method:raytracing,sec:method:temporalrendering}) 
and a snapshot-based (i.e., spatially dense but temporally sparse) rendering for the photometric loss (\cref{sec:method:spatialrendering}).
We use the image of warped events (IWE) \cite{Gallego18cvpr} to connect both branches.

\subsection{3D Gaussian Splatting}
\label{sec:method:gs}

In the typical 3D Gaussian Splatting (3DGS) setting \cite{Kerbl23tog}, 
a static scene is represented as a set of $\numGaussians$ Gaussians
$\gs = \{ (\gsmean_i, \gsvar_i, \gscolor_i, \gsopacity_i) \}_{i=1}^{\numGaussians}$,
where $\gsmean_i \in \mathbb{R}^3$ denotes the 3D mean position,
$\gsvar_i \in \mathbb{R}^{3 \times 3}$ the covariance matrix encoding the anisotropic spatial extent,
$\gscolor_i \in \mathbb{R}^3$ the color
and $\gsopacity_i \in [0,1]$ the opacity.
Each Gaussian defines a density function in space:
$\gsden_i(\worldX; \gsmean_i, \gsvar_i) \doteq e^{-\frac{1}{2} (\worldX - \gsmean_i)^\top \gsvar_i^{-1} (\worldX - \gsmean_i)}$.
The rendered appearance is obtained by projecting these 3D Gaussians into the image plane and blending their contributions according to visibility and opacity.
Projected Gaussians are at pixel locations $\gsmean'_i = \pi(\gsmean_i)$ and with covariances $\gsvar'_i \approx J_i \gsvar_i J_i^\top$,
where $J_i = \frac{\partial \pi(\worldX)}{\partial \worldX} \big|_{\worldX = \gsmean_i}$ is the Jacobian of the projection function $\pi:\Real^3 \to \Real^2$ that maps world coordinates to pixel coordinates.
The contribution of each Gaussian to a pixel $\pixelX$ is then given by:
$w_i(\pixelX) = \alpha_i 
\gsden_i(\pixelX; \gsmean'_i, \gsvar'_i)$.

The rendered color for each pixel $\mathbf{C}(\pixelX)$ is approximated by alpha compositing along the camera ray with correct ordering and blending based on depth:
\begin{equation}
\label{eq:gs:color}
\textstyle
\raycolor(\pixelX) = \sum_{i=1}^{N} \gscolor_i \, w_i(\pixelX) \, \prod_{j=1}^{i-1} (1 - w_j(\pixelX)).
\end{equation}

Finally, to obtain a differentiable depth rendering,
we associate each Gaussian with a mean depth value $Z_i = \mathbf{e}_3^\top \gsmean_i$ in camera coordinates.  
The rendered depth $D(\pixelX)$ is then given by the opacity-weighted expectation:
\begin{equation}
\label{eq:gs:depth}
D(\pixelX) = \frac{\sum_{i=1}^{N} Z_i \, w_i(\pixelX) \, \prod_{j<i} (1 - w_j(\pixelX))}{\sum_{i=1}^{N} w_i(\pixelX) \, \prod_{j<i} (1 - w_j(\pixelX)) + \epsilon}.
\end{equation}

\subsection{Event-by-event Ray Tracing}
\label{sec:method:raytracing}

An event camera asynchronously captures visual changes as soon as the log-intensity $L$ at a pixel $\pixelX$ exceeds a threshold $C_\text{th}$:
$\Delta \Lum(\pixelX_k,t_k) \doteq \Lum(\pixelX_k,t_k) - \Lum(\pixelX_k, \protect{t_k-\Delta t_k}) = \pol_k \, C_\text{th}$.
Each event $e_k \doteq (\pixelX_k, t_k, \pol_k)$ specifies the space-time coordinates $(\pixelX_k, t_k)$ and polarity $\pol_k \in \{+1, -1\}$ %
of the change.

Events are sparse in pixel space and quasi-continuous (dense) in time.
To fully leverage sparsity, the rendering of the 3DGS should be sparse rather than image rasterization. %
Hence, we propose the framework of \emph{event-by-event rendering} in the 3DGS pipeline,
inspired by recent advances in ray-tracing GS \cite{Moenne24tog,Wu25cvpr}.
The idea that each event should also carry information about depth, i.e., \emph{depth-augmented events}, originates in \cite{Weikersdorfer14icra} for the context of SLAM.

For each event $e_k$, we render the corresponding depth $\depth(\bx_k, t_k)$, which is now a function of both space and~time.
To this end, at each timestamp $t_k$, we compute the interpolated camera pose $(R(t_k), T(t_k))$ and the ray through the camera's optical center and pixel $\pixelX_k$.
Finally, GPU-accelerated ray tracing enables us to efficiently render event-by-event depth $\cD \doteq \{\depth(\bx_k, t_k)\}_{k=1}^{\numEvents}$, as illustrated in \cref{fig:depthpoint}, column (b).

Assuming a stationary scene viewed by a moving camera with linear and angular velocities $\linvel$ and $\angvel$, respectively, 
the per-event depth $\depth$ can be used to compute the per-event apparent motion %
via the motion field equation (\cref{sec:suppl:motionfield}) \cite{Trucco98book}:
\begin{equation}
\label{eq:motionField}
\velflow(\bx,t) = \frac1{\depth(\bx,t)}A(\bx)\linvel + B(\bx)\angvel.
\end{equation}
See the example in \cref{fig:depthpoint}, column (d).

\subsection{Geometric Loss}
\label{sec:method:temporalrendering}

To guide the estimation of the 3DGS parameters, we consider a geometric loss that is computed in an unsupervised manner 
following the Contrast Maximization (CMax) framework \cite{Gallego18cvpr} that is widely used for various motion estimation tasks
\cite{Gallego17ral,Kim21ral,Shiba22aisy,Peng21pami,Gu21iccv,Guo24tro,Guo24epba,Guo25iccv,Yamaki25cvprw,Shiba25iccv}.
Under the brightness constancy assumption, events $\cE \doteq \{e_k\}_{k=1}^{\numEvents}$ are caused by moving edges 
and can be motion-compensated by a warping operation if their motion is known: 
$\cE'_{\tref} \doteq \{e'_k\}_{k=1}^{\numEvents}$, where $e'_k \doteq (\bx'_k,\tref,\pol_k)$, at a reference time $\tref$.
We formulate the warp using the spatio-temporal optical flow $\velflow(\bx, t)$ \cite{Shiba22eccv},
which in the 3DGS setting can be obtained using \eqref{eq:motionField},
\begin{equation}
\label{eq:warp:oflow}
\bx'_k = \bx_k + (t_k-\tref) \, \velflow(\bx_k,t_k).
\end{equation}
Then, the warped events are aggregated to produce an image or histogram of warped events (IWE, top branch of \cref{fig:method})
\begin{equation}
\label{eq:IWE}
\textstyle
\IWE(\bx; \tref, \depth) \doteq \sum_{k=1}^{\numEvents} b_k \contrast \delta (\bx - \bx'_k),
\end{equation}
where $b_k=\pol_k$ if polarity is used and $b_k=1$ if polarity is not used. 
The Dirac delta is approximated by a Gaussian, 
$\delta(\bx-\bmu)\approx\cN(\bx;\bmu,\sigma^2=1 \text{px})$.

The IWE measures the alignment between the event data and the candidate motion $\velflow$. 
The true motion $\velflow^\ast$ leads to a sharp IWE, with motion-compensated edges.  
Hence, as geometric loss we use the IWE sharpness (without polarity, $b_k=1$), normalized by the value at zero flow \cite{Shiba22eccv}:
\begin{equation}
\begin{split}
\label{eq:loss:contrast}
    \loss_\text{c} & \doteq G(\bzero; -) \,/\, G(\velflow(\depth); \tref),\\
    G(\velflow(\depth); \tref) & = \textstyle \frac{1}{|\Omega|} \int_{\Omega} \| \nabla \IWE(\bx; \tref, \depth)\|_1 d\bx.
\end{split}
\end{equation}
Notice that we use the reciprocal of the contrast objective due to the minimization formulation, 
and the $L^1$-norm because it performs well for depth estimation \cite{Shiba24pami}.
\def\figWidth{0.24\linewidth}
\begin{figure}[t]
	\centering
    {\footnotesize
    \setlength{\tabcolsep}{1pt}
	\begin{tabular}{
	>{\centering\arraybackslash}m{\figWidth} 
	>{\centering\arraybackslash}m{\figWidth} 
	>{\centering\arraybackslash}m{\figWidth} 
	>{\centering\arraybackslash}m{\figWidth}}
 
        {\gframe{\includegraphics[width=\linewidth]{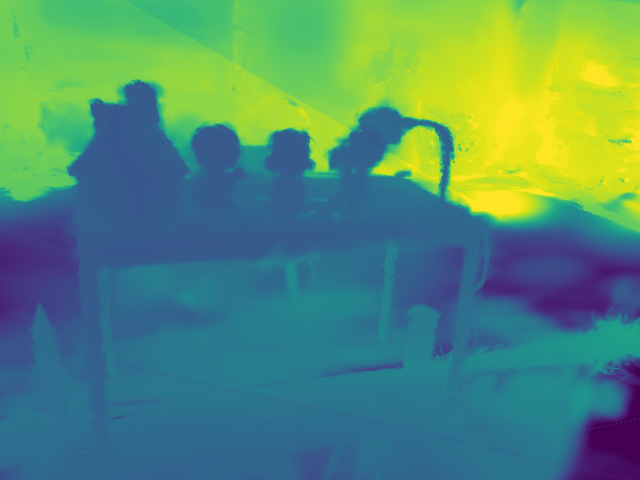}}}
        & {\gframe{\includegraphics[width=\linewidth]{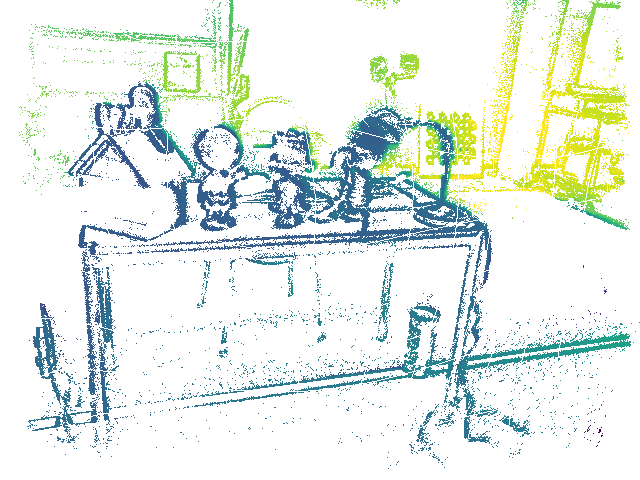}}}
        & {\gframe{\includegraphics[width=\linewidth]{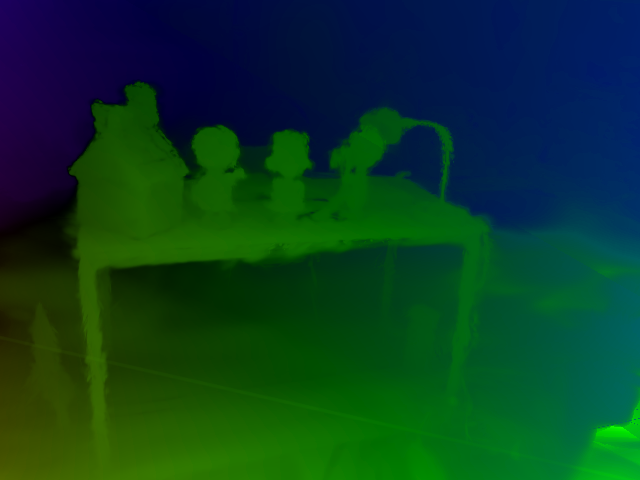}}}
        & {\gframe{\includegraphics[width=\linewidth]{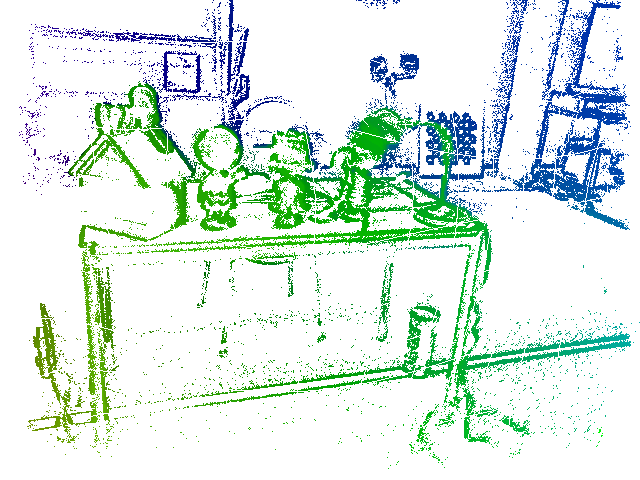}}}\\
   
        {\gframe{\includegraphics[trim={0 0 0 2.5cm},clip,width=\linewidth]{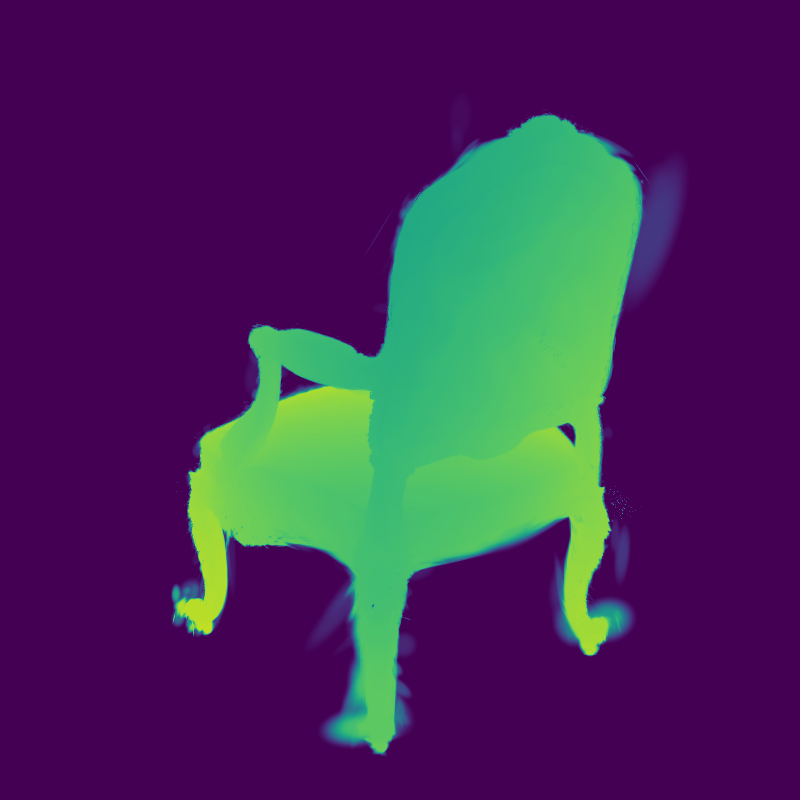}}}
        & {\gframe{\includegraphics[trim={0 0 0 2.5cm},clip,width=\linewidth]{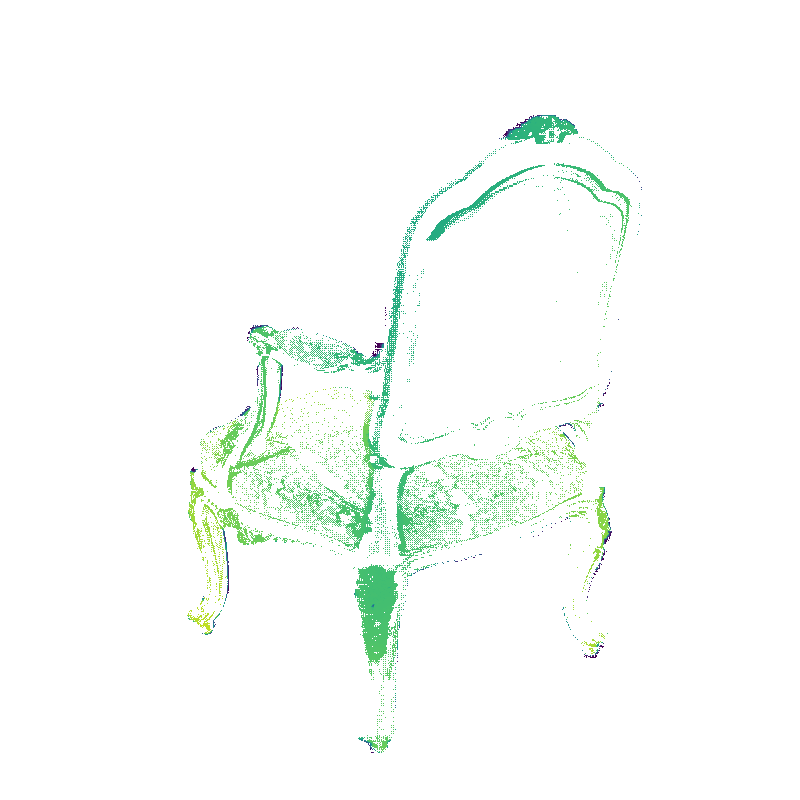}}}
        & {\gframe{\includegraphics[trim={0 0 0 2.5cm},clip,width=\linewidth]{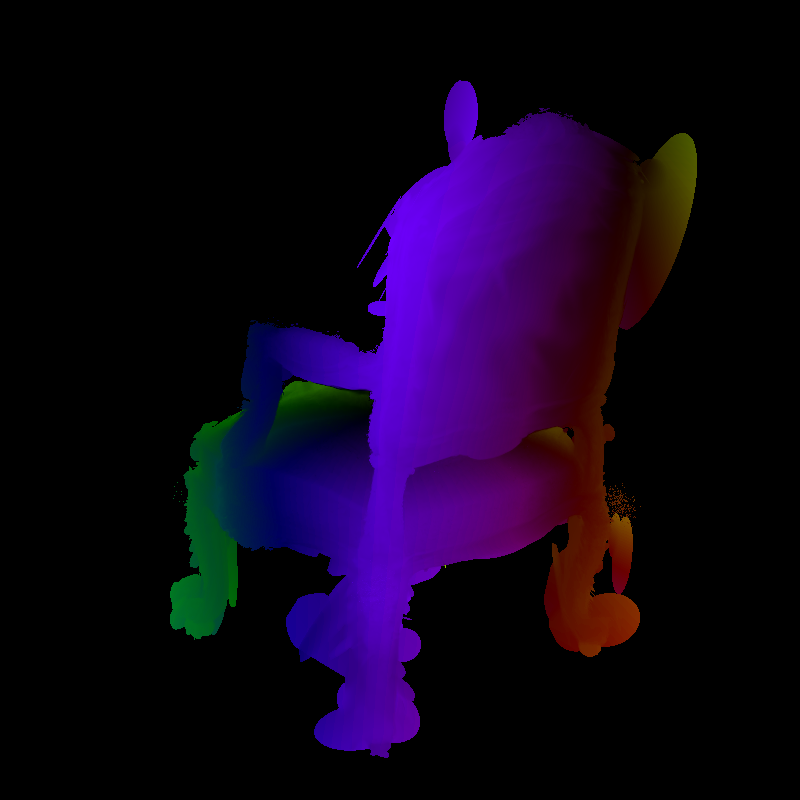}}}
        & {\gframe{\includegraphics[trim={0 0 0 2.5cm},clip,width=\linewidth]{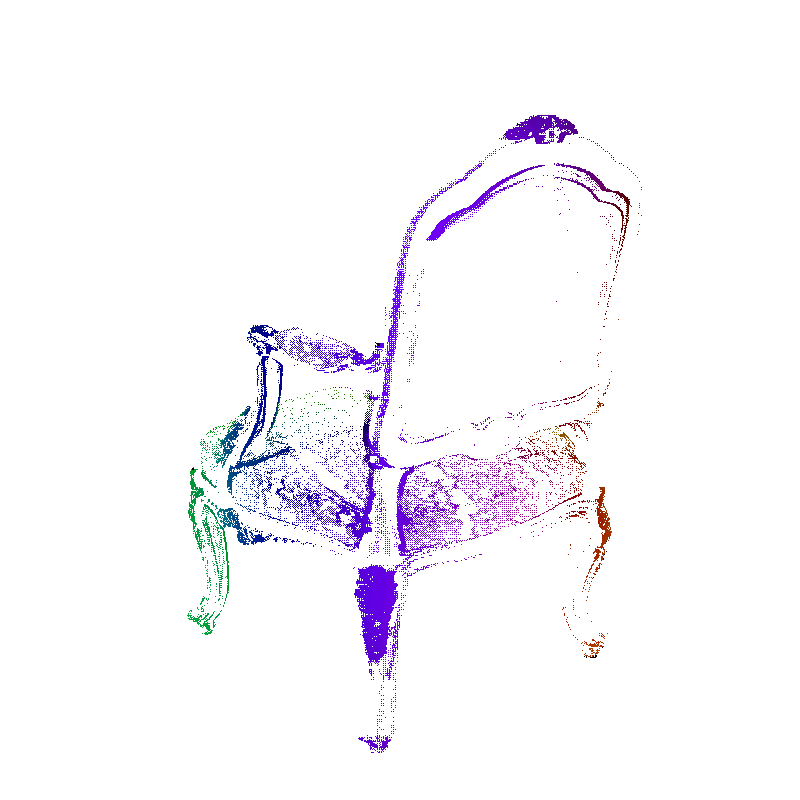}}}\\

        (a) Dense depth
		& (b) Sparse depth
		& (c) Dense flow
		& (d) Sparse flow
		\\
	\end{tabular}
	}
    \vspace{-1ex}
    \caption{\emph{Visualization of dense/sparse depth and optical flow}. 
    Sparse depth and optical flow are not simply obtained by masking the dense counterparts, 
    but by actual event-by-event ray tracing (\cref{sec:method:raytracing}).
    Top: using real events (EDS).
    Bottom: using synthetic events.
    The flow color notation is specified in \cref{fig:method}.}
    \label{fig:depthpoint}
    \vspace{-1ex}
\end{figure}

\def\textWidth{0.02\linewidth}
\def\figWidth{0.159\linewidth} %
\begin{figure*}[t]
	\centering
    {%
    \footnotesize
    \setlength{\tabcolsep}{1pt}
	\begin{tabular}{
	>{\centering\arraybackslash}m{\textWidth}
	>{\centering\arraybackslash}m{\figWidth} 
	>{\centering\arraybackslash}m{\figWidth} 
	>{\centering\arraybackslash}m{\figWidth} 
	>{\centering\arraybackslash}m{\figWidth} 
	>{\centering\arraybackslash}m{\figWidth}
	>{\centering\arraybackslash}m{\figWidth}}

            \rotatebox{90}{\makecell{EDS\_07}}
		& \gframe{\includegraphics[width=\linewidth]{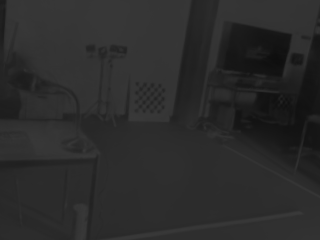}}
		& \gframe{\includegraphics[width=\linewidth]{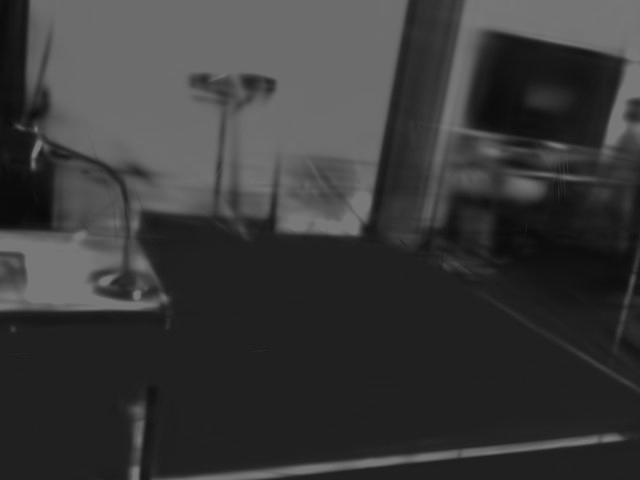}}
		& \gframe{\includegraphics[width=\linewidth]{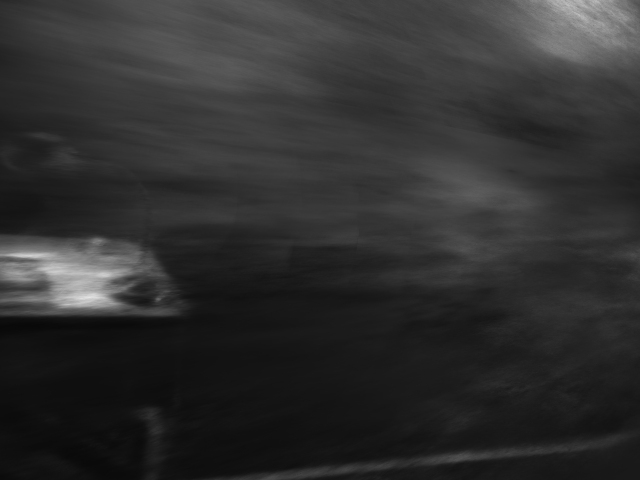}}
		& \gframe{\includegraphics[width=\linewidth]{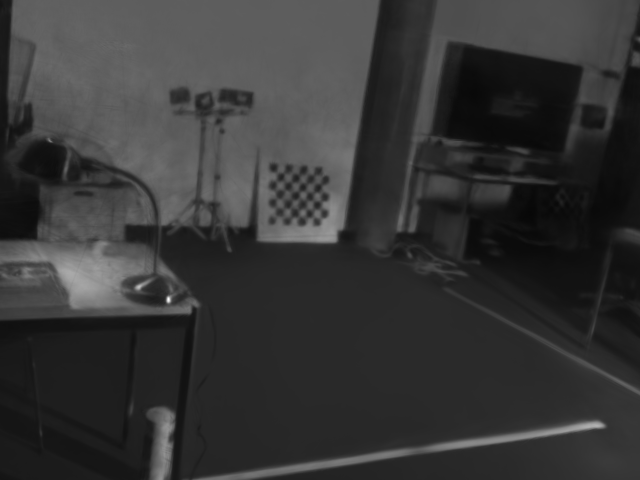}}
		& \gframe{\includegraphics[width=\linewidth]{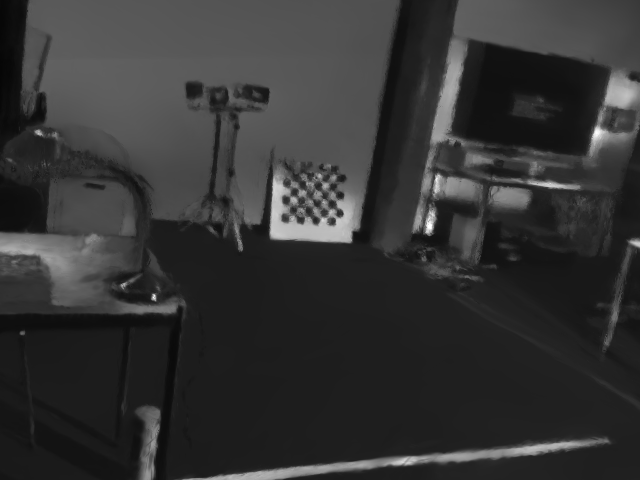}}
		& \gframe{\includegraphics[width=\linewidth]{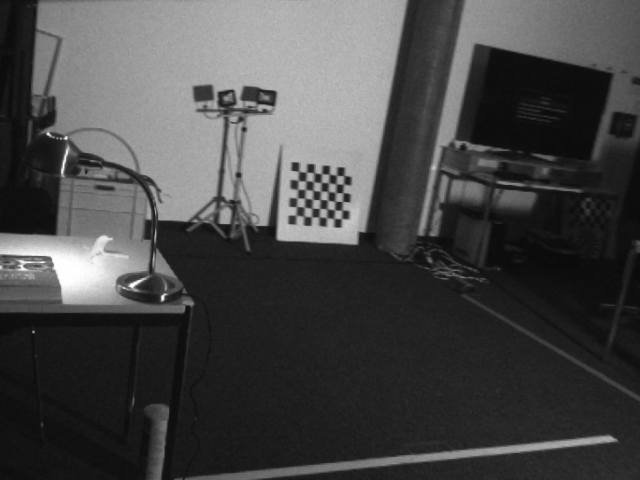}}
		\\

        \rotatebox{90}{\makecell{EDS\_11}}
		& \gframe{\includegraphics[width=\linewidth]{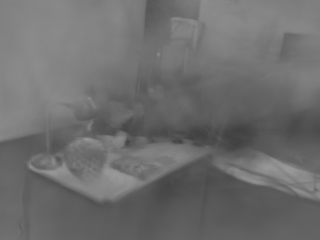}}
		& \gframe{\includegraphics[width=\linewidth]{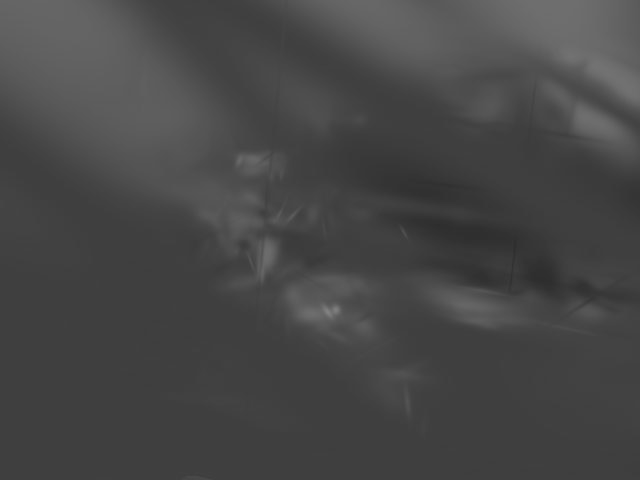}}
		& \gframe{\includegraphics[width=\linewidth]{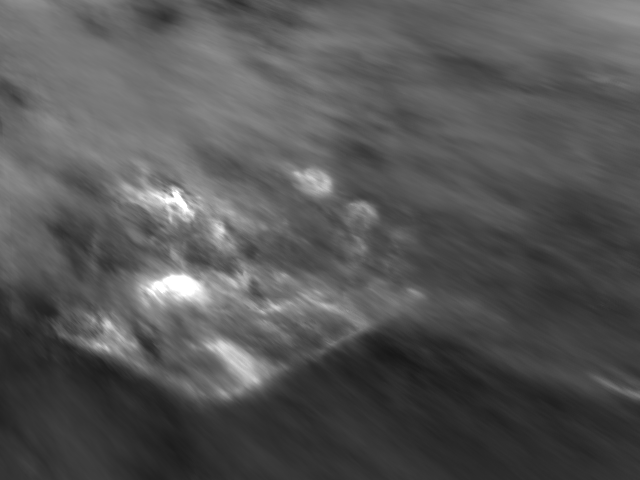}}
		& \gframe{\includegraphics[width=\linewidth]{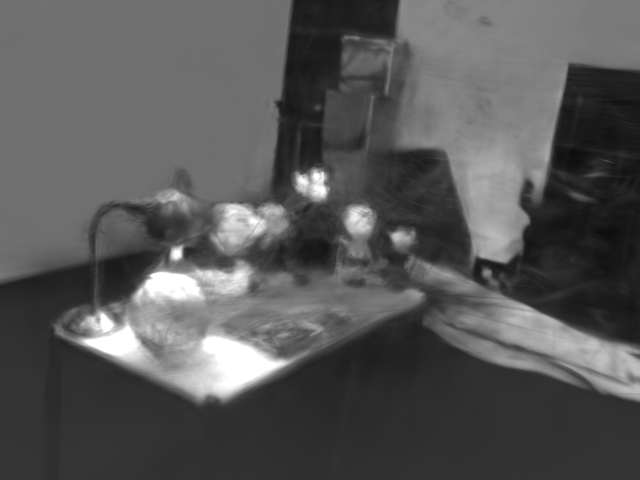}}
		& \gframe{\includegraphics[width=\linewidth]{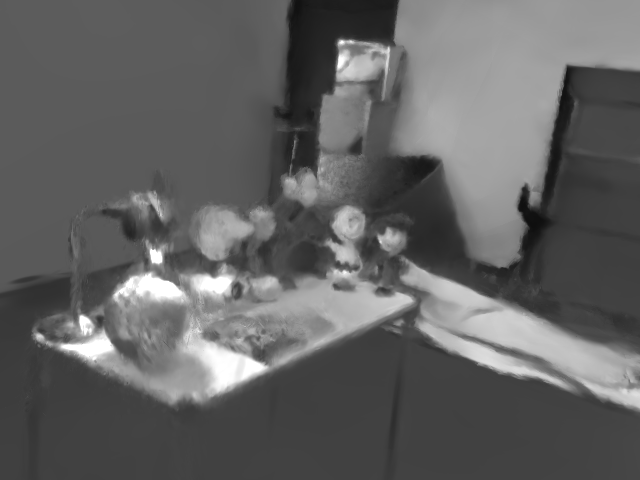}}
		& \gframe{\includegraphics[width=\linewidth]{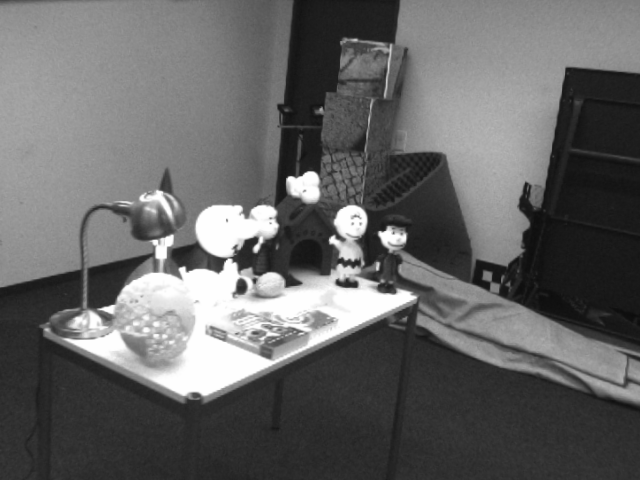}}
		\\   

            \rotatebox{90}{\makecell{EDS\_13}}
		& \gframe{\includegraphics[width=\linewidth]{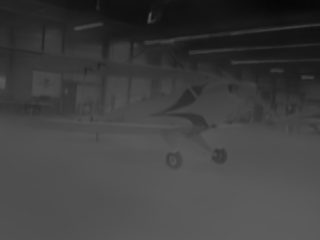}}
		& \gframe{\includegraphics[width=\linewidth]{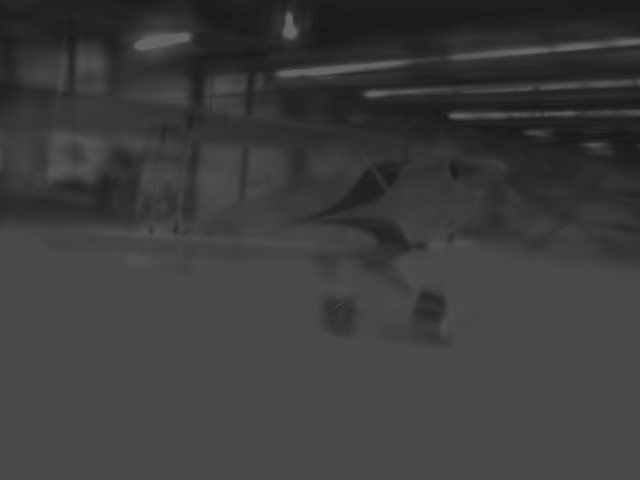}}
		& \gframe{\includegraphics[width=\linewidth]{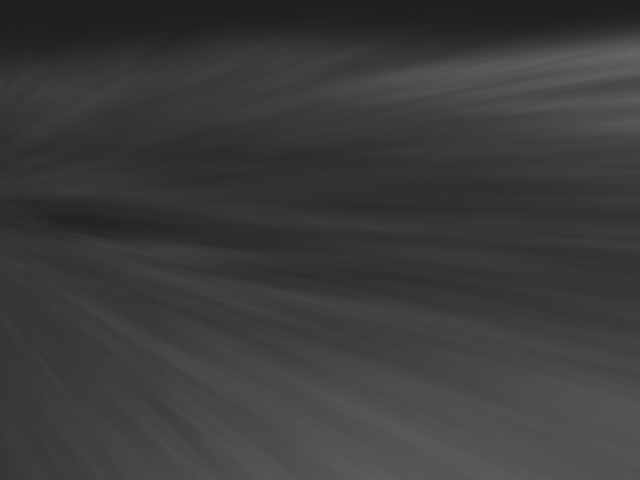}}
		& \gframe{\includegraphics[width=\linewidth]{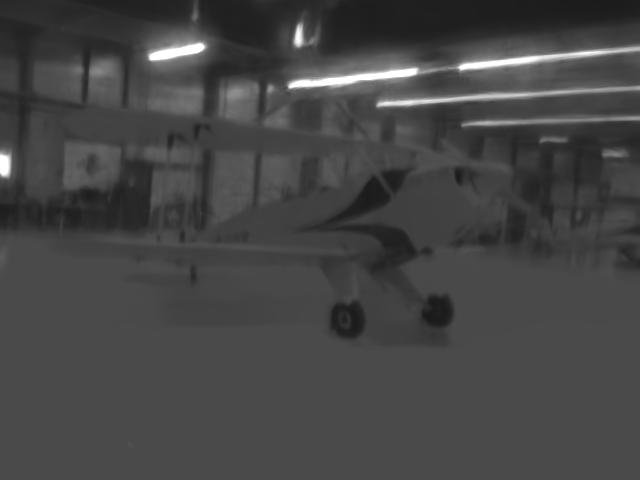}}
		& \gframe{\includegraphics[width=\linewidth]{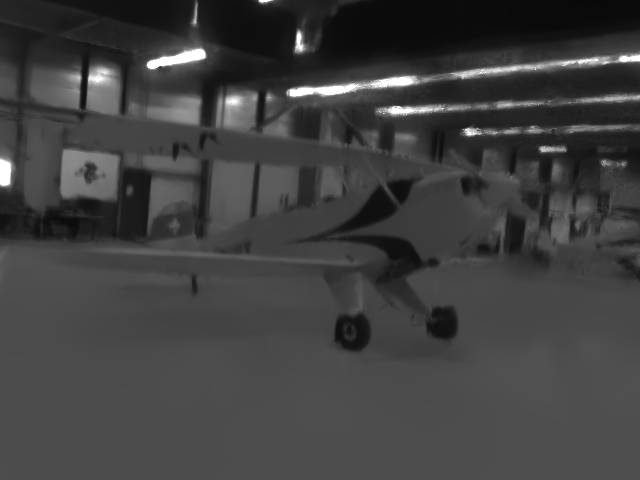}}
		& \gframe{\includegraphics[width=\linewidth]{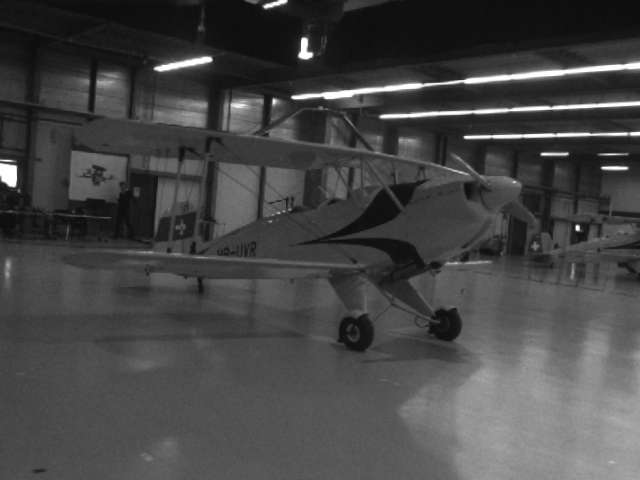}}
		\\

            \rotatebox{90}{\makecell{TUM\_1d-trans}}
		& \gframe{\includegraphics[width=\linewidth]{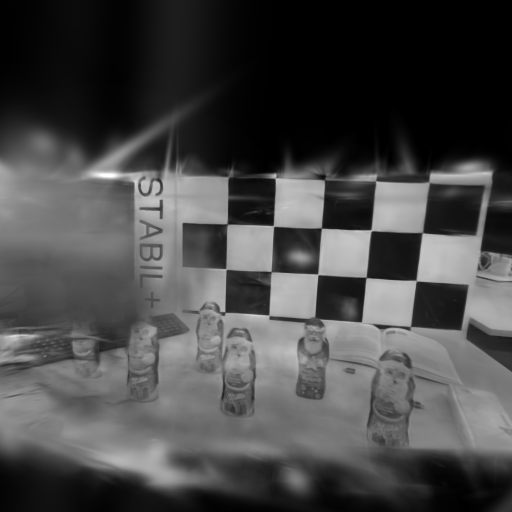}}
		& \gframe{\includegraphics[width=\linewidth]{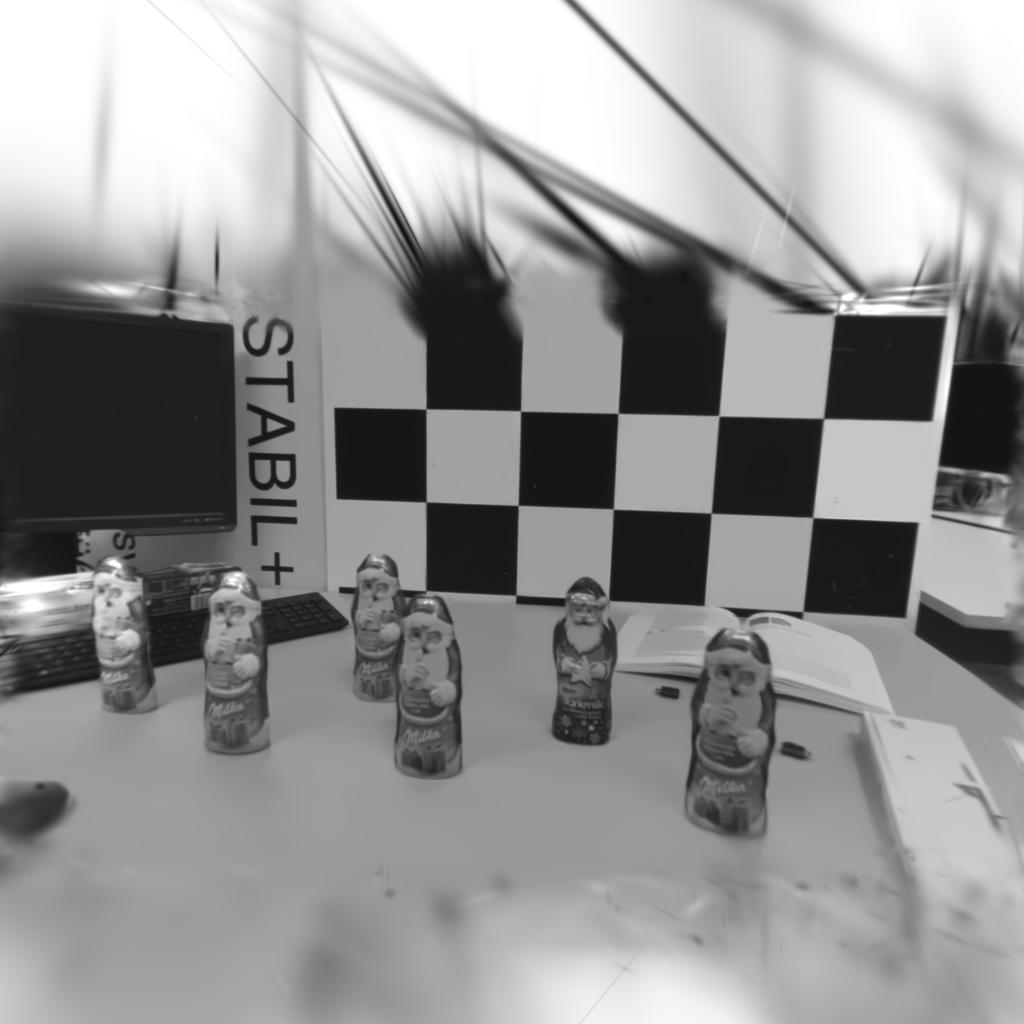}}
		& \gframe{\includegraphics[width=\linewidth]{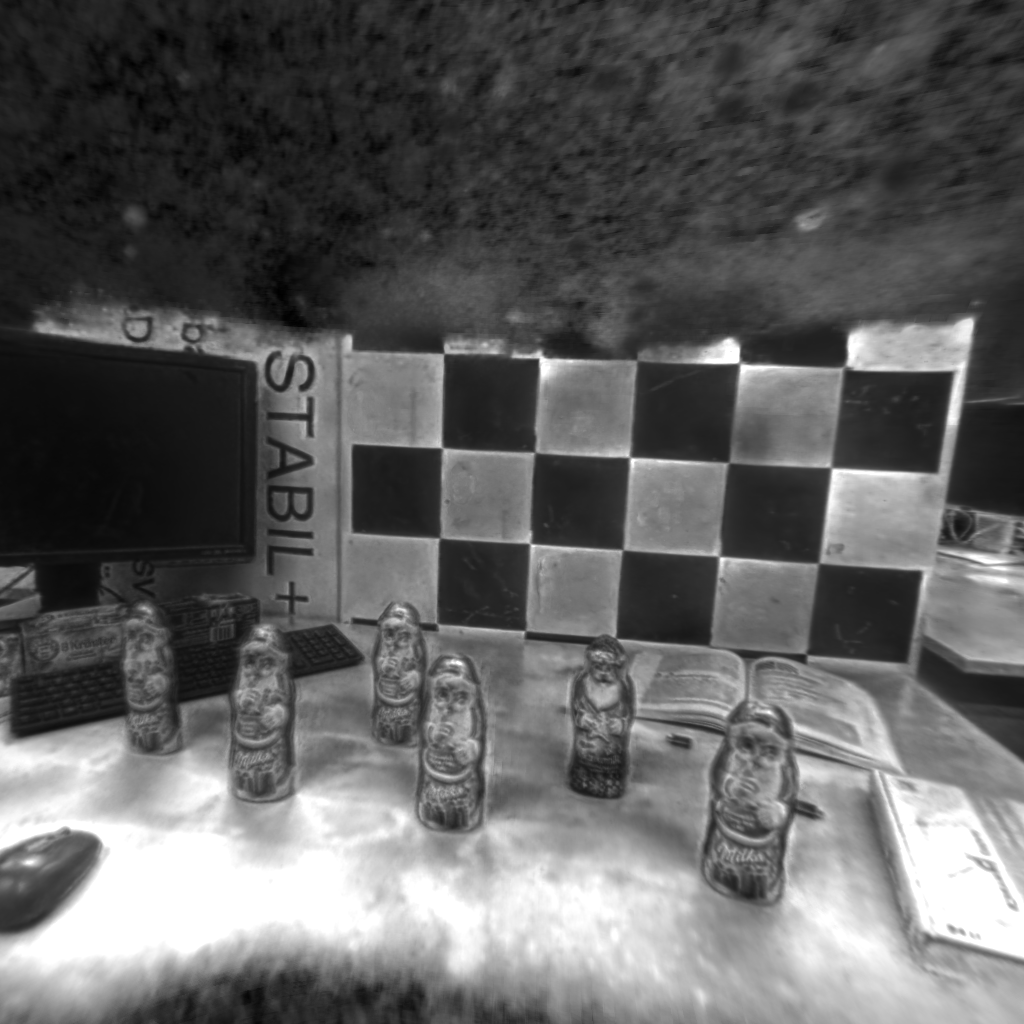}}
		& \gframe{\includegraphics[width=\linewidth]{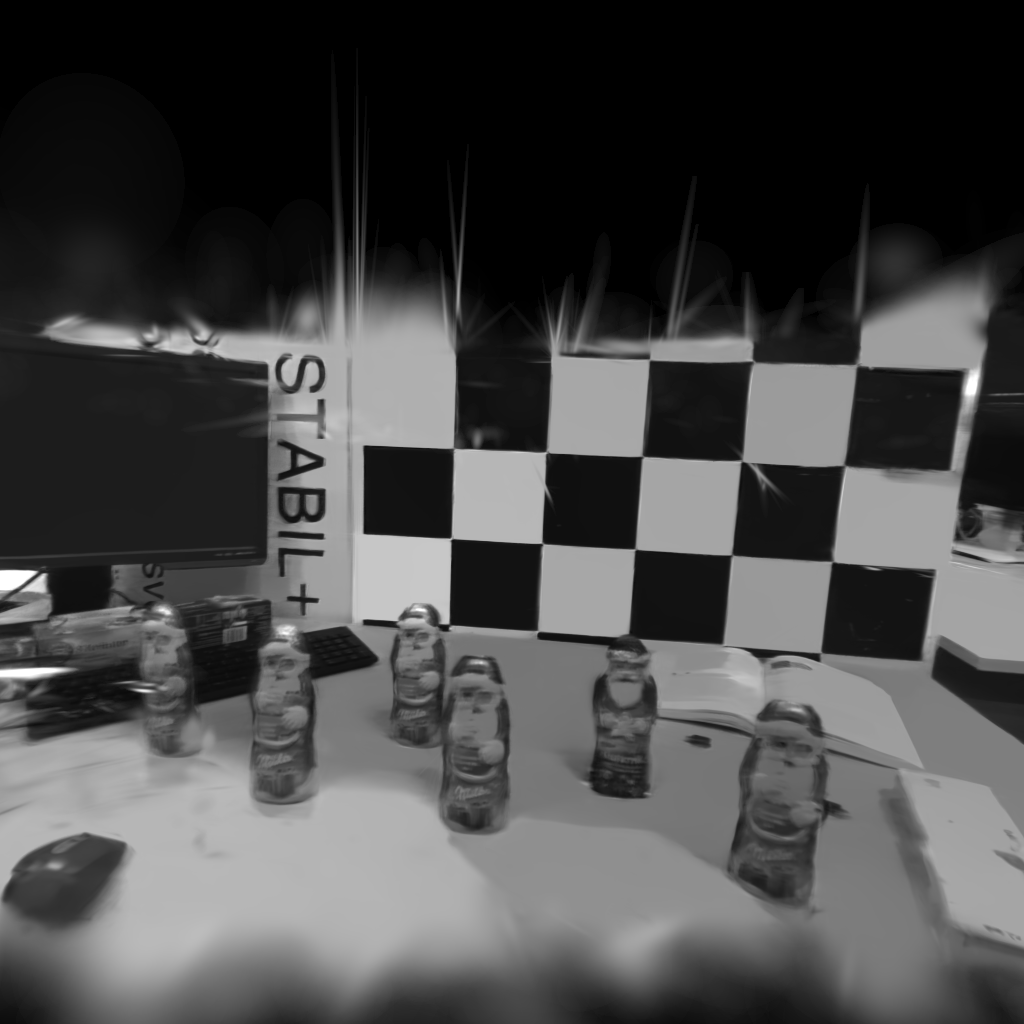}}
		& \gframe{\includegraphics[width=\linewidth]{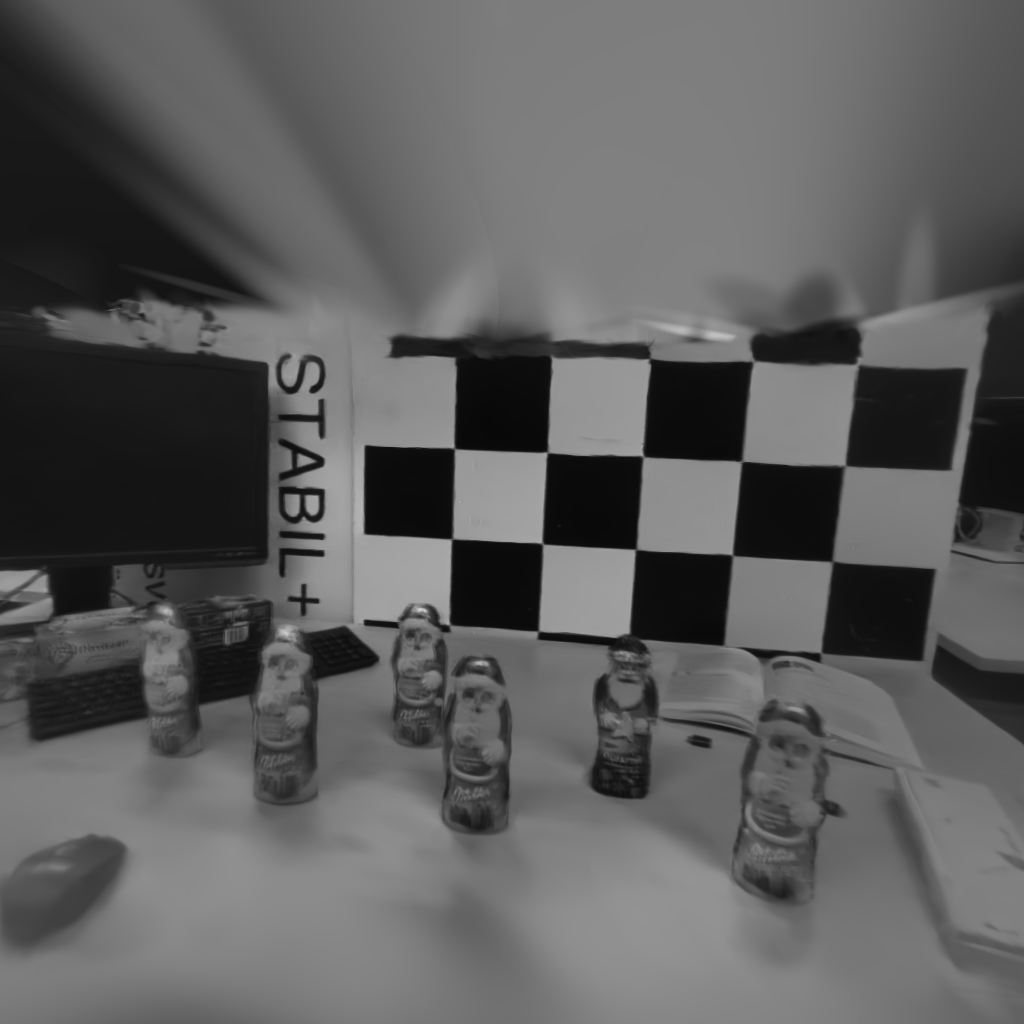}}
		& \gframe{\includegraphics[width=\linewidth]{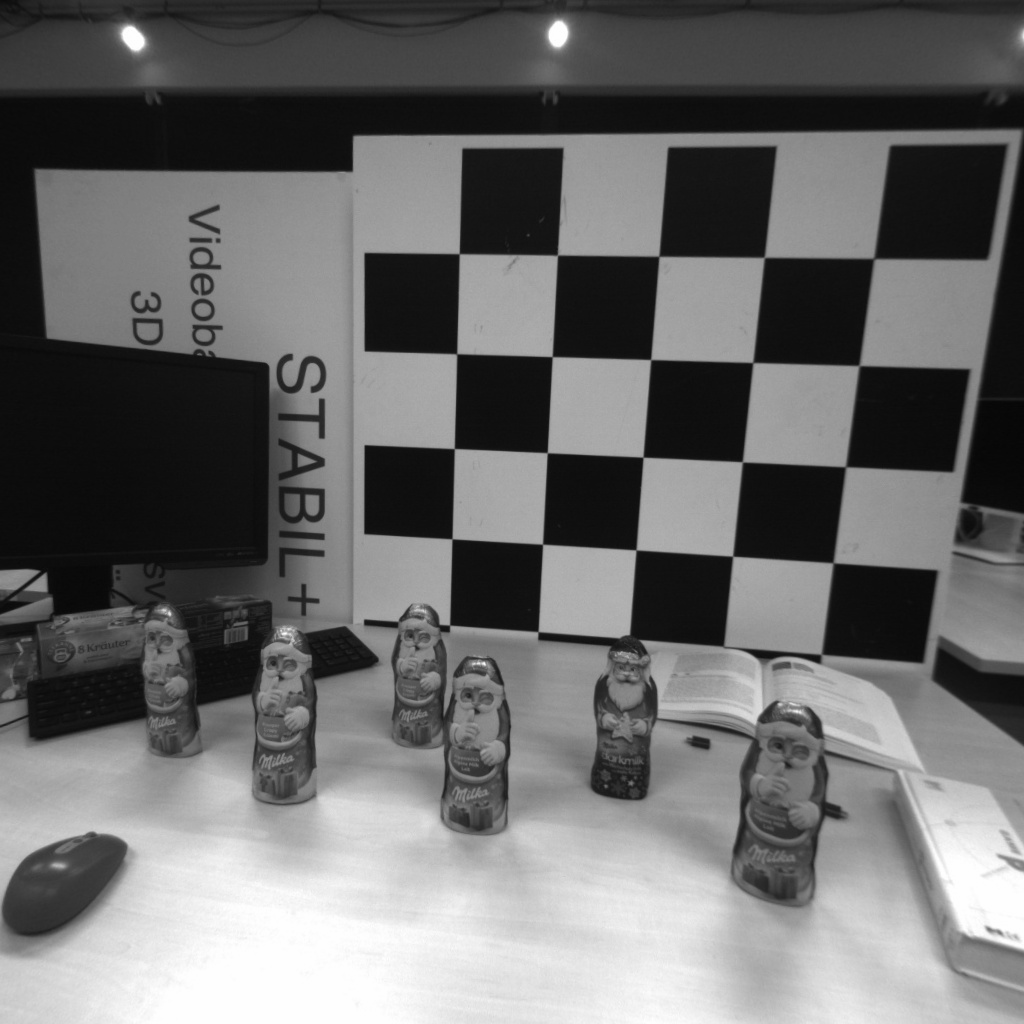}}
		\\

            \rotatebox{90}{\makecell{TUM\_desk2}}
		& \gframe{\includegraphics[width=\linewidth]{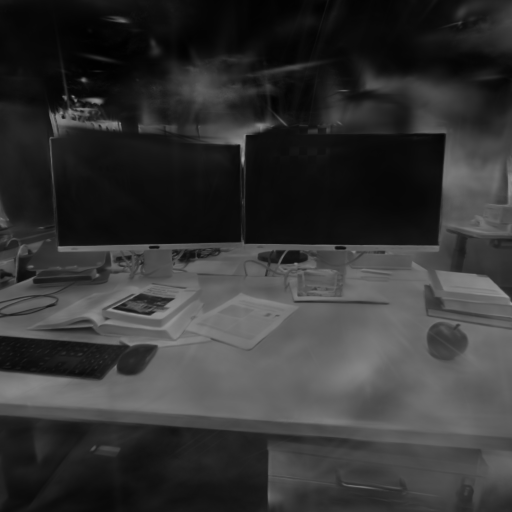}}
		& \gframe{\includegraphics[width=\linewidth]{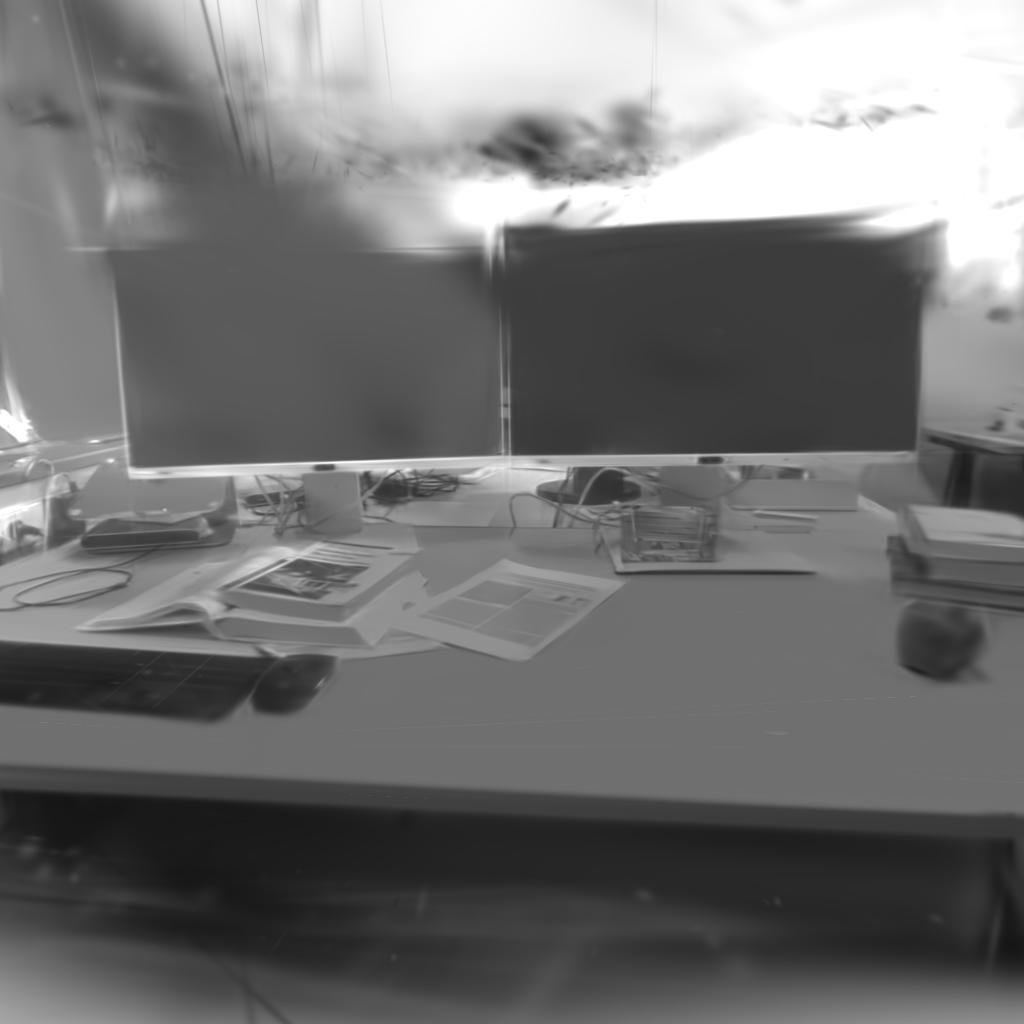}}
		& \gframe{\includegraphics[width=\linewidth]{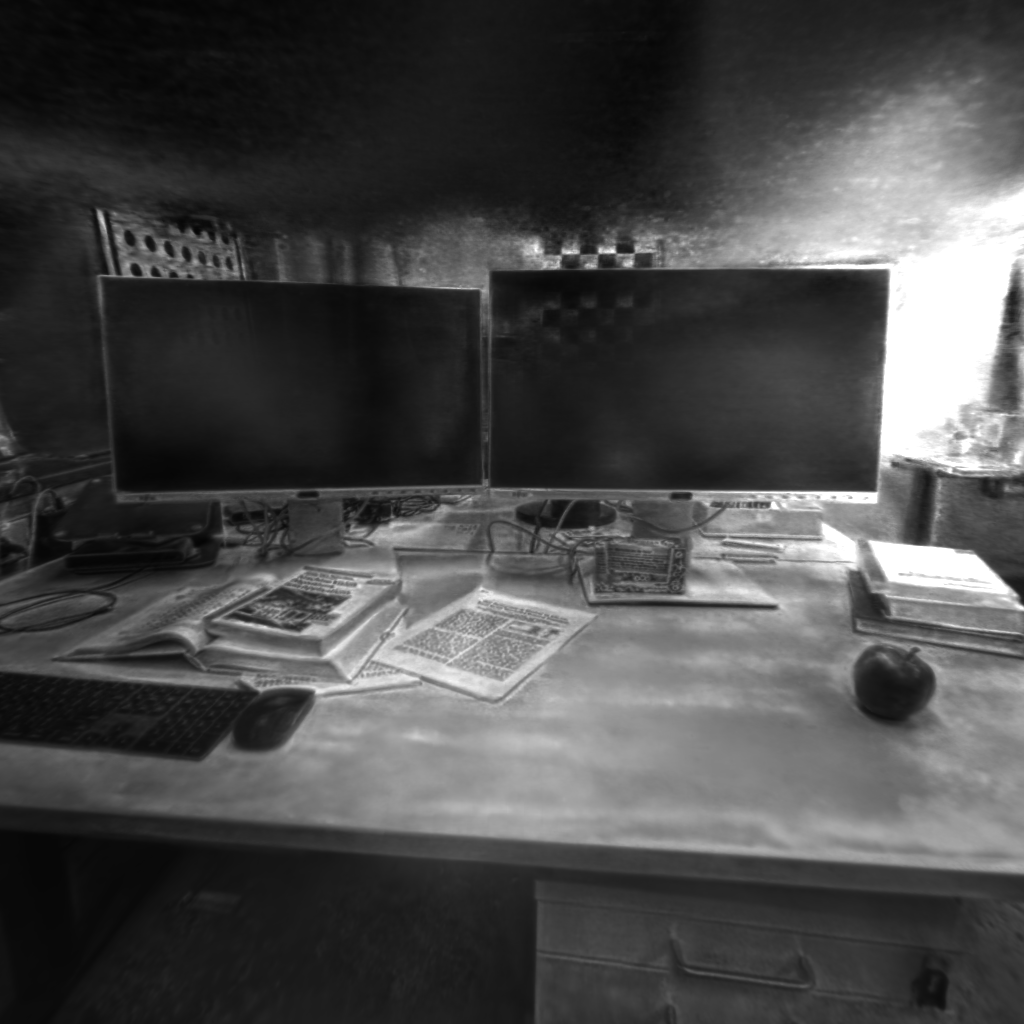}}
		& \gframe{\includegraphics[width=\linewidth]{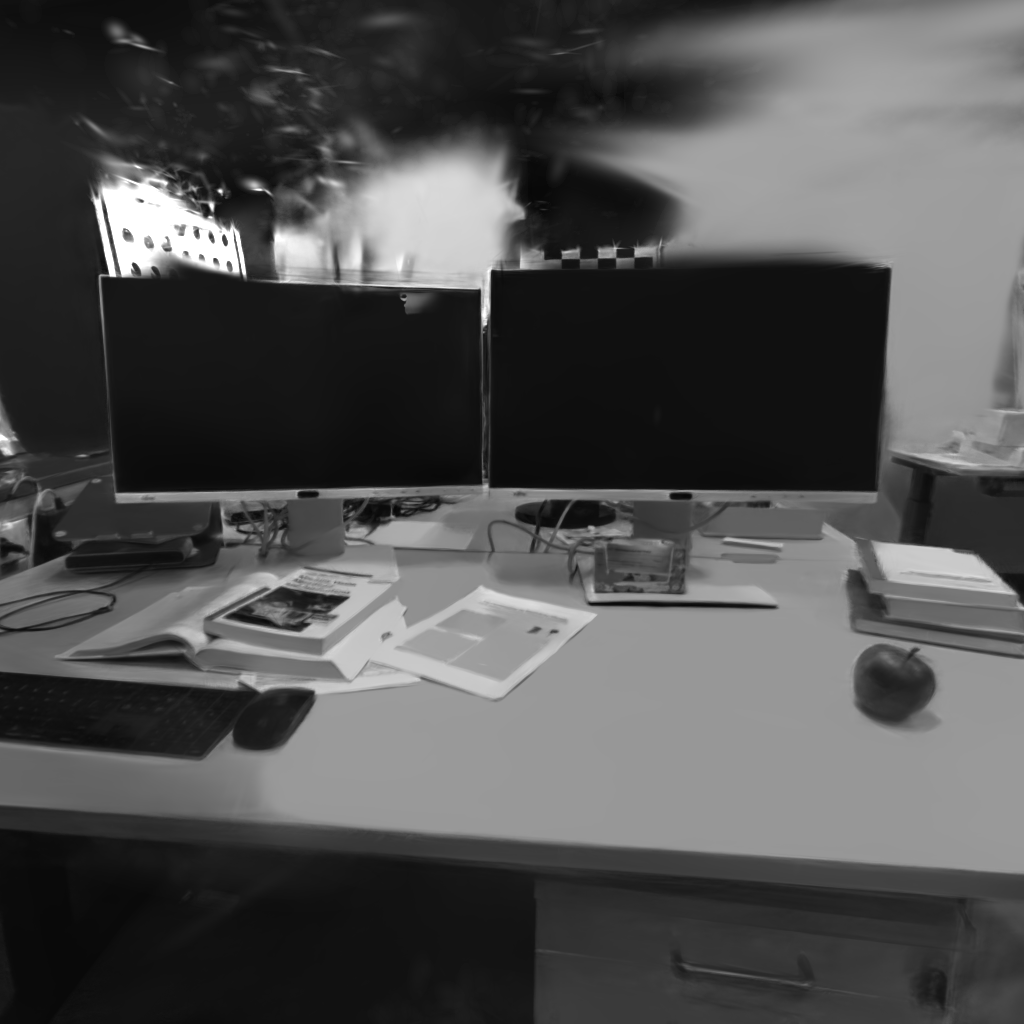}}
		& \gframe{\includegraphics[width=\linewidth]{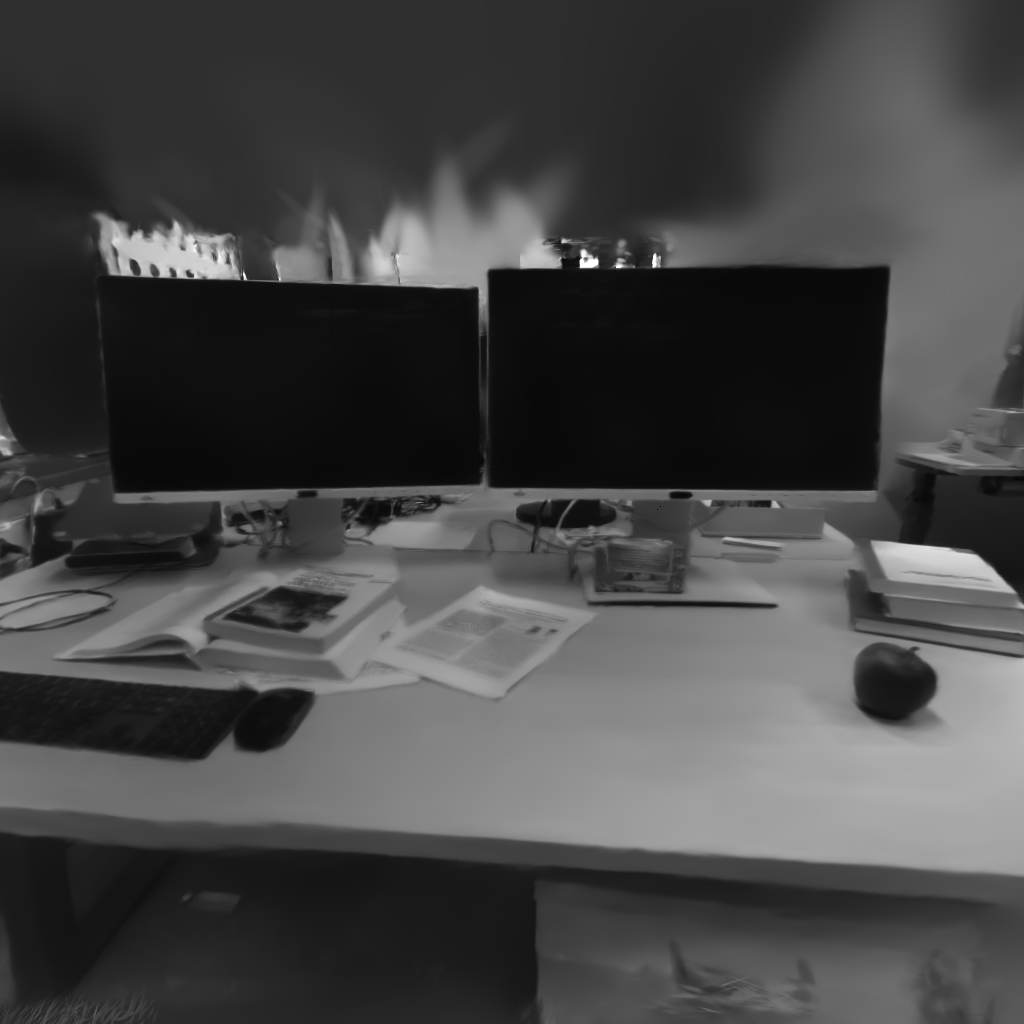}}
		& \gframe{\includegraphics[width=\linewidth]{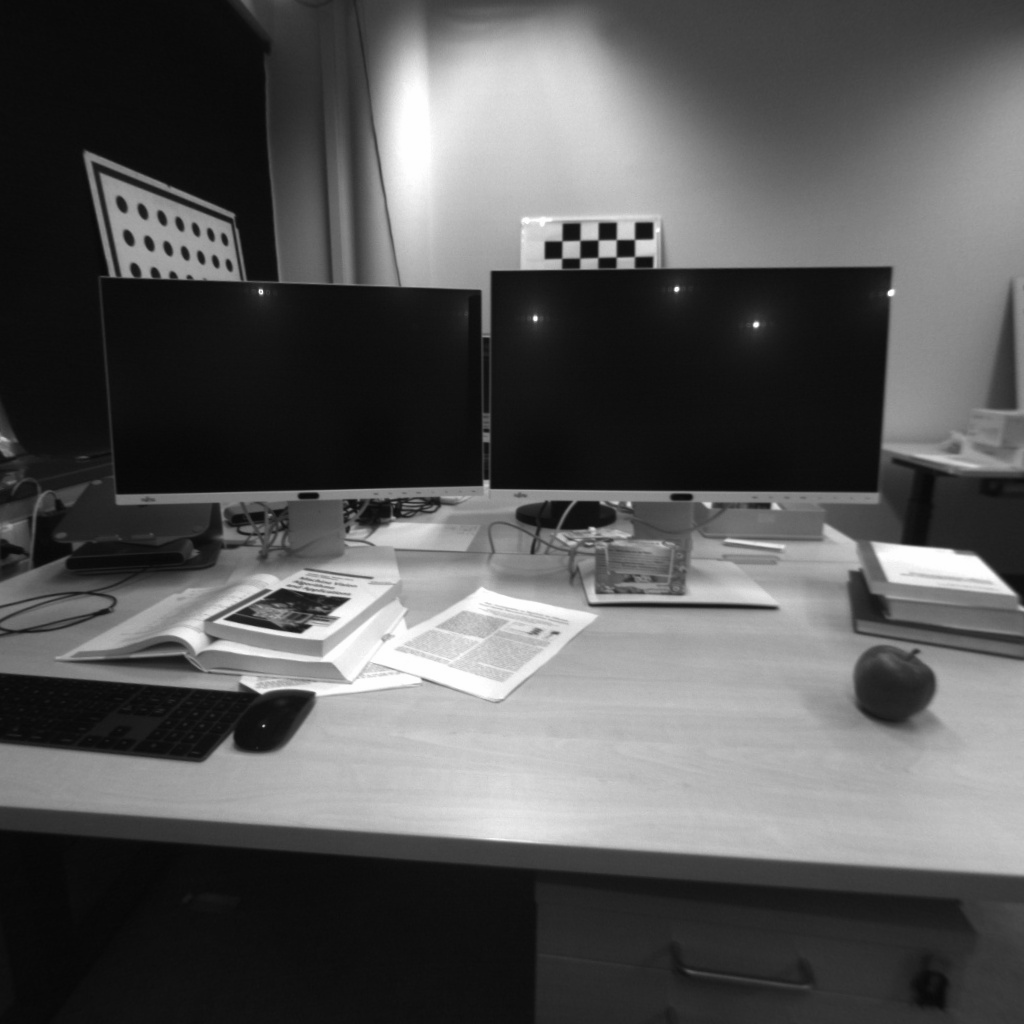}}
		\\
        
		& (a) E2VID + 3DGS
        & (b) IncEventGS \cite{Huang25cvpr}
		& (c) Robust E-NeRF \cite{Low23iccv}
		& (d) EventSplat \cite{Yura25cvpr}
		& (e) Ours
		& (f) GT
		\\
	\end{tabular}
	}
    \vspace{-1ex}
	\caption{\emph{Results on the real-world datasets EDS \cite{Hidalgo22cvpr} and TUM-VIE \cite{Klenk21iros}}. 
    The event camera's field of view in the TUM dataset is narrower than the GT (i.e., frame camera) in the vertical direction.
    }
    \label{fig:result:qualitativeReal}
\end{figure*}

\begin{table*}[t]
\centering
\adjustbox{max width=\linewidth}{%
\setlength{\tabcolsep}{3pt}
\begin{tabular}{ll*{9}{S[round-mode=places,round-precision=3]}}
\toprule
& & \multicolumn{6}{c}{\textbf{EDS} \cite{Hidalgo22cvpr}} &
\multicolumn{3}{c}{\textbf{TUM-VIE} \cite{Klenk21iros}} \\

\cmidrule(l{1mm}r{1mm}){3-8}
\cmidrule(l{1mm}r{1mm}){9-11}
\textbf{Metric} & \textbf{Method} & \textbf{\text{Avg.}} & \text{03}  & \text{07} & \text{08} & \text{11} & \text{13} & \textbf{\text{Avg.}} & \text{1d-trans} &\text{desk2}  \\
\midrule
\multirow{5}{*}{PSNR $\uparrow$} & 
E2VID + 3DGS & 15.51 & 15.67 & 15.05 & 14.03 & 13.83 & 18.96 & 9.524 & 9.382 & 9.664 \\
& Robust E-NeRF (ICCV'23) \cite{Low23iccv} & 16.25 & 19.19 & 14.78 & 14.75 & 14.43 & 18.10 & 11.79  & 9.612 & 13.97 \\
& IncEventGS (CVPR'25) \cite{Huang25cvpr} & 15.21 & 14.13 & 15.76 & 15.89 & 13.83 & 16.46 & 10.09 & 10.13 & 10.05  \\
& EventSplat (CVPR'25) \cite{Yura25cvpr} & 18.86 & \bnum{20.78} & 19.14 & 17.53 & \bnum{17.79} & 19.05 & \novalue & \novalue & \novalue \\
& \textbf{Ours} & \bnum{19.47} & 19.04 & \bnum{20.24} & \bnum{21.03} & 16.73 & \bnum{20.30} & \bnum{13.09} & \bnum{11.97} & \bnum{14.20}  \\
\midrule
\multirow{5}{*}{SSIM $\uparrow$} & 
E2VID + 3DGS & 0.692 & 0.716 & 0.689 & 0.642 & 0.691 & 0.723 & 0.516 & 0.525 & 0.507 \\
& Robust E-NeRF (ICCV'23) \cite{Low23iccv} & 0.739 & \bnum{0.846} & 0.815 & 0.735 & 0.569 & 0.729 & 0.573 & 0.504 & 0.642 \\
& IncEventGS (CVPR'25) \cite{Huang25cvpr} & 0.691 & 0.756 & 0.684 & 0.692 & 0.648 & 0.676 & 0.533 & 0.536 & 0.529 \\
& EventSplat (CVPR'25) \cite{Yura25cvpr} &  0.792 & 0.835 & 0.816 & 0.745 & 0.789 & 0.774 & \novalue & \novalue & \novalue \\
& \textbf{Ours} &  \bnum{0.816} & 0.819 & \bnum{0.855} & \bnum{0.814} & \bnum{0.790} & \bnum{0.804} & \bnum{0.716} & \bnum{0.665} & \bnum{0.766}  \\
\midrule
\multirow{5}{*}{LPIPS $\downarrow$} & 
E2VID + 3DGS & 0.375 & 0.266 & 0.378 & 0.402 & 0.415 & 0.415 & 0.759 & 0.790 & 0.728 \\
& Robust E-NeRF (ICCV'23) \cite{Low23iccv} & 0.543 & 0.324 & 0.476 & 0.567 & 0.700 & 0.650  & 0.588 &  0.721 & 0.454\\
& IncEventGS (CVPR'25) \cite{Huang25cvpr} & 0.561 & 0.356 & 0.557 & 0.631 & 0.588 & 0.674 & 0.685 & 0.707 & 0.663 \\
& EventSplat (CVPR'25) \cite{Yura25cvpr} & 0.362 & \bnum{0.239} & 0.351 & 0.424 & \bnum{0.391} & \bnum{0.407} & \novalue & \novalue & \novalue \\
& \textbf{Ours} & \bnum{0.357} & 0.272 & \bnum{0.335} & \bnum{0.369} & 0.396 & 0.414 & \bnum{0.411} & \bnum{0.497} & \bnum{0.324}  \\
\bottomrule
\end{tabular}
}
\vspace{-1ex}
\caption{\emph{Results on standard, real-world datasets EDS and TUM-VIE}. Best in bold.} 
\label{tab:result:eds}
\end{table*}

\subsection{Photometric Loss}
\label{sec:method:spatialrendering}

The IWE~\eqref{eq:IWE} represents not only motion-corrected edges,
but also their strength (e.g., intensity gradient) with respect to the flow direction \cite{Gallego18cvpr,Zhang22pami}.
Hence, we may use the IWE to design not only geometric loss terms but also photometric ones (bottom branch of \cref{fig:method}, and examples in \cref{fig:depthpoint} columns (a), (c)).
This is inspired by methods in the literature that define losses on brightness increment images obtained from grayscale information 
\cite{Bryner19icra,Paredes21cvpr,Hidalgo22cvpr,Shiba23pami}. 

Specifically, following the event generation model \cite{Gallego20pami}, 
the prediction of the scene's edge strength at time $\tref$ is:
\begin{equation}
\label{eq:brightnesschange}
    \hat{H} (\bx; \tref) 
    \doteq \frac{\partial \log \raycolor}{\partial t} \Delta t \approx - \nabla \log(\raycolor) \cdot \velflow \Delta t,
\end{equation}
where $\raycolor \equiv \raycolor(\bx)$ is the rendered frame \eqref{eq:gs:color} from the viewpoint of camera pose $(R(\tref), T(\tref))$ 
and $\velflow\equiv \velflow(\bx,\tref)$ is the motion field \eqref{eq:motionField} obtained using the rendered depth at time $\tref$ (bottom branch of \cref{fig:method}).
This corresponds to the instantaneous rate of brightness change in the optical flow direction \cite{Zhang22pami}.
Note that $\raycolor$ may represent color for the simulated/color event cameras, or gray (intensity) for the standard event cameras.
The dense-pixel (i.e., radiance) rendering happens once in each optimization step (see \cref{sec:experim:renderonce}).

Finally, photometric errors between the IWE (with $b_k=p_k$) and its prediction \eqref{eq:brightnesschange} are defined by the $L^2$-norm and the Structural Similarity Index Measure (SSIM) \cite{Wang04tip}: 
\begin{equation}
\begin{split}
\label{eq:loss:photometric}
    \loss_\text{p} & \doteq \frac{1}{|\Omega|} \| \IWE(\bx; \tref) - \hat{H}(\bx; \tref) \|^2, \\
    \loss_\text{s} & \doteq 1- \text{SSIM}\bigl(\IWE(\bx; \tref),\, \hat{H}(\bx; \tref) \bigr).
\end{split}
\end{equation}
In practice, the SSIM loss is computed for each polarity: $\loss_\text{s+},\loss_\text{s-}$, which we omit in the following notation for simplicity.

We find that %
warping is more useful to leverage the high temporal resolution than the simple pixel-wise accumulation of polarities %
used in most event-based GS and NeRF literature (\cite{Rudnev23cvpr,Huang25cvpr,Yura25cvpr}),
because the latter: 
($i$) may result in blurry edge images that discard the fine temporal resolution,
($ii$) incurs neutralization (cancellation of event polarities),
($iii$) requires two dense intensity renderings to compute the photometric loss,
($iv$) omits a dependency on the unknown depth/flow that can be useful during optimization \cite{Guo25iccv}.

\subsection{Combined Loss Function}
\label{sec:method:loss}

For each slice of events $\cE$ we use the middle timestamp as a reference, $\tref \doteq t_\text{mid}$.
The total loss is a weighted sum of the event-alignment loss (CMax) and the photometric losses:
\begin{equation}
\label{eq:theloss}
\loss \doteq \lambda_\text{c} \loss_\text{c} + \lambda_\text{p} \loss_\text{p} + \lambda_\text{s} \loss_\text{s}.
\end{equation}

\subsection{Initialization}
\label{sec:method:initialization}

The initialization of the 3D Gaussians is important.
For example, it is common practice for frame-based GS methods to use COLMAP \cite{Schoenberger16cvpr} to favor initial Gaussians on scene texture and edges.
Indeed, prior work EventSplat \cite{Yura25cvpr} uses intensity reconstruction and runs COLMAP for initialization.
However, it relies on the pretrained E2VID model \cite{Rebecq19pami} as prior.
We propose using the $\IWE(\bx;t_\text{mid})$ without polarity and the rendered image $\raycolor(\pixelX)$ for initialization, keeping the rest of the pipeline untouched.
This favors initial 3D Gaussians around scene structures because the IWE responds to edges.
We find that IWEs produce better initialization than images of pixel-wise accumulation of events because of their sharpness, which narrows down the initial possible locations of the Gaussian centers (see \cref{sec:suppl:initialization}).

\section{Experiments}
\label{sec:experim}

\subsection{Datasets, Metrics, and Baselines}
\label{sec:experim:dataset}

\textbf{Datasets}.
We use standard datasets for event-based NeRF and GS works, both on simulated and real data.
\emph{EDS} \cite{Hidalgo22cvpr} is a real-world dataset of indoor scenarios, recorded with a VGA event camera ($640\times 480$ px), an RGB camera, an IMU, and ground-truth poses from motion capture.
The sequences include challenging scenes, such as flickering light sources.
\emph{TUM-VIE} \cite{Klenk21iros} is another real-world dataset, acquired with an HD event camera ($1280 \times 720$ px, i.e., 1 megapixel) and with ground-truth poses.
It consists of indoor and outdoor sequences recorded with the sensor rig mounted on a helmet.
We use indoor sequences following prior work.
\emph{Robust E-NeRF} \cite{Low23iccv} contributes a synthetic color event dataset with a $800\times800$ px resolution and color pixels following the Bayer pattern.

\textbf{Evaluation Metrics}.
Following prior work \cite{Low23iccv,Yura25cvpr}, reconstruction performance is measured with standard metrics 
on view synthesis quality:
Peak Signal-to-Noise Ratio (PSNR),
SSIM,
and Learned Perceptual Image Patch Similarity (LPIPS).
Real-world datasets use poses from colocated frame-based cameras for the evaluation of rendering.
In addition, we follow prior work and apply gamma correction before computing the evaluation metrics.

\textbf{Baselines}.
Our baselines are among the best event-only NeRF and GS methods in the literature.
First, we use the two-stage approach of E2VID image reconstruction \cite{Rebecq19pami} and frame-based GS, termed ``E2VID + 3DGS''.
For event-based GS methods (the two-rendering approaches), we retrain \emph{IncEventGS} \cite{Huang25cvpr},
and copy the results from \emph{EventSplat} \cite{Yura25cvpr} because it has no available code.
We also compare with the state-of-the-art NeRF method \emph{Robust E-NeRF} \cite{Low23iccv} that uses event-by-event loss computation.

\textbf{Hyper-parameters.}
For all sequences the contrast threshold is set to $\contrast=0.25$.
The loss weights in \eqref{eq:theloss} are set to $\lambda_\text{c}=0.125, \lambda_\text{p}=500 ,\lambda_\text{s}=1 $.
The number of events is $\numEvents=125$k for EDS and synthetic data \cite{Low23iccv}, and $\numEvents=500$k for TUM-VIE.
We further test the robustness of the method to the choice of $\numEvents$. %
The initialization steps are $10$k, and the entire training steps are $40$k for all sequences.

\def\textWidth{0.02\linewidth}
\def\figWidth{0.16\linewidth}
\begin{figure*}[t]
	\centering
    {%
    \footnotesize
    \setlength{\tabcolsep}{1pt}
	\begin{tabular}{
	>{\centering\arraybackslash}m{\textWidth}
	>{\centering\arraybackslash}m{\figWidth} 
	>{\centering\arraybackslash}m{\figWidth} 
	>{\centering\arraybackslash}m{\figWidth} 
	>{\centering\arraybackslash}m{\figWidth}
	>{\centering\arraybackslash}m{\figWidth}}

            \rotatebox{90}{\makecell{Chair}}
		& \gframe{\includegraphics[trim={0 0 0 1.5cm},clip,width=\linewidth]{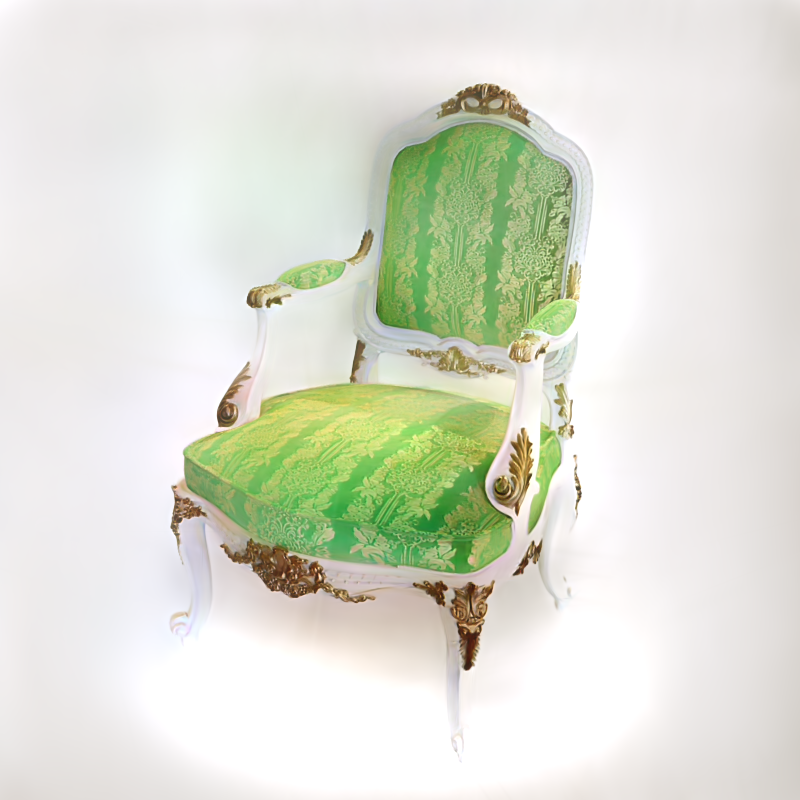}}
		& \gframe{\includegraphics[trim={0 0 0 1.5cm},clip,width=\linewidth]{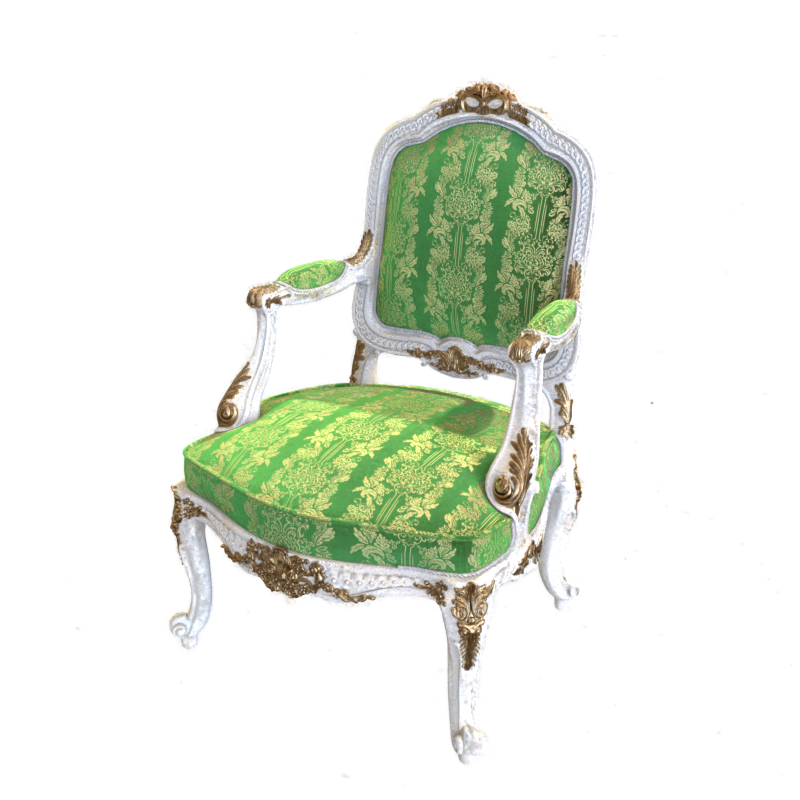}}
		& \gframe{\includegraphics[trim={0 0 0 1.5cm},clip,width=\linewidth]{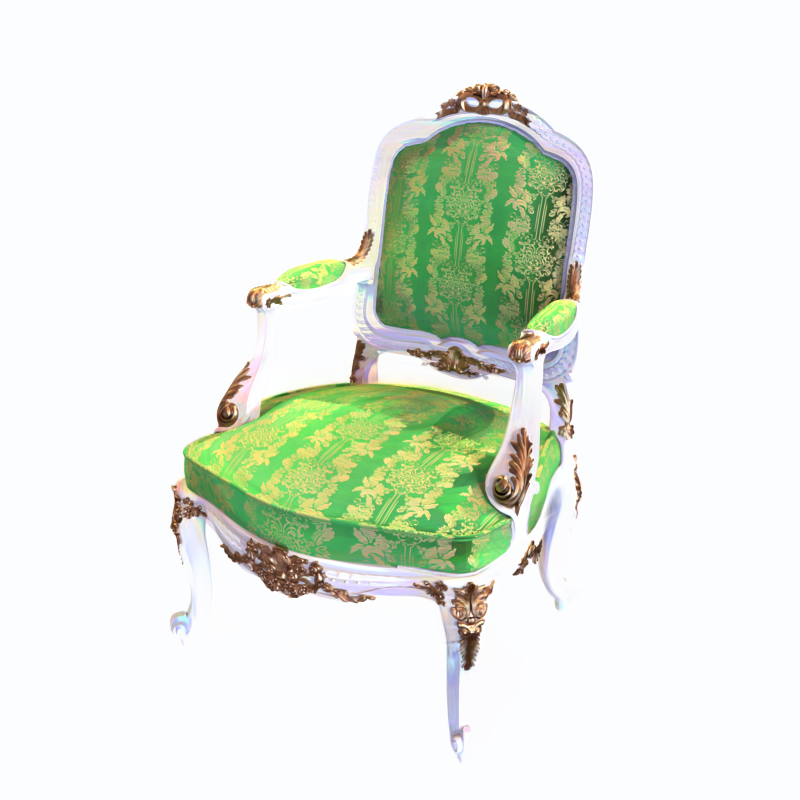}}
		& \gframe{\includegraphics[trim={0 0 0 1.5cm},clip,width=\linewidth]{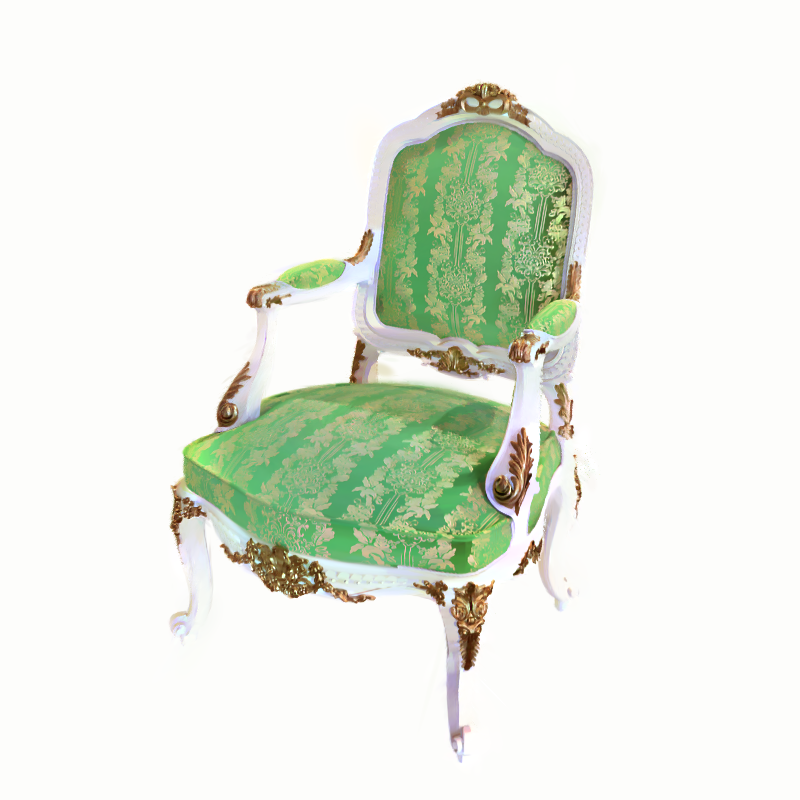}}
		& \gframe{\includegraphics[trim={0 0 0 1.5cm},clip,width=\linewidth]{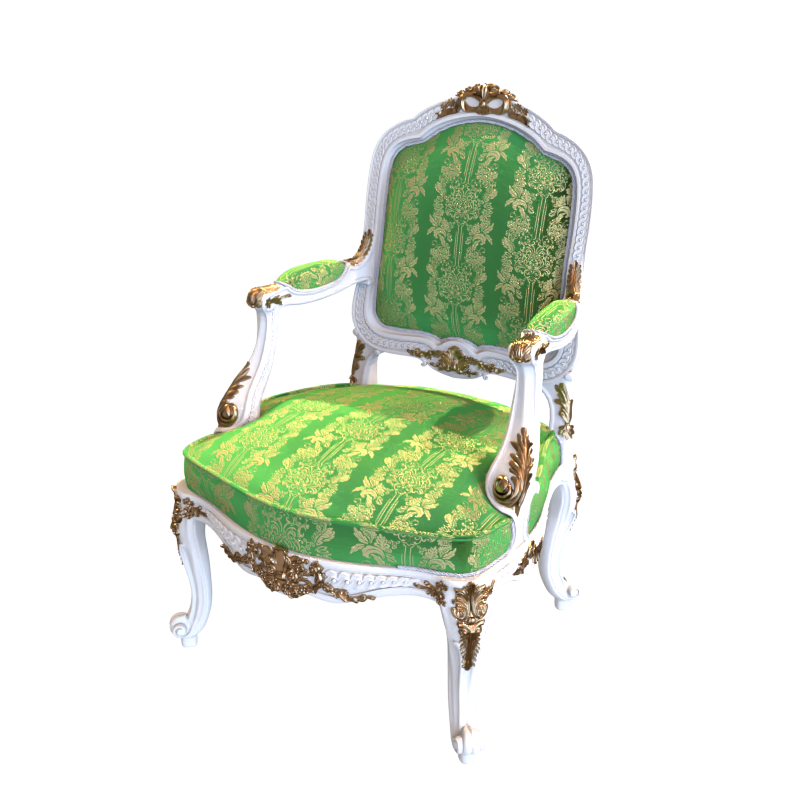}}
		\\

            \rotatebox{90}{\makecell{Drums}}
		& \gframe{\includegraphics[width=\linewidth]{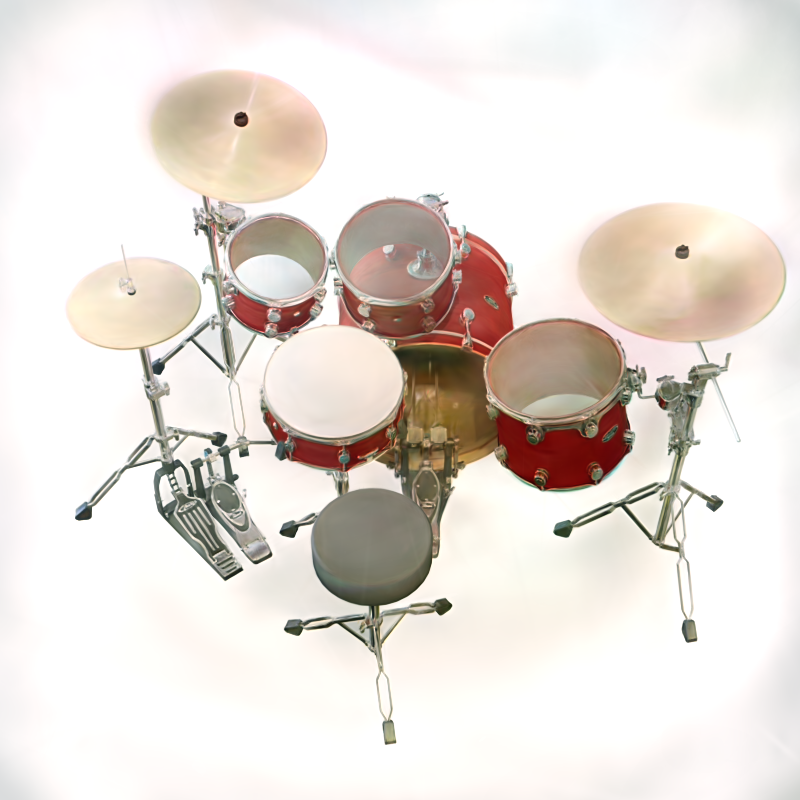}}
		& \gframe{\includegraphics[width=\linewidth]{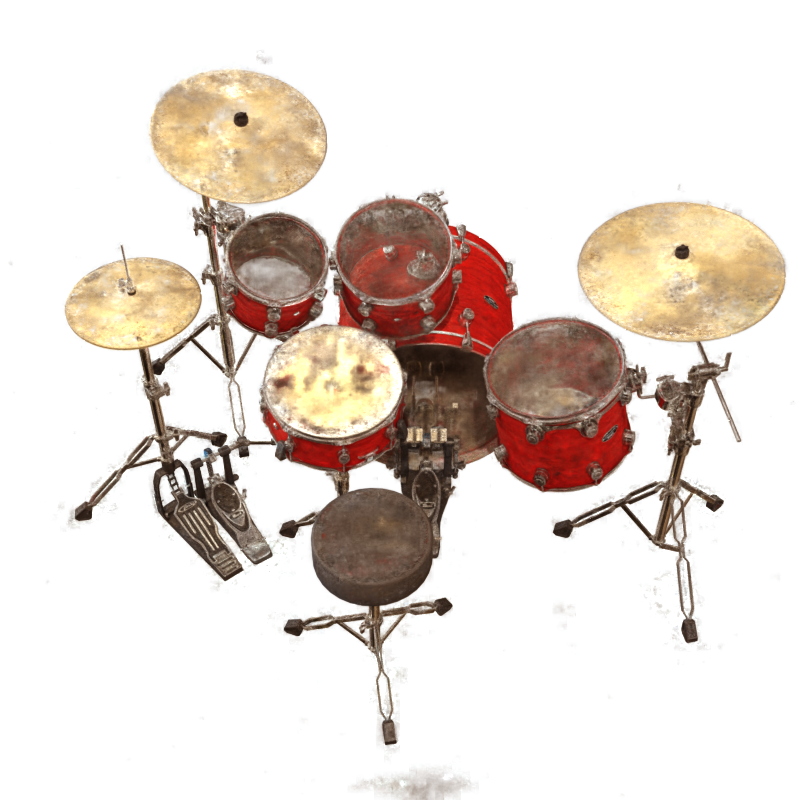}}
		& \gframe{\includegraphics[width=\linewidth]{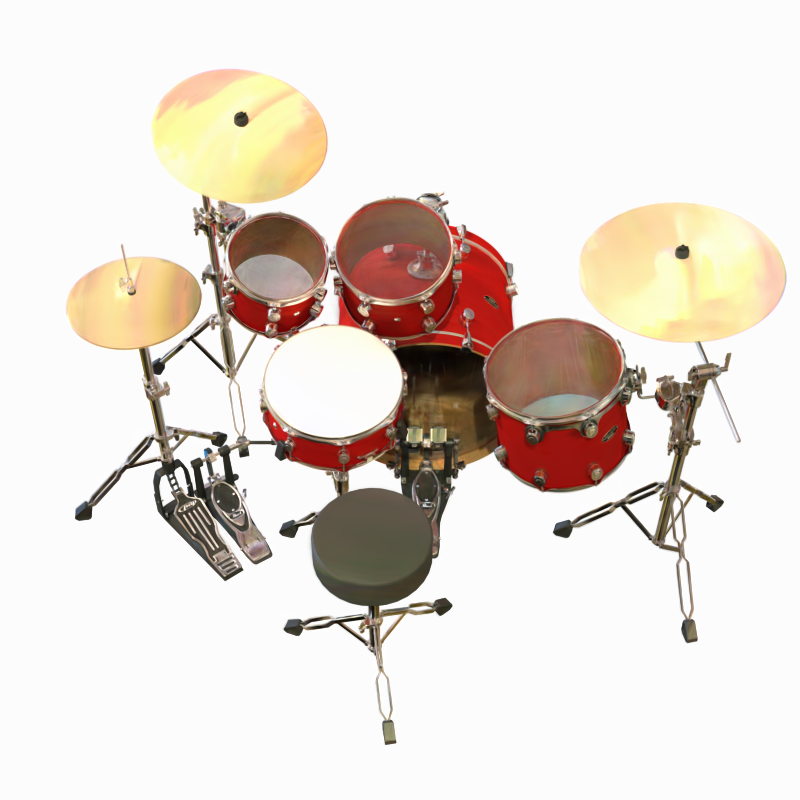}}
		& \gframe{\includegraphics[width=\linewidth]{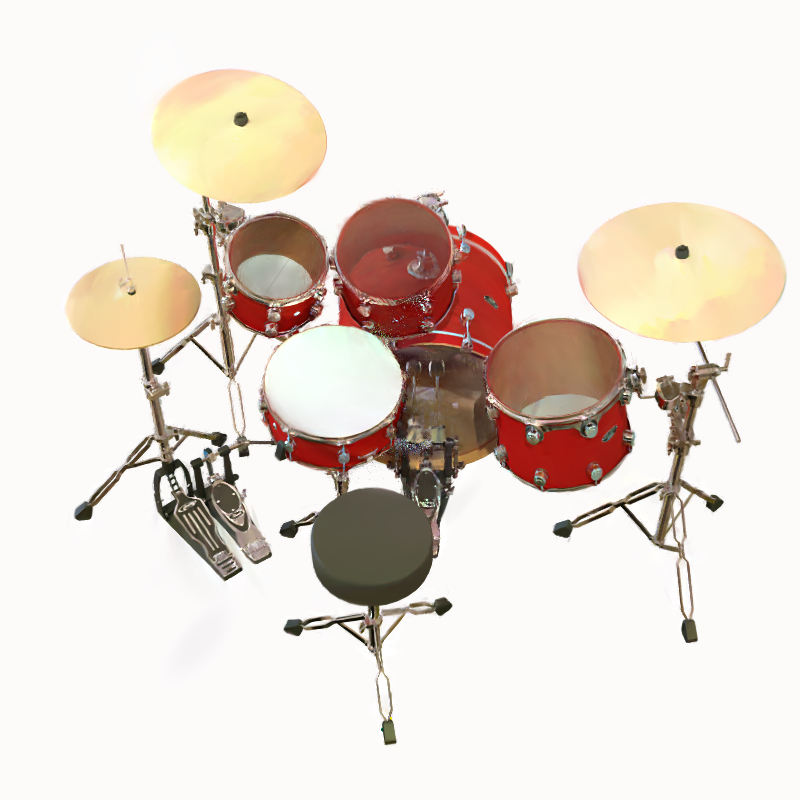}}
		& \gframe{\includegraphics[width=\linewidth]{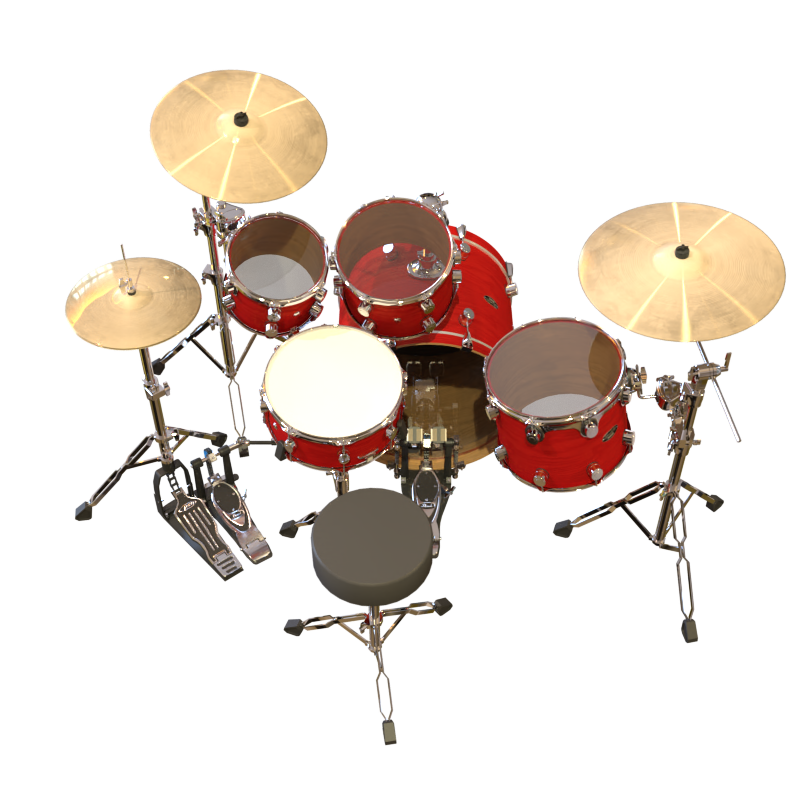}}
		\\

            \rotatebox{90}{\makecell{Materials}}
		& \gframe{\includegraphics[trim={0 0 0 5cm},clip,width=\linewidth]{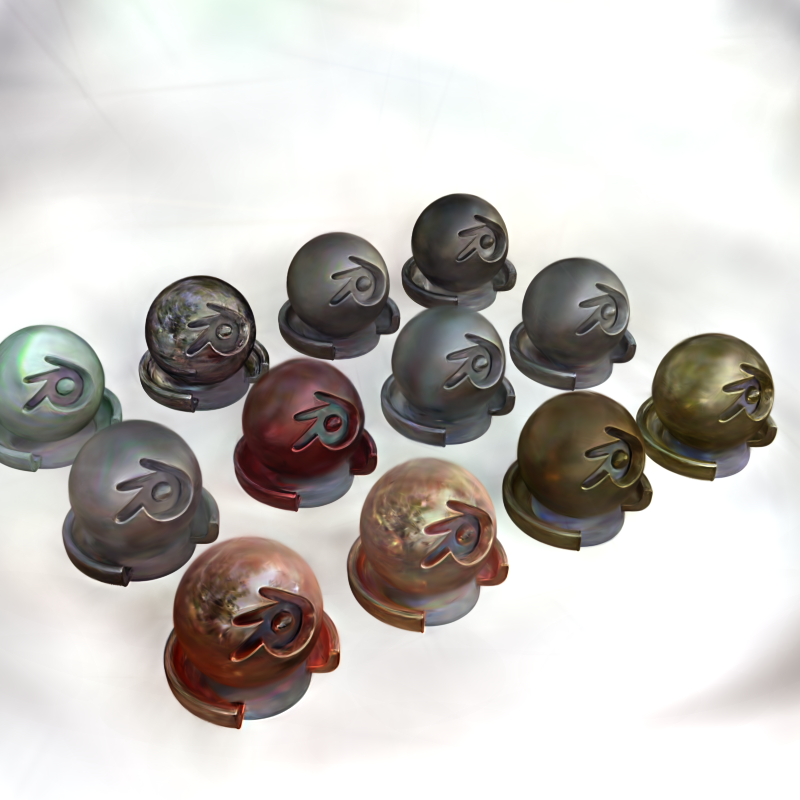}}
		& \gframe{\includegraphics[trim={0 0 0 5cm},clip,width=\linewidth]{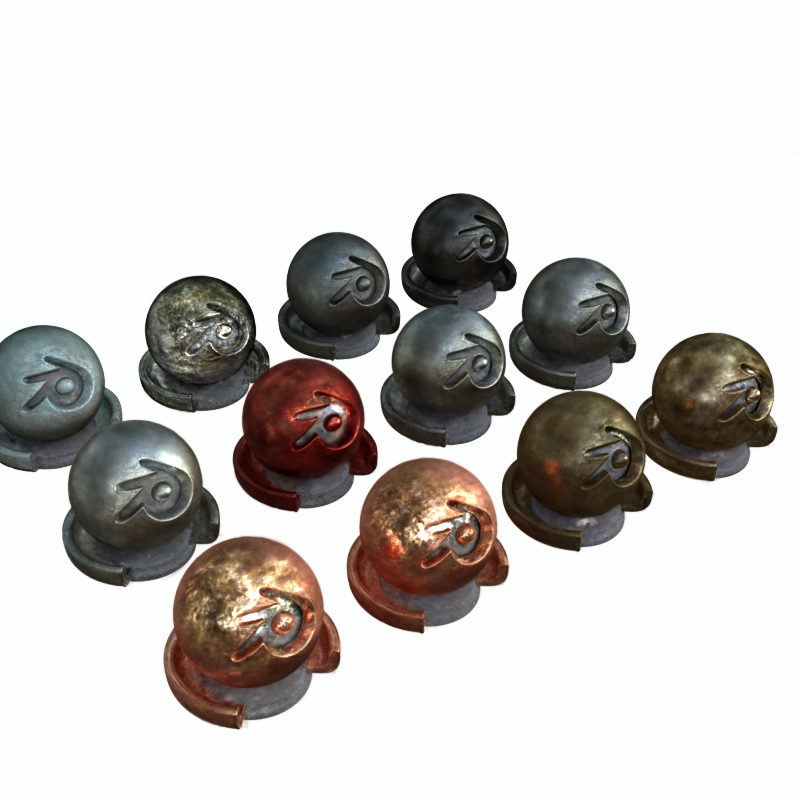}}
		& \gframe{\includegraphics[trim={0 0 0 5cm},clip,width=\linewidth]{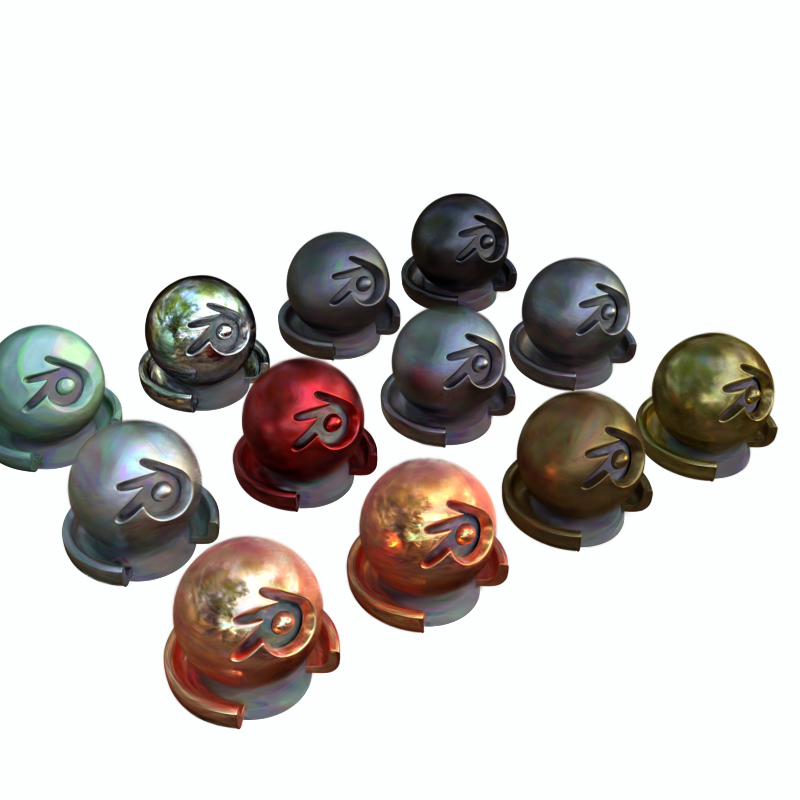}}
		& \gframe{\includegraphics[trim={0 0 0 5cm},clip,width=\linewidth]{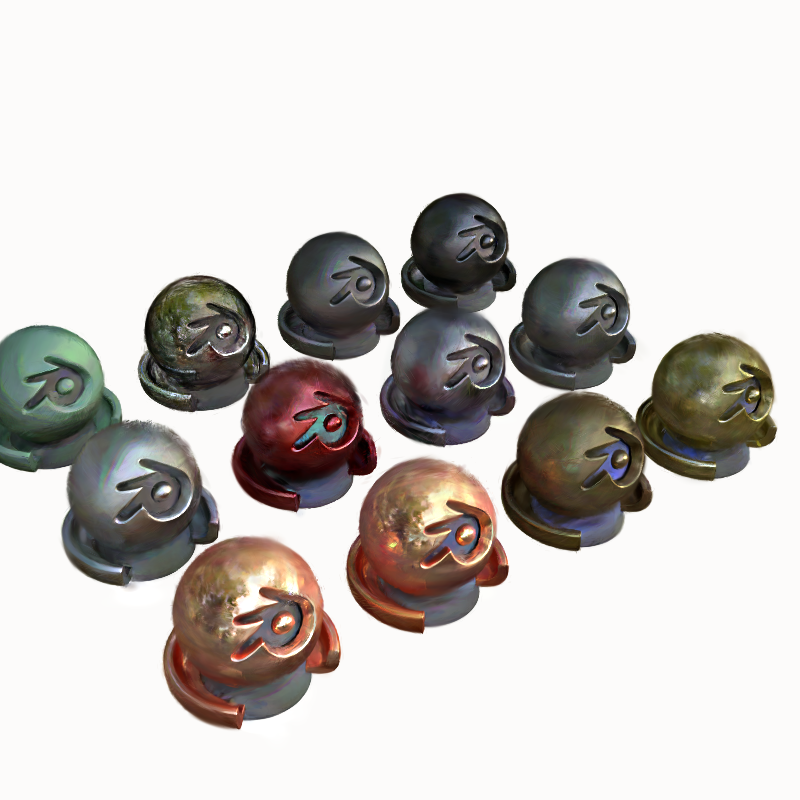}}
		& \gframe{\includegraphics[trim={0 0 0 5cm},clip,width=\linewidth]{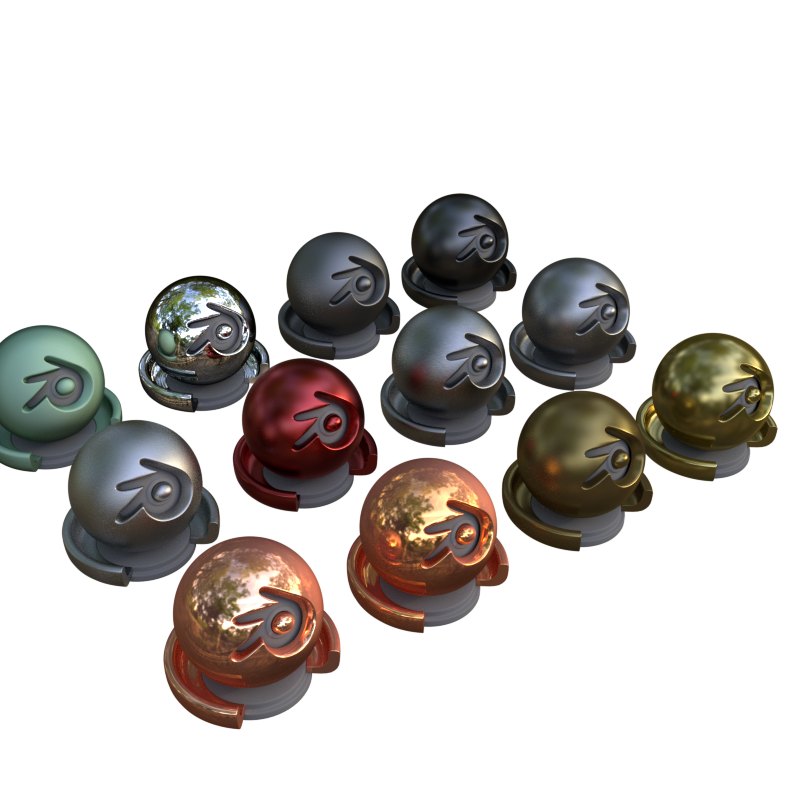}}
		\\

		& (a) E2VID + 3DGS
		& (b) Robust E-NeRF \cite{Low23iccv}
		& (c) EventSplat \cite{Yura25cvpr}
		& (d) Ours
		& (e) GT
		\\
	\end{tabular}
	}
    \vspace{-1ex}
	\caption{\emph{Qualitative results on the color synthetic dataset \cite{Low23iccv}}.}
    \label{fig:result:qualitativeSim}
\end{figure*}

\begin{table*}[t]
\centering
\adjustbox{max width=.95\linewidth}{%
\setlength{\tabcolsep}{4pt}
\begin{tabular}{ll *{8}{S[round-mode=places,round-precision=3]}}
\toprule
\textbf{Metric} & \textbf{Method}  & \textbf{\text{Avg.}} & \text{Chair}  & \text{Drums} & \text{Ficus} & \text{Hotdog} & \text{Lego} & \text{Materials} & \text{Mic} \\
\midrule
\multirow{4}{*}{PSNR $\uparrow$} & 
E2VID + 3DGS & 19.29 & 21.39 & 19.86 & 19.90 & 15.55 & 18.17 & 20.08 & 20.10  \\
& Robust E-NeRF (ICCV'23) \cite{Low23iccv} & 26.7657 & \bnum{30.24} & 23.15 & \bnum{30.71} & 18.07 & 27.34 & 24.98 & 32.87 \\
& EventSplat (CVPR'25) \cite{Yura25cvpr} & \bnum{28.14} & 28.69 & \bnum{25.81} & 29.90 & \bnum{22.91} & \bnum{29.22} & \bnum{27.16} & \bnum{33.27} \\
& Ours  & 23.11  & 26.42 & 23.34 & 25.36 & 17.76 & 18.08 & 23.50 & 27.30 \\
\midrule
\multirow{4}{*}{SSIM $\uparrow$} & 
E2VID + 3DGS & 0.917 & 0.934 & 0.915 & 0.922 & 0.897 &  0.895 & 0.901 & 0.957 \\
& Robust E-NeRF (ICCV'23) \cite{Low23iccv}  & 0.945 & \bnum{0.958} & 0.897 & \bnum{0.971} & \bnum{0.953} & 0.934 & 0.923 & 0.981 \\
& EventSplat (CVPR'25) \cite{Yura25cvpr}  & \bnum{0.953} & 0.953 & \bnum{0.947} & 0.966 & 0.940 & \bnum{0.945} & \bnum{0.936} & \bnum{0.986} \\
& Ours & 0.927 & 0.941 & 0.921 & 0.938 & 0.911 & 0.901 & 0.910 & 0.968  \\
\midrule
\multirow{4}{*}{LPIPS $\downarrow$} & 
E2VID + 3DGS  & 0.118 & 0.076 & 0.094 & 0.108 & 0.208 & 0.145 & 0.125 & 0.069 \\
& Robust E-NeRF (ICCV'23) \cite{Low23iccv} & 0.057 & \bnum{0.040} & 0.091 & \bnum{0.022} & \bnum{0.095} & 0.074 & \bnum{0.052} & 0.029 \\
& EventSplat (CVPR'25) \cite{Yura25cvpr} & \bnum{0.051} & 0.047 & \bnum{0.052} & 0.028 & 0.098 & \bnum{0.055} & 0.060 & \bnum{0.015}  \\
& Ours & 0.0737 & 0.054 & 0.066 & 0.046 & 0.160 & 0.097 & 0.061 & 0.032 \\
\bottomrule
\end{tabular}
}
\vspace{-1ex}
\caption{\emph{Quantitative results on the color synthetic dataset \cite{Low23iccv}}. 
The Bayer pattern is challenging for the proposed warp-based method.}
\label{tab:result:robustenerf}
\end{table*}

\subsection{Results on Real-World Datasets}
\label{sec:experim:real}

\Cref{fig:result:qualitativeReal} shows the results on EDS and TUM-VIE datasets.
Throughout the scenes, the proposed method consistently achieves successful reconstructions (we encourage readers to watch the video).
Notably, our reconstructions recover fine details:
($i$) gradual (mild) intensity changes, e.g., shadows and reflections on the desk in \emph{TUM-desk2},
($ii$) fewer artifacts due to noisy events, e.g., walls on \emph{EDS-07,11}.
($iii$) sharp edges in details, e.g., airplane and background in \emph{EDS-13}. 
Also, EDS sequences contain lots of events due to flickering lights.
Surprisingly, our method converges and successfully reconstructs the scene while relying on the contrast loss, which may be sensitive to flickering events.

The quantitative comparison is reported in \cref{tab:result:eds}.
Our method consistently achieves state-of-the-art results: \emph{the best results on average across all three metrics},
despite not relying on pretrained depth estimation models \cite{Huang25cvpr}, or video-guided initialization and cubic splines for pose interpolation \cite{Yura25cvpr}.
Notice that there are some limitations of the quantitative evaluation on the real-world sequences, such as the high dynamic range (HDR) of event cameras, and the disparity between the event camera and the frame camera.

\subsection{Results on Synthetic Data}
\label{sec:experim:sim}

Due to the influence of synthetic RGB-based novel-view-synthesis datasets \cite{Mildenhall21nerf}, the method is also tested on such data, converted into events via an event camera simulator \cite{Rebecq18corl}.
Note that such sequences are unrealistic because they lack the noise and most dynamic effects characteristic of event cameras.
Results are shown in \cref{fig:result:qualitativeSim}.
The RGB Bayer pattern is challenging for warp-based methods, such as the proposed one, since ($i$) warped pixels may not fall into the same location among different colors \cite{Zhang22pami}, which complicates the demosaicing operation, 
and ($ii$) the color distribution is imbalanced (green pixels are twice as many as red/blue). 
Nonetheless, our method achieves successful color reconstruction.
Following the same color correction steps as \cite{Low23iccv,Yura25cvpr}, our results show fewer object artifacts and fewer floaters on the background.
Quantitative results are given in \cref{tab:result:robustenerf}, where we achieve competitive values.

\Cref{fig:depthpoint} shows rendered depth (sparse or dense) obtained on this data and real-world data. 
We find that Gaussian-based depth estimation achieves high-quality results, especially around occlusions.
We report the  quantitative comparisons with  EMVS \cite{Rebecq18ijcv} %
in the supplementary.

\subsection{Runtime}
\label{sec:runtime}

The training takes $30$--$45$ minutes for EDS and synthetic sequences \cite{Low23iccv},
and $80$--$130$ minutes for TUM-VIE.
The rendering takes roughly $3$~ms for $\numGaussians = 0.1M$, and $30$~ms for $\numGaussians = 1M$, using a PyTorch implementation on an NVIDIA RTX6000 (Ada).
Our method is significantly faster than other methods:
both Robust E-NeRF \cite{Low23iccv} and IncEventGs \cite{Huang25cvpr} take $3$ h to train on EDS under the same settings.
EventSplat \cite{Yura25cvpr} does not have publicly available code but reports $1$--$3$ h for the same number of iterations on EDS.

\section{Ablations}
\label{sec:ablation}

\subsection{Effect of Temporal Window Selection}
\label{sec:experim:renderonce}

\begin{figure}[ht!]
  \centering
  {{\includegraphics[trim={0 5.5cm 6.6cm 0},width=\linewidth]{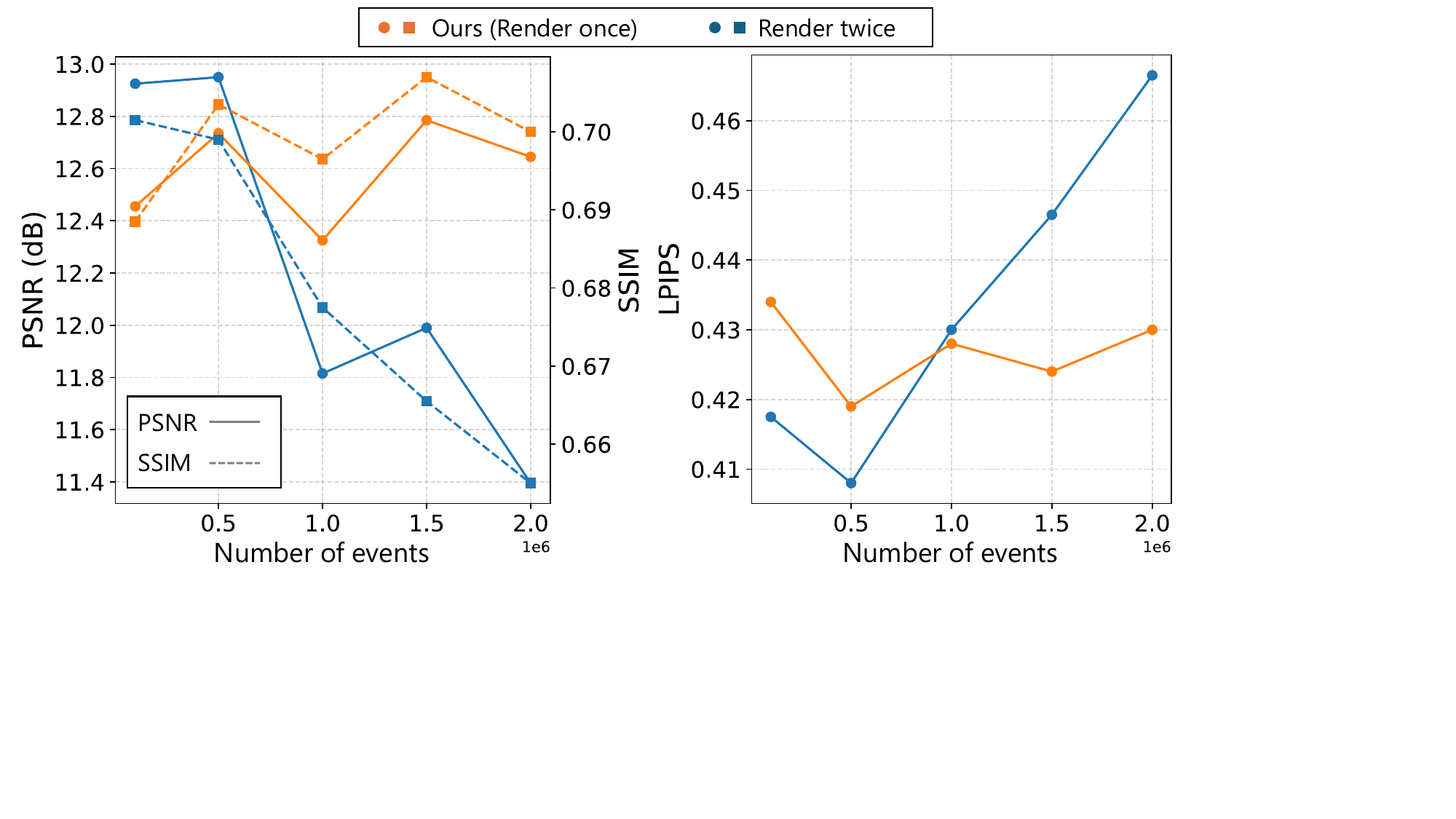}}}\\
  \vspace{-1ex}
\caption{\emph{Robustness with respect to the time window selection}. 
We compare the proposed pipeline and its render-twice variant for different numbers of events $\numEvents$.
Due to the warp that reduces blur in the scene, the proposed method shows robustness against the choice of $\numEvents$, achieving consistently good values.}
\label{fig:renderonce}
\end{figure}

We further investigate the efficacy of our two-branch pipeline, which renders intensity just once (\cref{fig:method}).
Most event-based GS methods render dense intensity twice and subtract one from another to obtain an edge-like image (i.e., $\Delta \raycolor = \raycolor(t_2) - \raycolor(t_1)$) 
that is compared to the brightness increment obtained by pixel-wise accumulation of the event data.
A clear advantage of the proposed pipeline over the above ``render-twice'' pipeline is the robustness with respect to the choice of $\numEvents$.
\Cref{fig:renderonce} reports reconstruction performance for different $\numEvents$,
using two sequences from the TUM-VIE dataset: \emph{1d-trans} and \emph{desk2}.
For larger $\numEvents$ the edges become more blurry in the render-twice pipeline, and therefore, the reconstruction quality degrades.
However, the proposed render-once pipeline shows consistent results regardless of $\numEvents$,
which is desirable because it is a sensible parameter that depends on many factors, such as camera resolution, scene texture, and camera motion.

\begin{table}[t]
\centering
\adjustbox{max width=\linewidth}{%
\setlength{\tabcolsep}{3pt}
\begin{tabular}{ll *{8}{S[round-mode=places,round-precision=3]}}
\toprule
\textbf{Metric} & \textbf{Method} & \text{Synthetic \cite{Low23iccv}}  & \text{EDS} & \text{TUM-VIE} \\
\midrule
\multirow{3}{*}{PSNR $\uparrow$} & 
Ours & \bnum{23.11} & \bnum{19.47} &  13.09 \\
& w/o contrast loss & 9.60 & 15.52 & \bnum{13.45} \\
& w/o initialization & 20.82 & 17.34 & 11.36\\
\midrule
\multirow{3}{*}{SSIM $\uparrow$} & 
Ours & \bnum{0.927} & \bnum{0.816}  & \bnum{0.716} \\
& w/o contrast loss & 0.81 & 0.7439 & 0.715 \\
& w/o initialization & 0.908 & 0.759 & 0.633\\
\midrule
\multirow{3}{*}{LPIPS $\downarrow$} & 
Ours & \bnum{0.073} & \bnum{0.357} & 0.411 \\
& w/o contrast loss & 0.405 & 0.5808 & \bnum{0.405} \\
& w/o initialization & 0.121 & 0.442 & 0.561\\
\bottomrule
\end{tabular}
}
\vspace{-1ex}
\caption{\emph{Ablation on the contrast loss and the initialization}.}
\label{tab:result:ablation}
\vspace{-2ex}
\end{table}

\subsection{Contrast Loss and Initialization}
\label{sec:ablation:contrast}

We conduct ablation studies on the contrast loss and initialization, as shown in \cref{tab:result:ablation}.
Here, ``w/o initialization'' starts from random $10^5$ Gaussians and skips the proposed initialization step (\cref{sec:method:initialization}).
Our method achieves the best or second-best results among all metrics and datasets.
Notably, the SSIM improves with the contrast loss, showcasing the efficacy of the proposed ray-tracing rendering and loss.

\section{Limitations}
\label{sec:limitations}

Our method follows an unsupervised approach based on the contrast loss, which assumes brightness constancy and therefore suffers in the presence of flickering events.
Although the pipeline converges on the EDS dataset, we find that the presence of large amounts of flickering events make appearance recovery and depth estimation results unstable.

The proposed framework assumes static scenes, and is not expected to work well on dynamic scenes.
However, following recent advances in frame-based 4D GS \cite{Wu24cvpr,Yin25siggraph},
event-based 4D GS would be a relevant future direction. %

\vspace{-.5ex}
\section{Conclusion}
\label{sec:conclusion}

We propose the first framework for event-based Gaussian Splatting that fully leverages the spatio-temporal properties of event data. 
Our method targets the majority of modern event cameras, which are monochrome (i.e., grayscale) and with VGA or higher (1 megapixel) resolution.
The rendering pipeline consists of two explicit pathways:
spatially sparse and temporally dense (i.e., event-by-event) pathway for geometry (depth) recovery, 
and spatially dense and temporally sparse (i.e., a snapshot) pathway for appearance (radiance) estimation.
A thorough evaluation reveals that the proposed method
($i$) achieves state-of-the-art performance on real-world data without using extra priors, 
and ($ii$) effectively tackles the trade-off revolved around the choice of the number of events to process.

\section*{Acknowledgments}
We thank Mr. Yura Toshiya and Dr. Hidenobu Matsuki for useful discussions.
This work was partially supported by JST-Research and Development Program for Next-generation Edge AI Semiconductors Grant Number JPMJES2513.
This work has been partially supported by the German Federal Ministry of Research, Technology and Space (BMFTR) under the Robotics Institute Germany (RIG).
Funded by the Deutsche Forschungsgemeinschaft (DFG, German Research Foundation) under Germany’s Excellence Strategy – EXC 2002/1 ``Science of Intelligence'' – project number 390523135.

\ifarxiv

\section{Supplementary}
\label{sec:suppl}

In the supplementary, we first 
provide additional details about the motion field \cref{sec:suppl:motionfield}.  
Then, we report the sensitivity analysis of the loss weights (\cref{sec:suppl:sensitivity}) 
and the detailed analysis of the runtime (\cref{sec:suppl:runtime}).
We also provide details on the proposed initialization of the scene Gaussians \cref{sec:suppl:initialization} 
and compare it to existing work on event-based structure-from-motion (SfM) (\cref{sec:suppl:emvs}).
An example of a failure case due to flickering lights is given in \cref{sec:suppl:failurecases}.
Finally, additional results are provided on dense/sparse depth, flow, and rendered intensity (\cref{sec:suppl:moreResults}).

\subsection{Video}
We encourage readers to inspect the attached video,
which summarizes the method and the results.

\subsection{Motion Field from Depth and Camera Pose}
\label{sec:suppl:motionfield}
Let us provide further details on the geometric equation \cref{eq:motionField}. 
Assuming a stationary scene viewed by a moving camera with linear and angular velocities $\linvel$ and $\angvel$, respectively, 
the scene depth $\depth$ can be used to compute the apparent motion on the image plane via the well-known motion field equation 
\begin{equation}
\label{eq:motionField:suppl}
\velflow(\bx) = \frac1{\depth(\bx)}A(\bx)\linvel + B(\bx)\angvel,
\end{equation}
where the quantities are assumed to be given at time $t$ (and omitted, for simplicity of this instantaneous equation).
The $2\times 3$ matrices $A(\bx)$ and $B(\bx)$ solely depend on the pixel coordinates $\bx=(x,y)^\top$, given the focal length $f$:
\begin{equation}
\label{eq:motionfield:A}
A(\bx) = \begin{bmatrix}
        -f & 0  & x\\
        0  & -f & y
  \end{bmatrix},
\end{equation}  
\begin{equation}
\label{eq:motionfield:B}
B(\bx) = \frac{1}{f}
        \begin{bmatrix}
        x y & -(f^2+x^2) &  f \, y \\
        f^2+y^2 & -x y   & -f \, x
  \end{bmatrix},  
\end{equation}
An alternative way to write \cref{eq:motionField:suppl} is as the product of a $2\times 6$ matrix (called feature sensitivity matrix, interaction matrix, or image Jacobian matrix for a
point feature \cite{Bryner19icra}) and the $6\times 1$ twist given by the camera's generalized velocity $(\linvel^\top,\angvel^\top)^\top$.

\subsection{Sensitivity Analysis}
\label{sec:suppl:sensitivity}

\Cref{tab:result:sensitivity} reports the sensitivity analysis regarding the loss weights: $\loss_c, \loss_p$, using the EDS dataset.
The metrics are averaged over all five sequences.
The results confirm the efficacy of all proposed loss terms in leading to a successful convergence of the GS model.

Notably, we find that \emph{event collapse} (e.g., \cite{Shiba22aisy}) occurs with a large weight of the Contrast loss $\loss_c$.
The collapse is observed in \cref{fig:collapse}c) as corrupted depth (many small Gaussians with various distances).
In \cref{fig:collapse}b) IWE, the lamp on the desk shows undesired local optima of the warp.
\begin{table}[ht]
\centering
\adjustbox{max width=\linewidth}{%
\setlength{\tabcolsep}{3pt}
\begin{tabular}{ll*{6}{S[round-mode=places,round-precision=3]}}
\toprule
$\lambda_c$ & $\lambda_s$ & \textbf{PSNR $\uparrow$} & \textbf{SSIM $\uparrow$} & \textbf{LPIPS $\downarrow$} \\
\midrule
0.1 & 0.1 & 18.994 & 0.801 & 0.389 \\
0.1 & 1 & \bnum{19.584} & \bnum{0.812} & \bnum{0.359} \\
0.1 & 10 & 17.282 & 0.773 & 0.423 \\
1 & 0.1 & 18.432 & 0.790 & 0.398 \\ 
1 & 1 & 19.094 & 0.805 & 0.361 \\ 
1 & 10 & 18.593 & 0.802 & 0.359 \\ 
10 & 0.1 & 16.666 & 0.753 & 0.448 \\
10 & 1 & 16.288 & 0.752 & 0.436 \\
10 & 10 & 16.946 & 0.770 & 0.398 \\
\bottomrule
\end{tabular}
}
\caption{\emph{Sensitivity analysis of the loss weights}.} 
\label{tab:result:sensitivity}
\end{table}

\def\figWidth{0.49\linewidth}
\begin{figure}[ht]
	\centering
    {\footnotesize
    \setlength{\tabcolsep}{1pt}
	\begin{tabular}{
	>{\centering\arraybackslash}m{\figWidth} 
	>{\centering\arraybackslash}m{\figWidth}}
 
        {\gframe{\includegraphics[width=\linewidth]{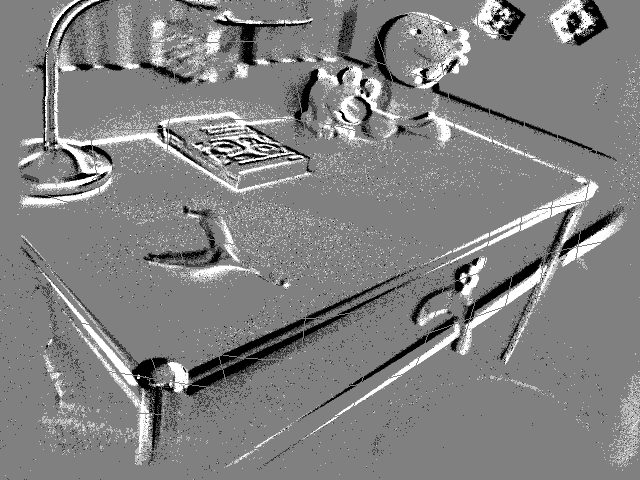}}}
        & {\gframe{\includegraphics[trim={0 0 8.5cm 0},width=\linewidth]{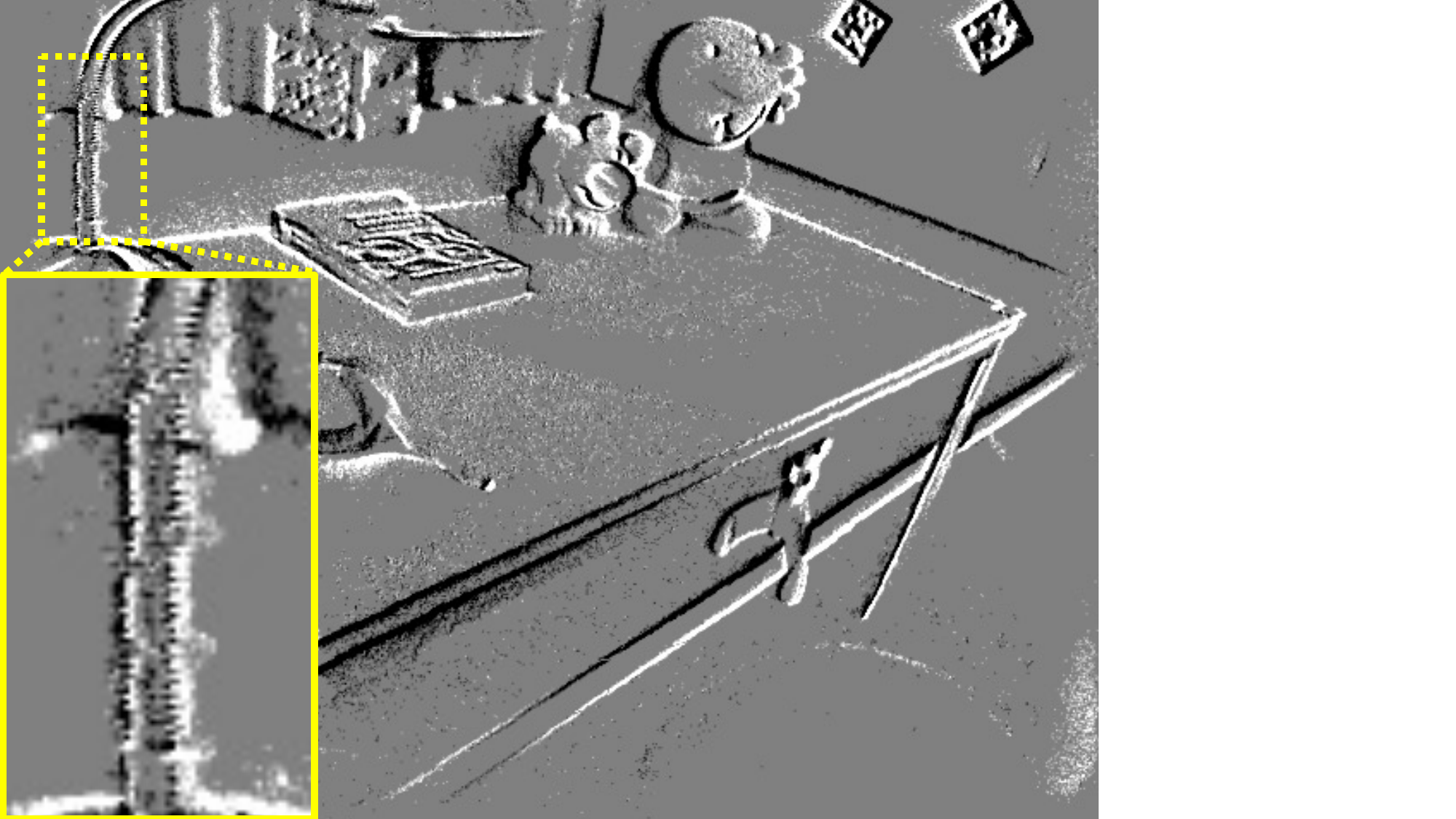}}}\\

        (a) Raw events
		& (b) IWE (collapsed)
		\\[1ex]
   
        {\gframe{\includegraphics[trim={0 0 0 2.5cm},clip,width=\linewidth]{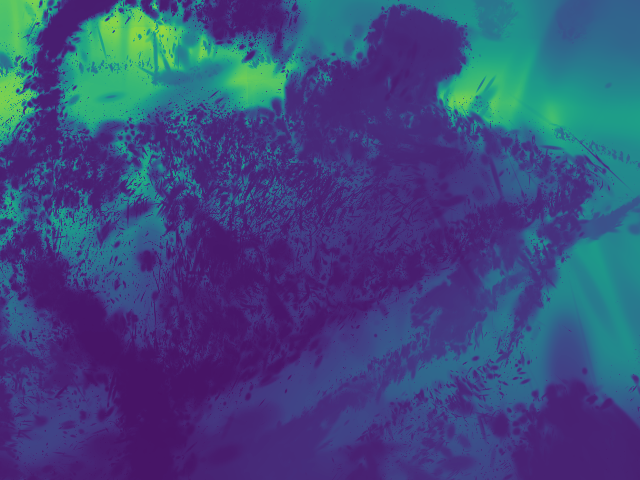}}}
        & {\gframe{\includegraphics[trim={0 0 0 2.5cm},clip,width=\linewidth]{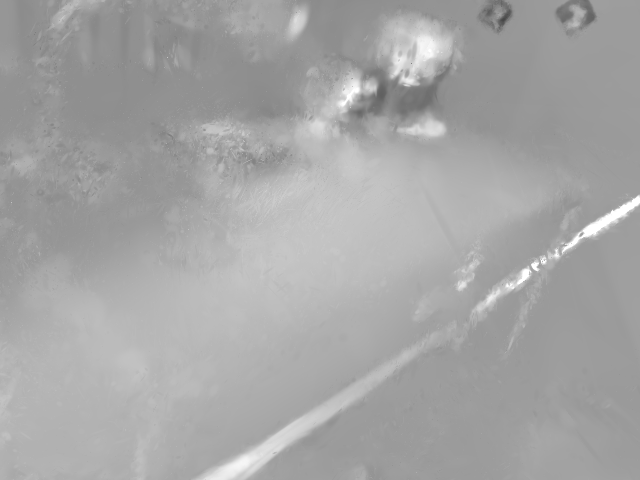}}}\\

        (c) Depth (corrupted)
		& (d) Rendered intensity
		\\

	\end{tabular}
	}
    \caption{\emph{Examples of corrupted depth for large $\loss_c$}.}
    \label{fig:collapse}
\end{figure}

\subsection{Runtime per Each Training Step}
\label{sec:suppl:runtime}

In \cref{sec:runtime}, we report the training runtime for the total steps to converge.
Here, we provide a detailed runtime analysis of each step in \cref{fig:runtime:ngaussian}.
Larger scenes have more Gaussians (i.e., larger $\numGaussians$).
The runtime of the proposed method scales sub-linearly with the scene size, despite having the warp (i.e., $O(\numEvents)$) and IWE (i.e., $O(\numEvents + \numPixels)$) creation steps.
For reference, we also report the render-twice variant of the proposed pipeline.
Our pipeline is slightly faster; however, we do not observe any significant differences.
\begin{figure}[t]
  \centering
  {{\includegraphics[width=0.8\linewidth]{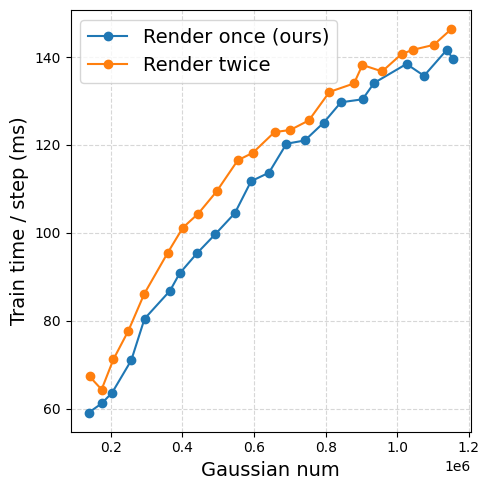}}}\\
\caption{\emph{Detailed analysis on the runtime for different number of Gaussians $\numGaussians$ used to model the scene}.}
\label{fig:runtime:ngaussian}
\end{figure}

\subsection{Initialization}
\label{sec:suppl:initialization}

Our method starts from a random distribution of $100$k Gaussians.
During the initial steps ($10$k steps out of the total $40$k steps), 
we run the system pipeline in \cref{fig:method} using $\raycolor(\bx)$, directly, instead of $\hat{H}(\bx; \tref)$ (Eq.~\eqref{eq:brightnesschange}).
After the initial $10$k steps,
the Gaussians converge, as shown in \cref{fig:rebuttal:initialization}~(a).
The motivation of the initial steps is to favor initial Gaussians on scene texture and edges,
and we find that IWEs produce better initialization than images of pixel-wise accumulation of events (e.g., \cref{fig:rebuttal:initialization}~(b)) because of their sharpness, which more concretely determines the location of the centers of the Gaussians.
\def\figWidth{0.49\linewidth}
\begin{figure}[ht]
	\centering
    {\footnotesize
    \setlength{\tabcolsep}{1pt}
	\begin{tabular}{
	>{\centering\arraybackslash}m{\figWidth} 
	>{\centering\arraybackslash}m{\figWidth}}

        {\gframe{\includegraphics[width=\linewidth]{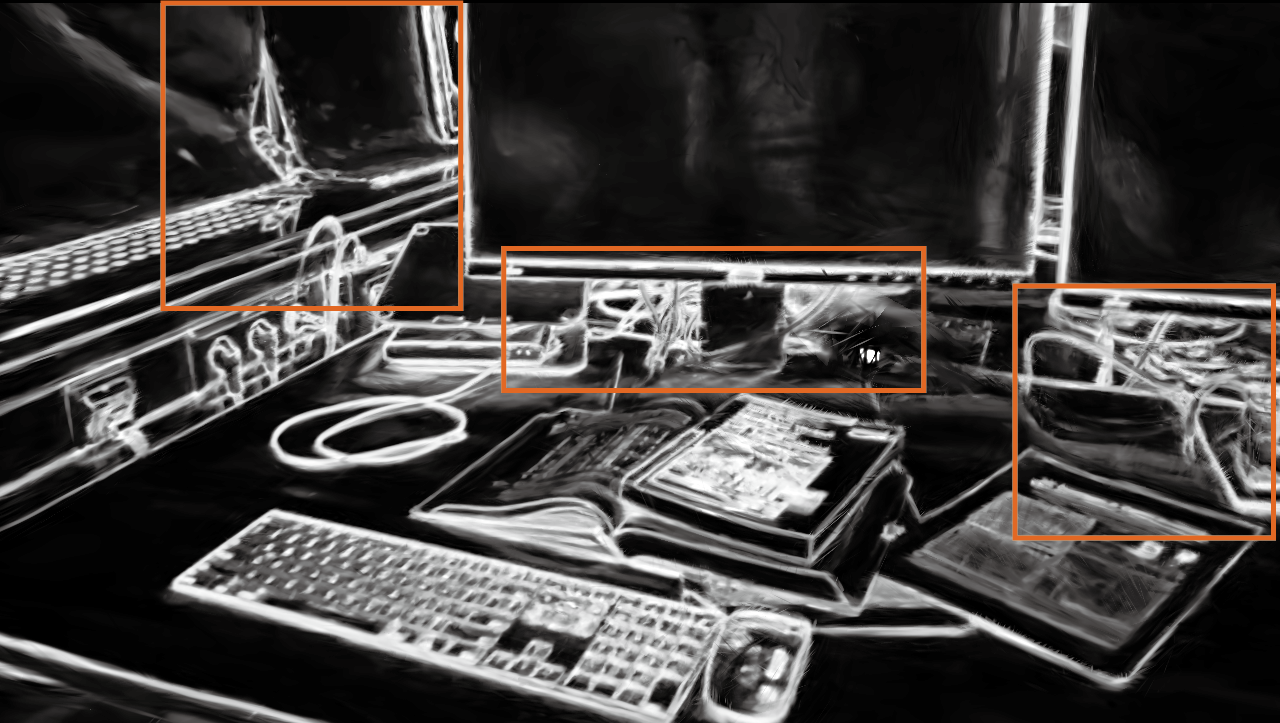}}}
        & {\gframe{\includegraphics[width=\linewidth]{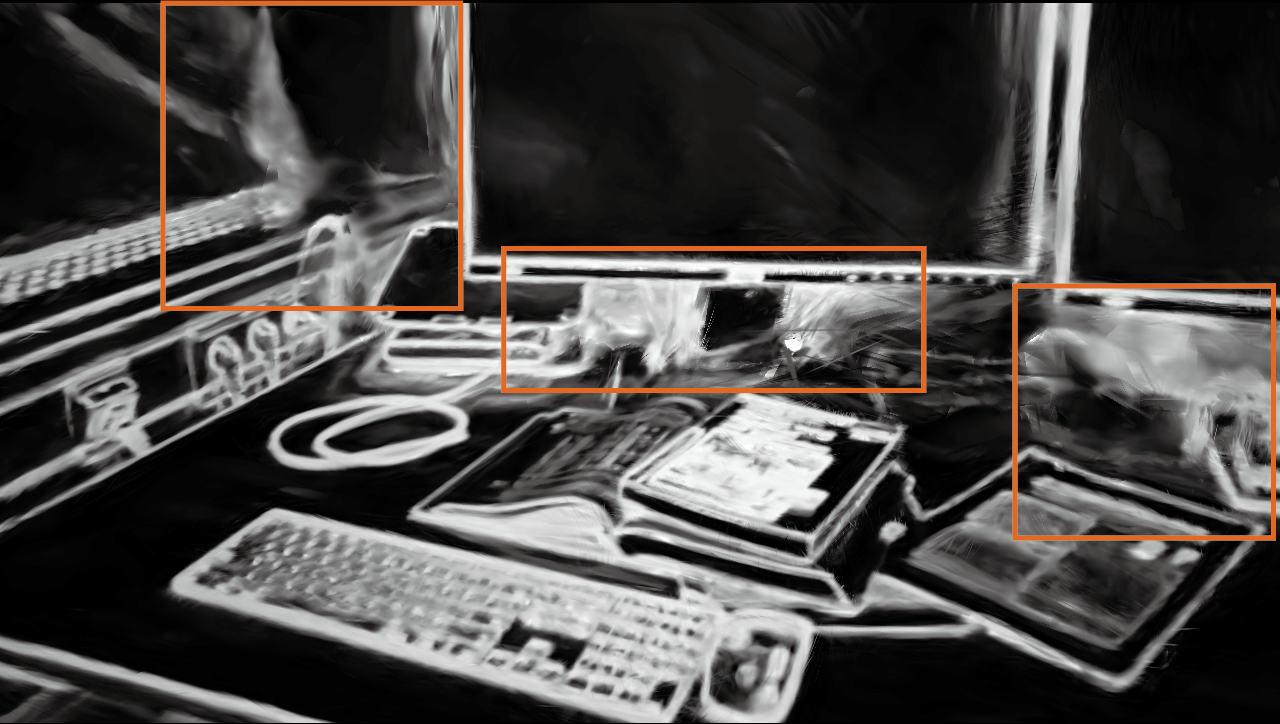}}} \\
        
        (a) $\raycolor(\bx)$ optimized by IWE
        & (b) $\raycolor(\bx)$ optimized by event frame
		\\
   
    \end{tabular}
	}
    \caption{IWE-based initialization (a) reconstructs better structure than event-frame--based initialization (b). 
    See the region enclosed by the orange rectangles.}
    \label{fig:rebuttal:initialization}
\end{figure}

\subsection{Comparison with SfM Methods}
\label{sec:suppl:emvs}

As discussed in \cref{sec:method:initialization,sec:suppl:initialization}, the proposed framework initializes the scene geometry via optimization without polarity information,
which has a similar effect as COLMAP, i.e., SfM.
Here, using the synthetic dataset from \cite{Low23iccv}, which has accurate ground truth geometry, we now visualize and compare initial point cloud estimation results.
For the evaluation of 3D points, we use the Chamfer Distance (CD):
\begin{equation}
\begin{split}
\label{eq:chamferDistance}
    \mathbf{CD}(X,\hat{X}) = & \frac{1}{|X|}
    \sum_{x\in{X}}{\min_{\hat{x}\in{{\hat{X}}}}{\| x-\hat{x}\|^2_2}}
    \\ & + \frac{1}{|\hat{X}|}\sum_{\hat{x}\in{\hat{X}}}{\min_{x\in{X}}{\|x-\hat{x}\|^2_2}},
\end{split}
\end{equation}
which measures the 3D Euclidean distance between the predicted points $\hat{X}$ and the ground truth (GT) points $X$.

\Cref{fig:emvs} displays qualitative 3D point estimation results.
As baselines, we use the frame-based pipeline (``E2VID~\cite{Rebecq19pami} $+$ VGGT~\cite{wang25cvpr_vggt}'') and Event-based Multi-view Stereo (EMVS)~\cite{Rebecq18ijcv}. %
Our method consistently recovers fine details of the scene, such as the thin edges of the chair and drum, and the cables of the mic.
On the other hand, the event-based baseline, EMVS \cite{Rebecq18ijcv}, struggles to recover the entire scene and is limited to the points visible from a small range of viewpoints in the entire trajectory.
EMVS is not suitable for the $360$-degree trajectory that is typical for the GS and NeRF settings, since the 3D space is represented as voxels (DSIs) with the perspective projection.

Quantitative results are reported in \Cref{tab:result:point}.
Our method achieves the smallest CD among all sequences except for the \emph{hotdog} sequence.
``E2VID $+$ VGGT'' recovers the \emph{hotdog} sequence the best, possibly due to its simple shape; however, it struggles to estimate correct 3D points for other sequences.
The overall results show that the proposed initialization provides more plausible initial geometry than the conventional event-based or event-to-frame SfM methods.

\subsection{Failure Cases}
\label{sec:suppl:failurecases}

As mentioned in \cref{sec:limitations}, flickering lights are challenging for event-based methods based on brightness constancy. 
\Cref{fig:rebuttal:flicer} shows a scene from the EDS dataset where many events are generated away from object edges due to flickering lights, which produces noisy reconstructions.
\def\figWidth{0.49\linewidth}
\begin{figure}[ht]
	\centering
    {\small
    \setlength{\tabcolsep}{1pt}
	\begin{tabular}{
	>{\centering\arraybackslash}m{\figWidth} 
	>{\centering\arraybackslash}m{\figWidth}}
 
        {\gframe{\includegraphics[width=\linewidth]{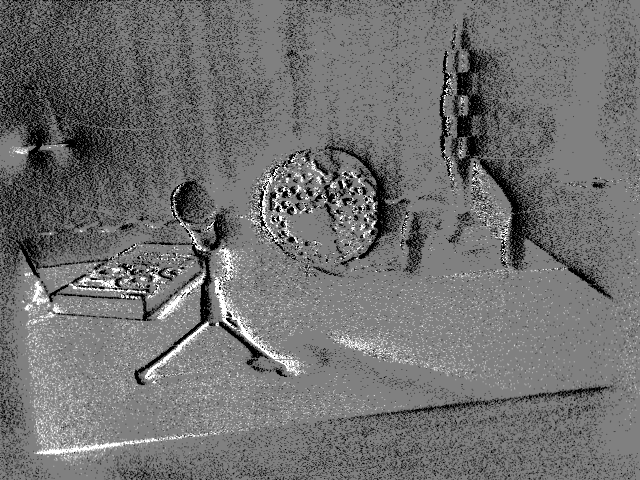}}}
        & {\gframe{\includegraphics[width=\linewidth]{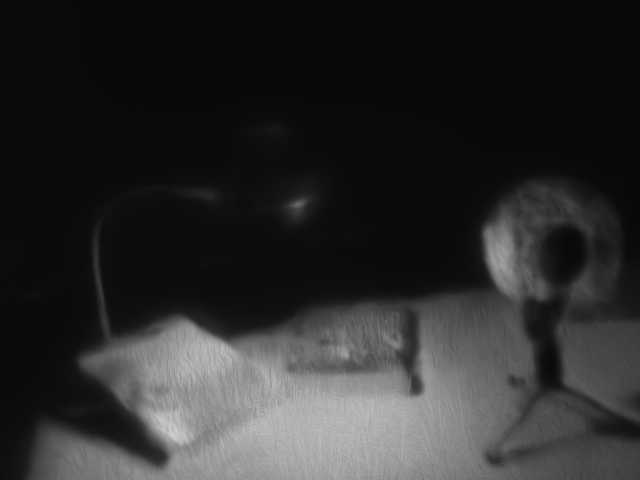}}} \\

        (a) Scene with flicker events
		& (b) Reconstruction
		\\

    \end{tabular}
	}
    \caption{Flickering events produce blurred results.}
    \label{fig:rebuttal:flicer}
\end{figure}

\begin{figure*}[t]
  \centering
  {\includegraphics[clip,trim={0cm 0.5cm 12.5cm 0cm},width=0.9\linewidth]{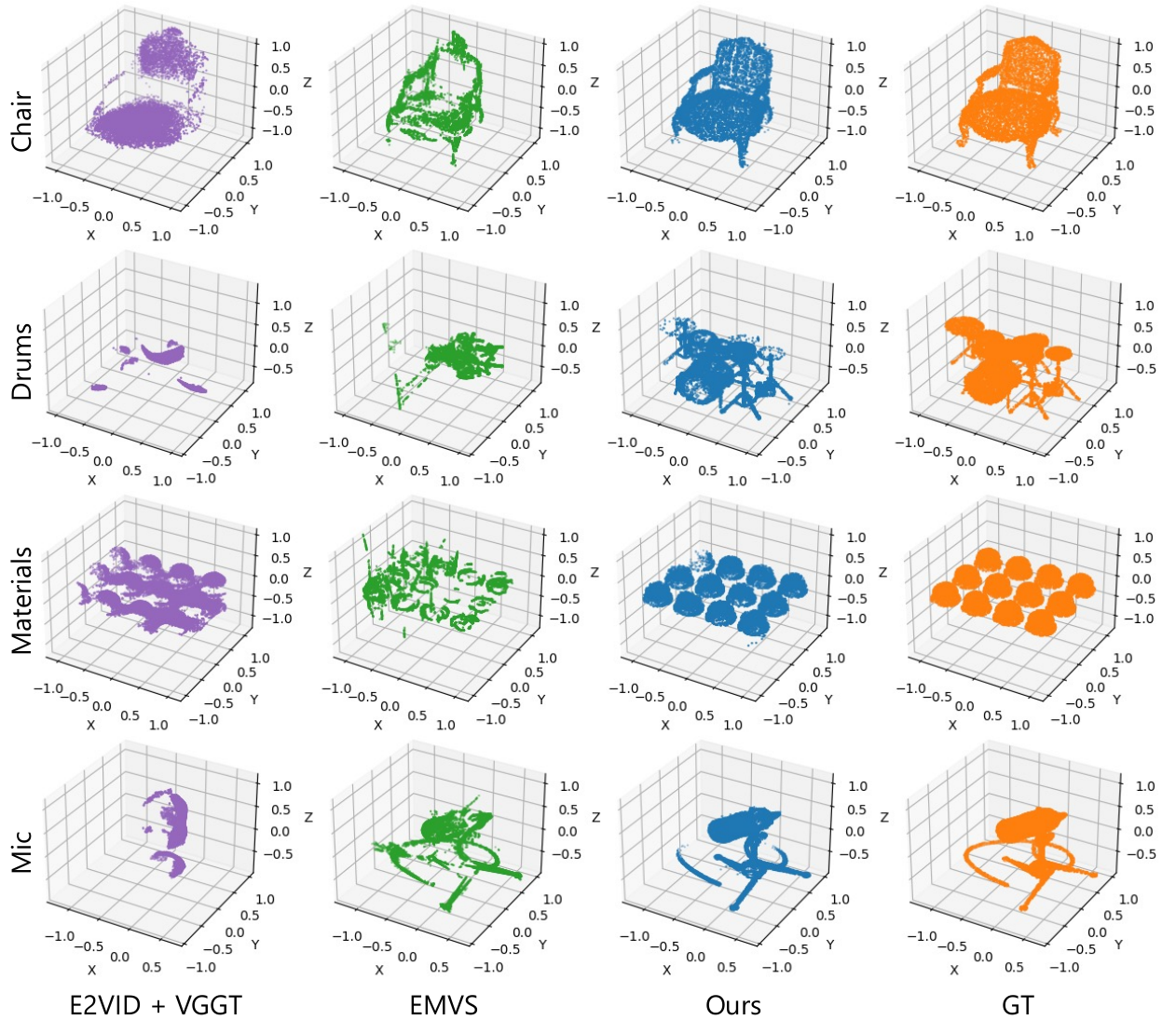}}
\caption{\emph{Results on the point cloud estimation.} 
For comparison, we use E2VID \cite{Rebecq19pami} $+$ VGGT \cite{wang25cvpr_vggt} and EMVS \cite{Rebecq18ijcv}.}
\label{fig:emvs}
\end{figure*}

\begin{table*}[t]
\centering
\adjustbox{max width=.95\linewidth}{%
\setlength{\tabcolsep}{4pt}
\begin{tabular}{l*{8}{S[round-mode=places,round-precision=3]}}
\toprule
\textbf{Method}  & \textbf{\text{CD} $\downarrow$ (Avg.)} & \text{Chair}  & \text{Drums} & \text{Ficus} & \text{Hotdog} & \text{Lego} & \text{Materials} & \text{Mic} \\
\midrule
E2VID \cite{Rebecq19pami} $+$ VGGT \cite{wang25cvpr_vggt} & 34.82 & 25.31 & 82.34 & \novalue & \bnum{5.284} & 4.568 & 11.50 & 79.92  \\
EMVS \cite{Rebecq18ijcv} & 35.09 & 9.757 & 79.33 & 7.260 & 51.53 & 71.39 & 18.89 & 7.49 \\
Ours & \bnum{3.559} & \bnum{3.127} & \bnum{1.204} & \bnum{0.949} & 11.49 & \bnum{3.056} & \bnum{1.351} & \bnum{3.734} \\
\bottomrule
\end{tabular}
}
\vspace{-1ex}
\caption{\emph{Quantitative results on point cloud estimation using data \cite{Low23iccv}}.
The CD is given in mm.
``E2VID + VGGT'' does not converge on the ficus sequence.}
\label{tab:result:point}
\end{table*}

\subsection{Further Qualitative Results}
\label{sec:suppl:moreResults}

\Cref{fig:suppl:moreRes} shows further results on depth, flow, and intensity reconstruction using the three datasets.
\def\figWidth{0.163\linewidth}
\begin{figure*}[t]
	\centering
    {\footnotesize
    \setlength{\tabcolsep}{1pt}
	\begin{tabular}{
	>{\centering\arraybackslash}m{\figWidth} 
	>{\centering\arraybackslash}m{\figWidth} 
	>{\centering\arraybackslash}m{\figWidth} 
	>{\centering\arraybackslash}m{\figWidth} 
	>{\centering\arraybackslash}m{\figWidth} 
	>{\centering\arraybackslash}m{\figWidth}}
 
        {\gframe{\includegraphics[width=\linewidth]{images/depthflow/eds08/eds08_dense_depth_910.png}}}
        & {\gframe{\includegraphics[width=\linewidth]{images/depthflow/eds08/eds08_sparse_depth_910.png}}}
        & {\gframe{\includegraphics[width=\linewidth]{images/depthflow/eds08/eds08_dense_flow_910.png}}}
        & {\gframe{\includegraphics[width=\linewidth]{images/depthflow/eds08/eds08_sparse_flow_910.png}}}
        & {\gframe{\includegraphics[width=\linewidth]{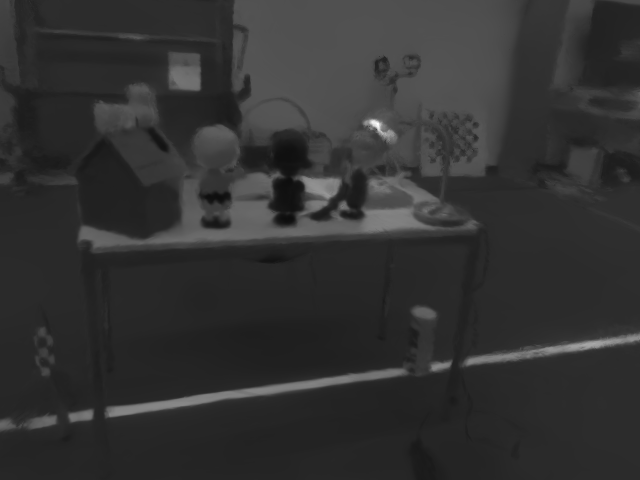}}}
        & {\gframe{\includegraphics[width=\linewidth]{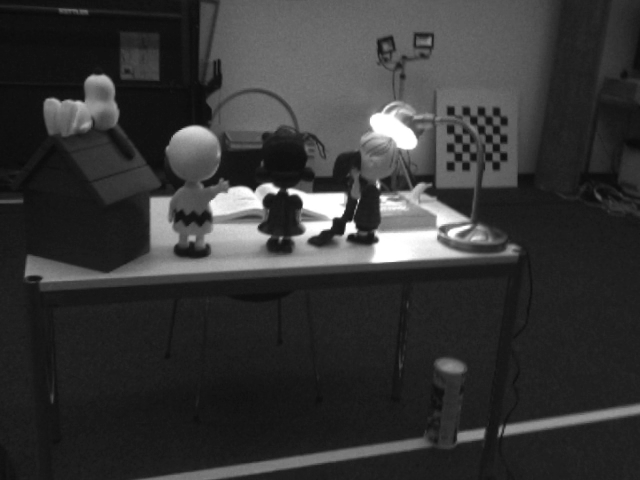}}}\\

        {\gframe{\includegraphics[width=\linewidth]{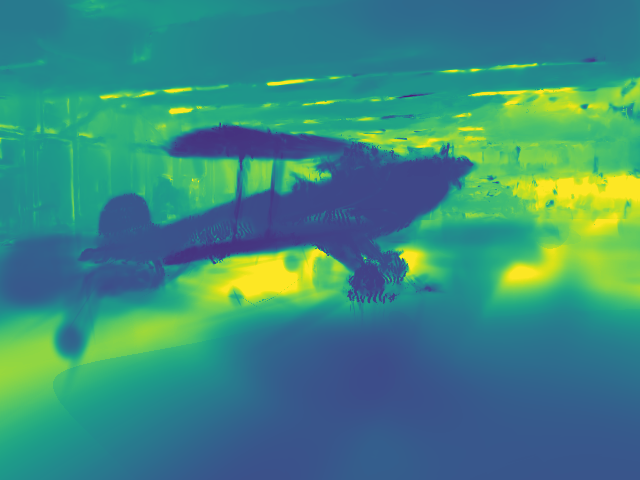}}}
        & {\gframe{\includegraphics[width=\linewidth]{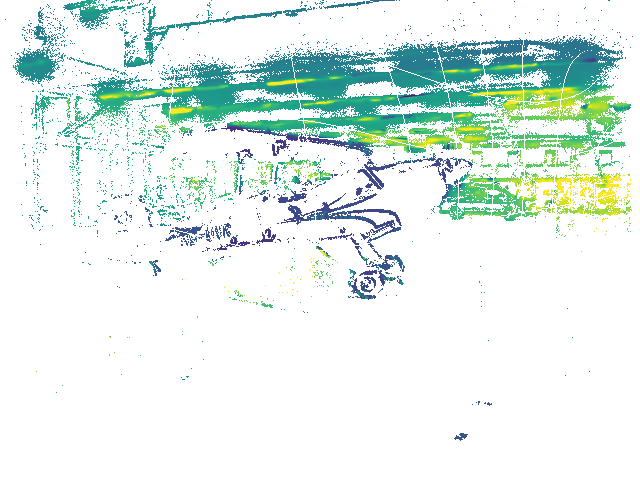}}}
        & {\gframe{\includegraphics[width=\linewidth]{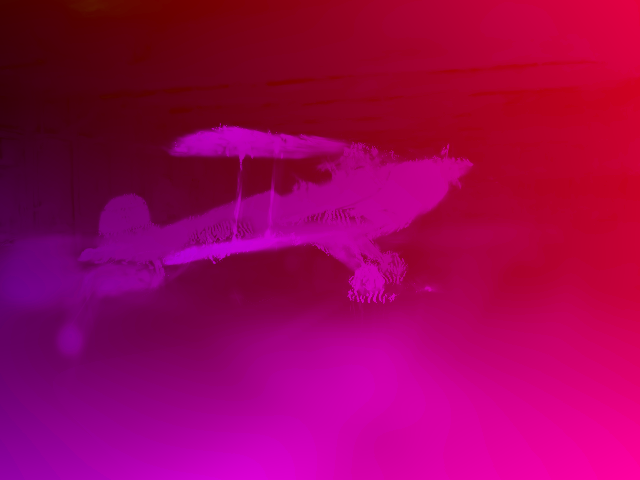}}}
        & {\gframe{\includegraphics[width=\linewidth]{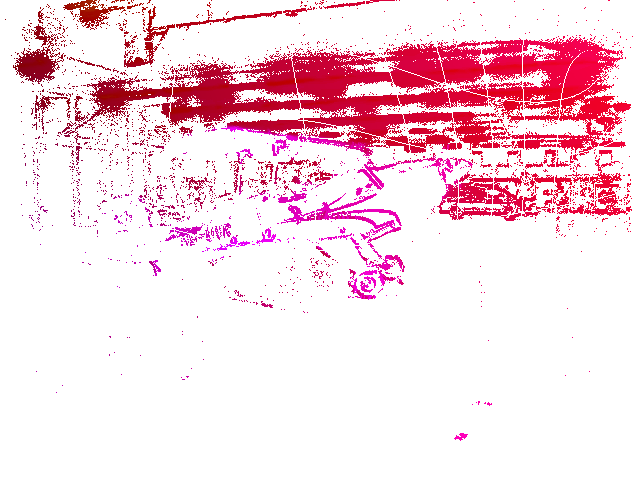}}}
        & {\gframe{\includegraphics[width=\linewidth]{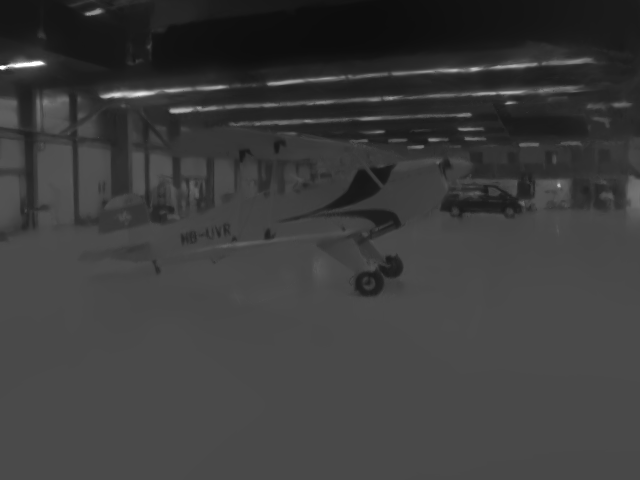}}}
        & {\gframe{\includegraphics[width=\linewidth]{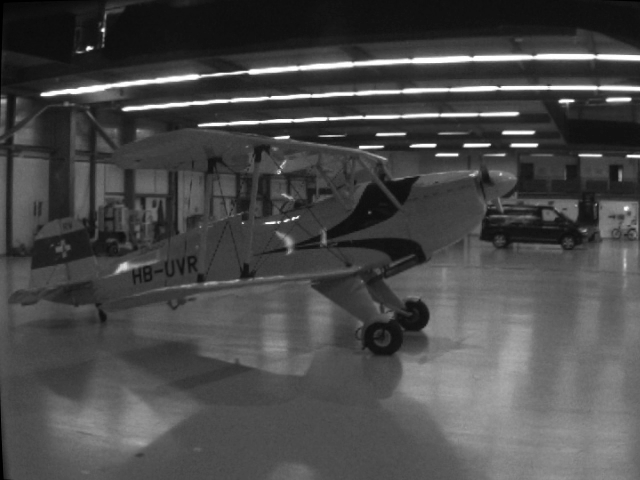}}}\\

        {\gframe{\includegraphics[trim={0 0 0 2.5cm},clip,width=\linewidth]{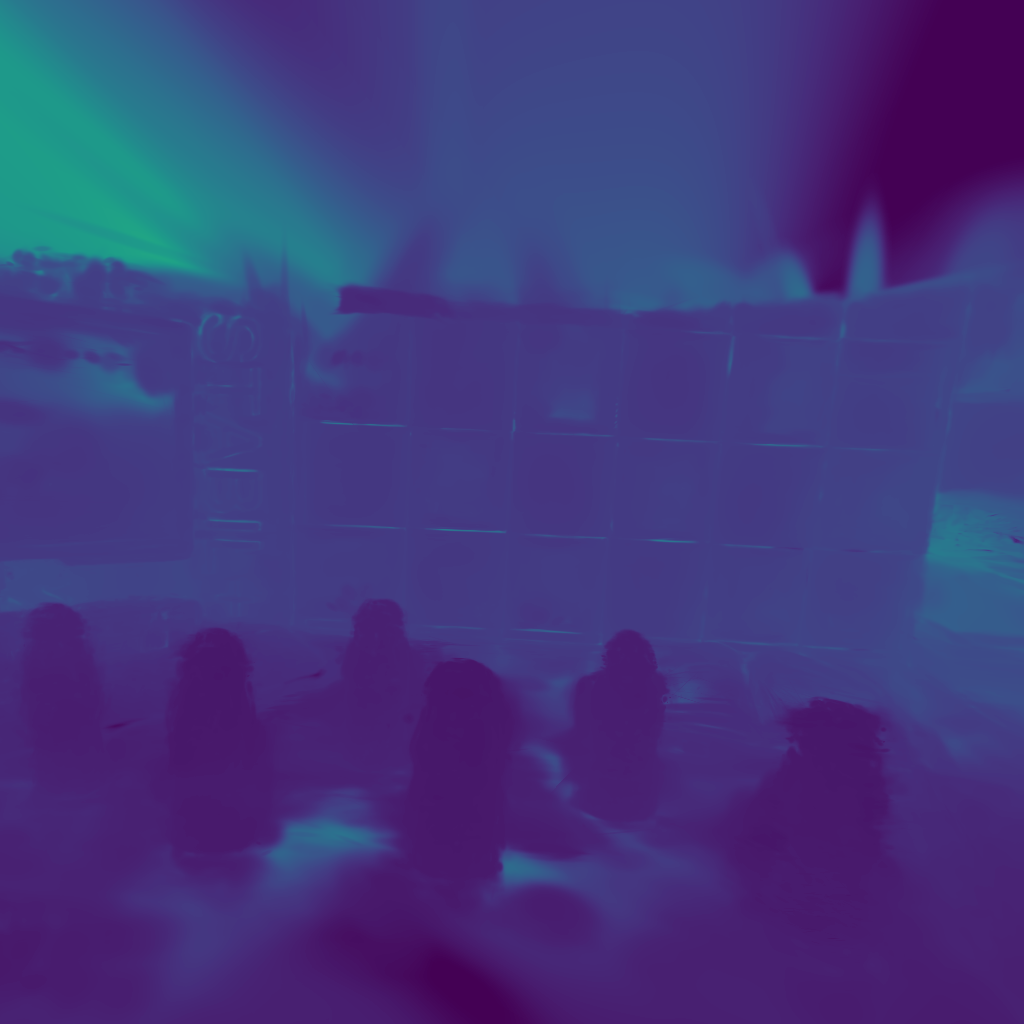}}}
        & {\gframe{\includegraphics[trim={0 0 0 2.5cm},clip,width=\linewidth]{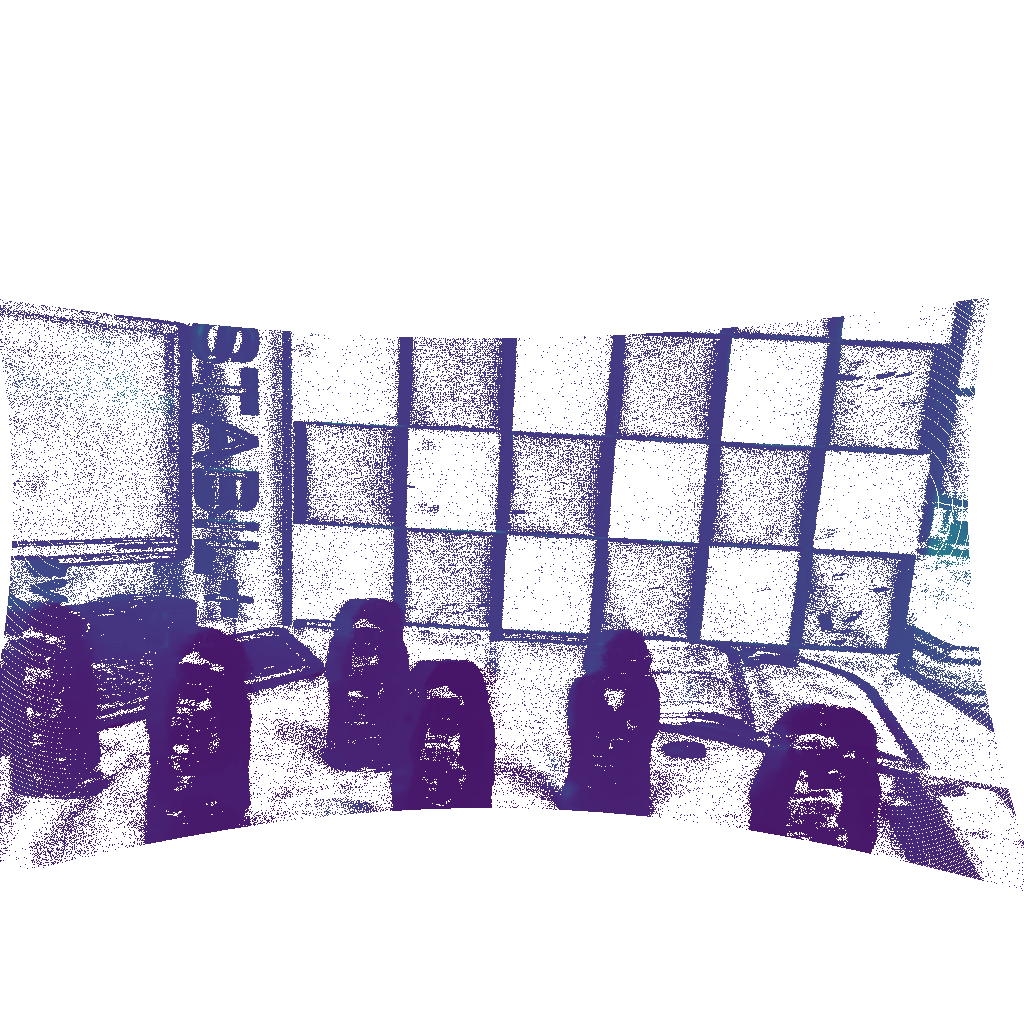}}}
        & {\gframe{\includegraphics[trim={0 0 0 2.5cm},clip,width=\linewidth]{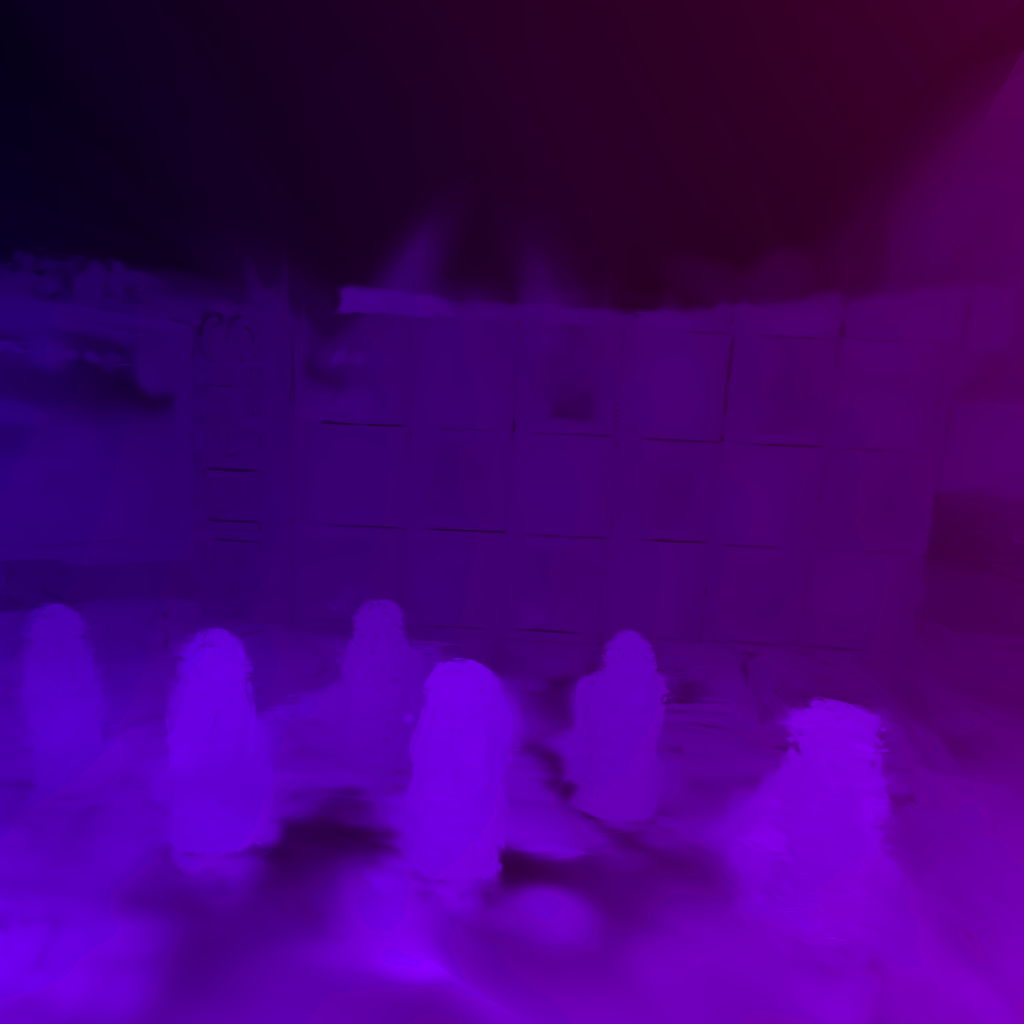}}}
        & {\gframe{\includegraphics[trim={0 0 0 2.5cm},clip,width=\linewidth]{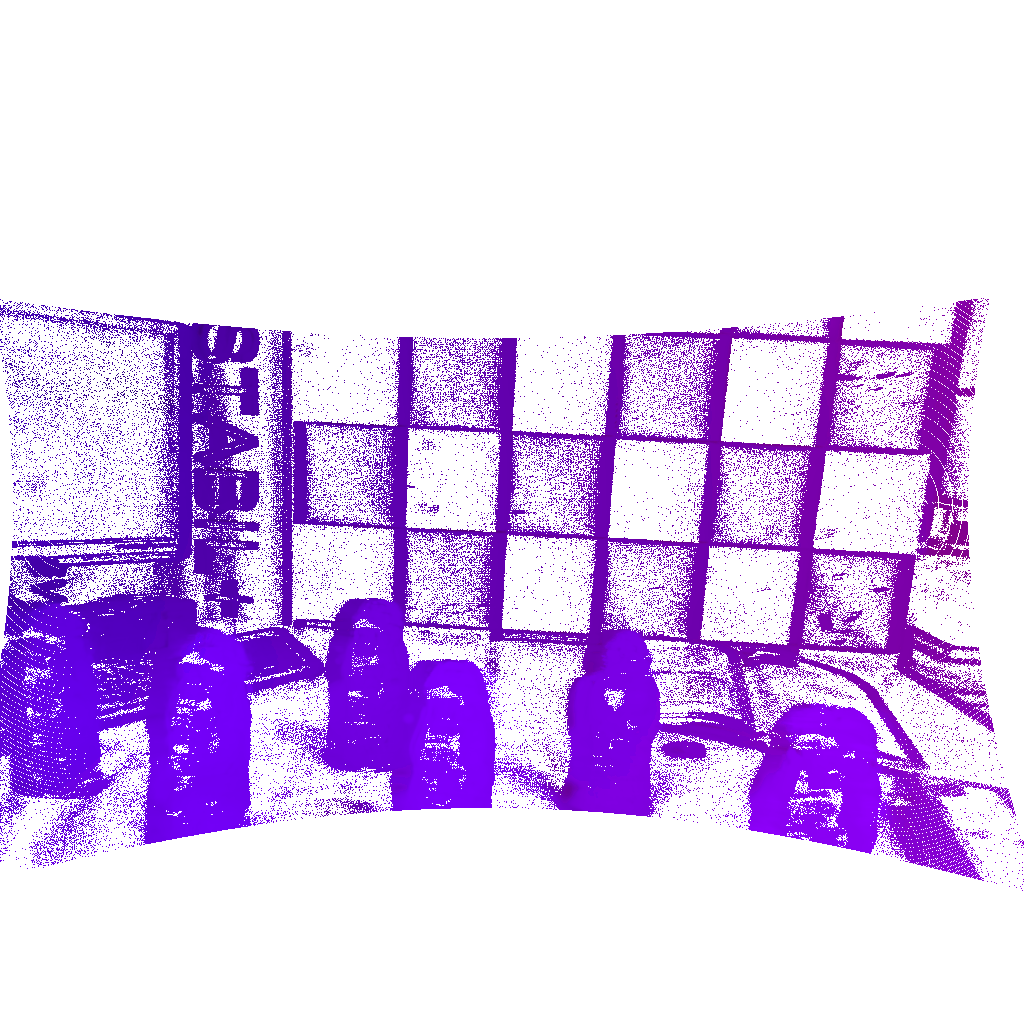}}}
        & {\gframe{\includegraphics[trim={0 0 0 2.5cm},clip,width=\linewidth]{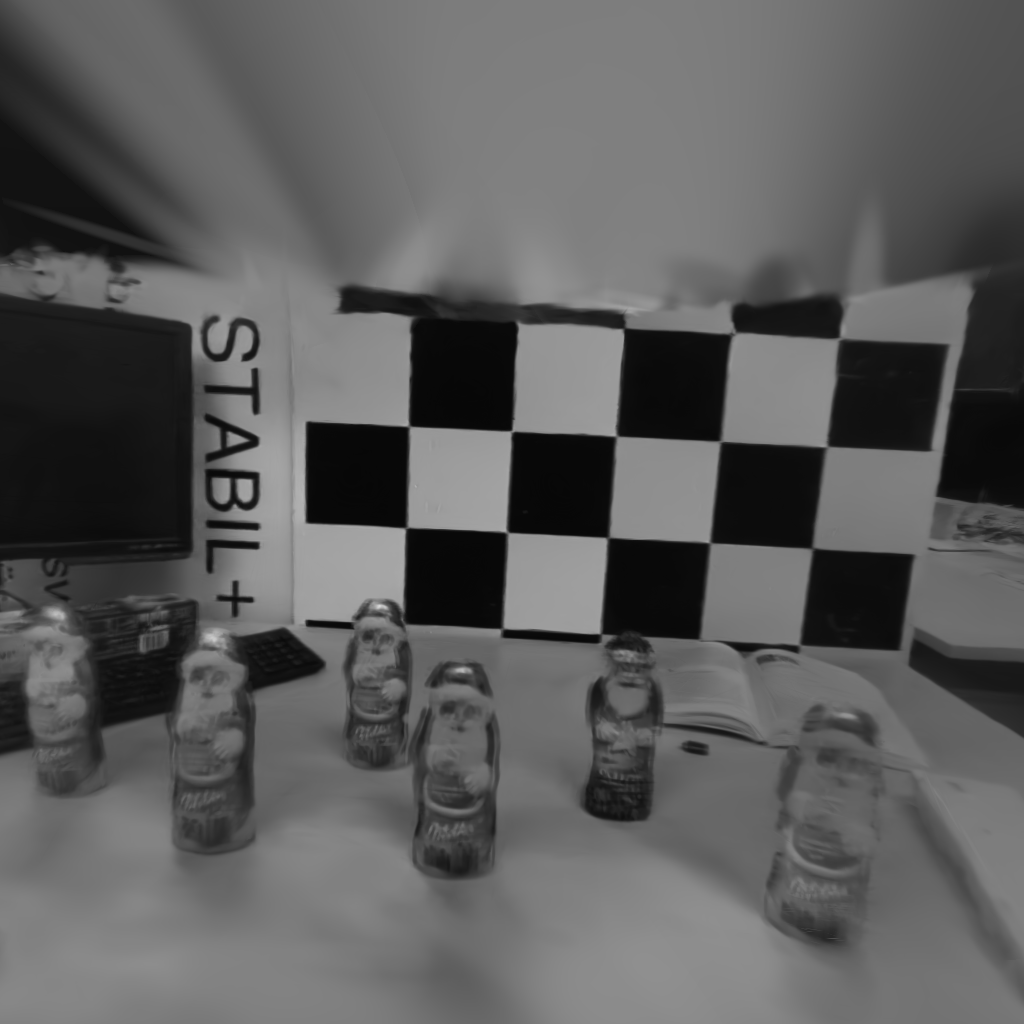}}}
        & {\gframe{\includegraphics[trim={0 0 0 2.5cm},clip,width=\linewidth]{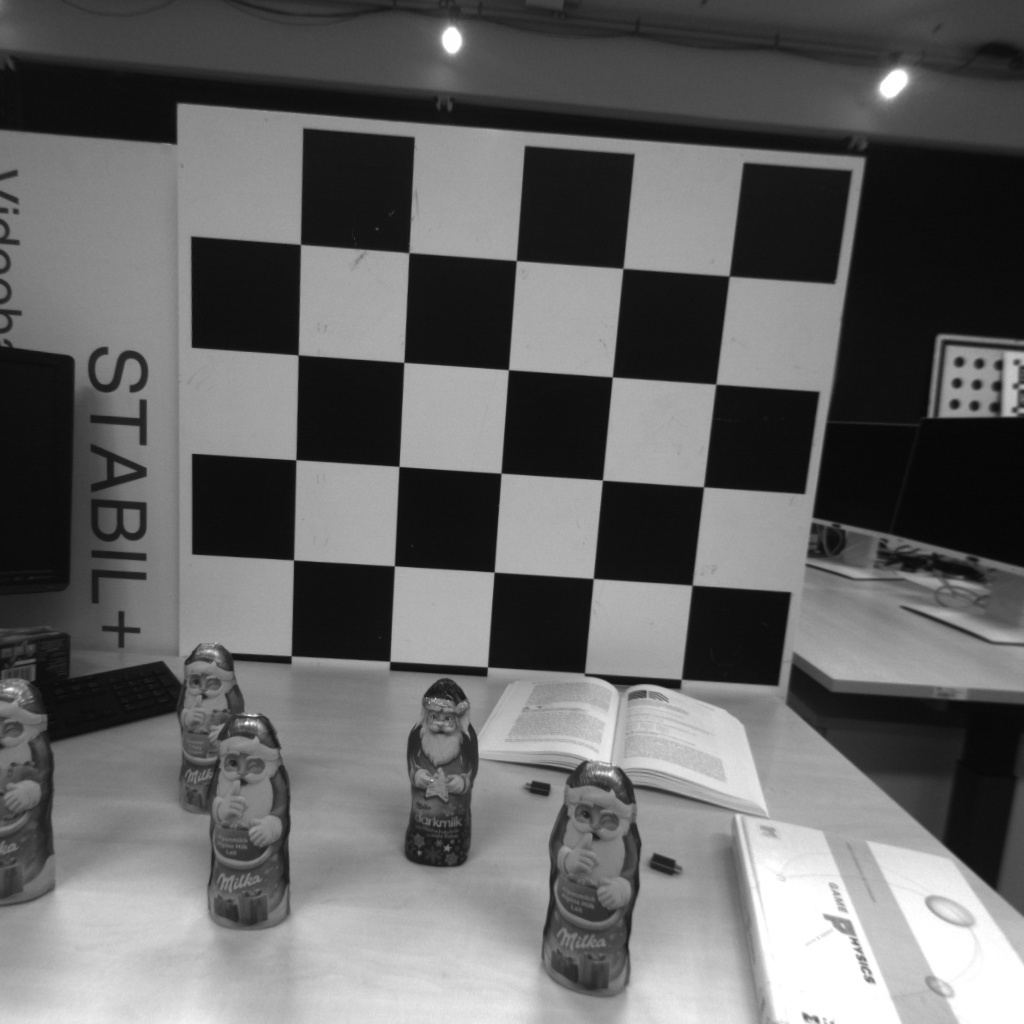}}}\\
   
        {\gframe{\includegraphics[trim={0 0 0 2.5cm},clip,width=\linewidth]{images/depthflow/robust_enerf_chair/chair_dense_depth.png}}}
        & {\gframe{\includegraphics[trim={0 0 0 2.5cm},clip,width=\linewidth]{images/depthflow/robust_enerf_chair/chair_sparse_depth.png}}}
        & {\gframe{\includegraphics[trim={0 0 0 2.5cm},clip,width=\linewidth]{images/depthflow/robust_enerf_chair/chair_dense_flow.png}}}
        & {\gframe{\includegraphics[trim={0 0 0 2.5cm},clip,width=\linewidth]{images/depthflow/robust_enerf_chair/chair_sparse_flow.png}}}
        & {\gframe{\includegraphics[trim={0 0 0 2.5cm},clip,width=\linewidth]{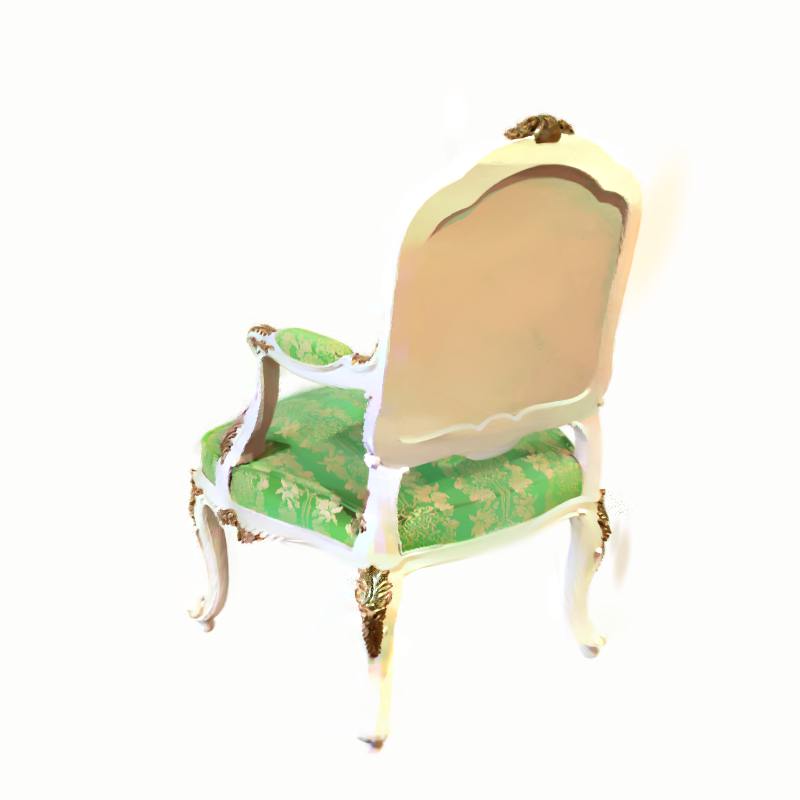}}}
        & {\gframe{\includegraphics[trim={0 0 0 2.5cm},clip,width=\linewidth]{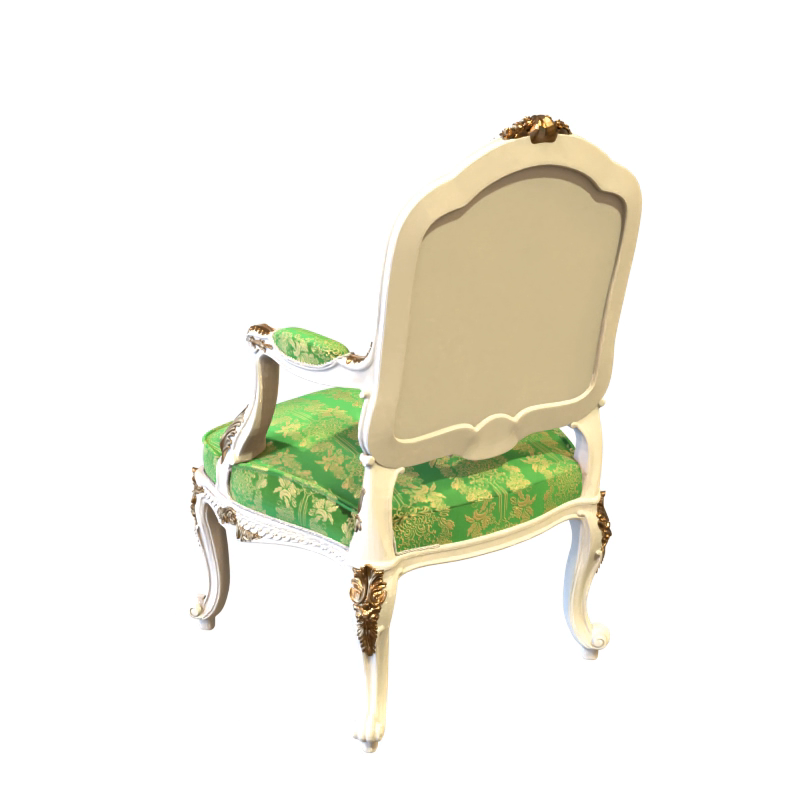}}}\\

        {\gframe{\includegraphics[trim={0 0 0 2.5cm},clip,width=\linewidth]{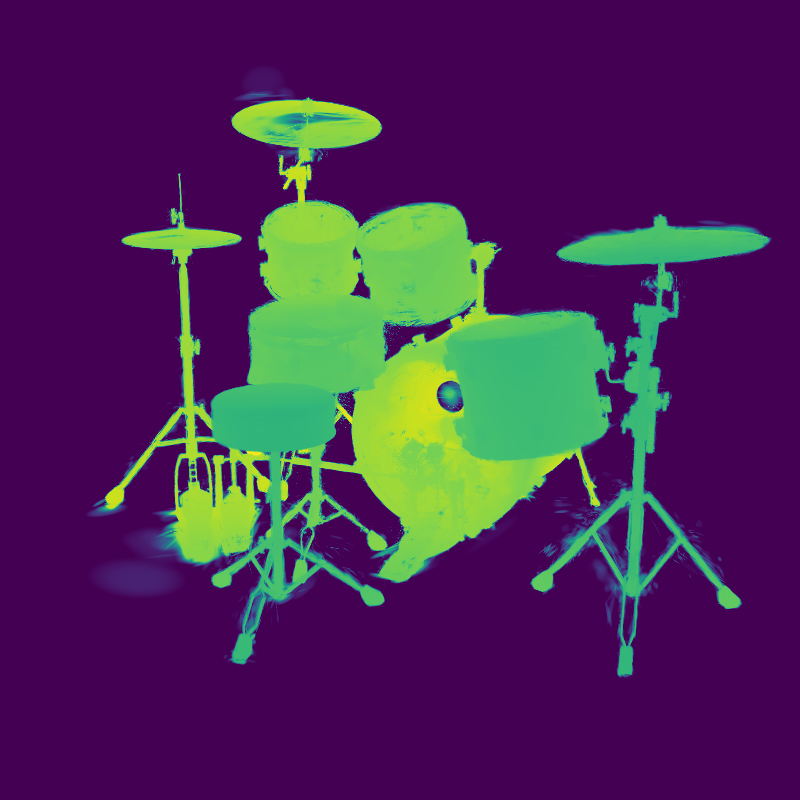}}}
        & {\gframe{\includegraphics[trim={0 0 0 2.5cm},clip,width=\linewidth]{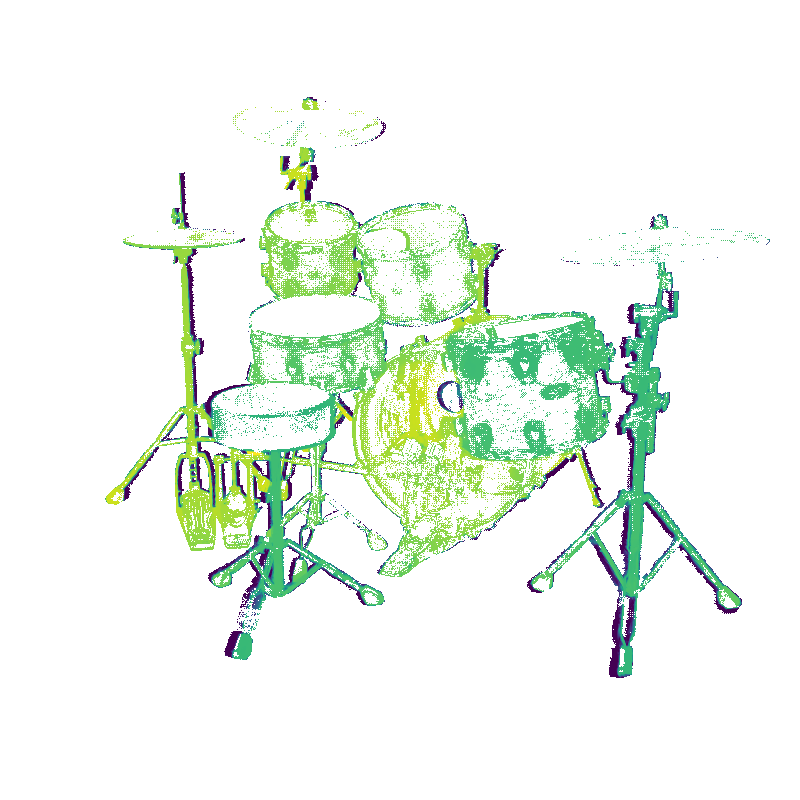}}}
        & {\gframe{\includegraphics[trim={0 0 0 2.5cm},clip,width=\linewidth]{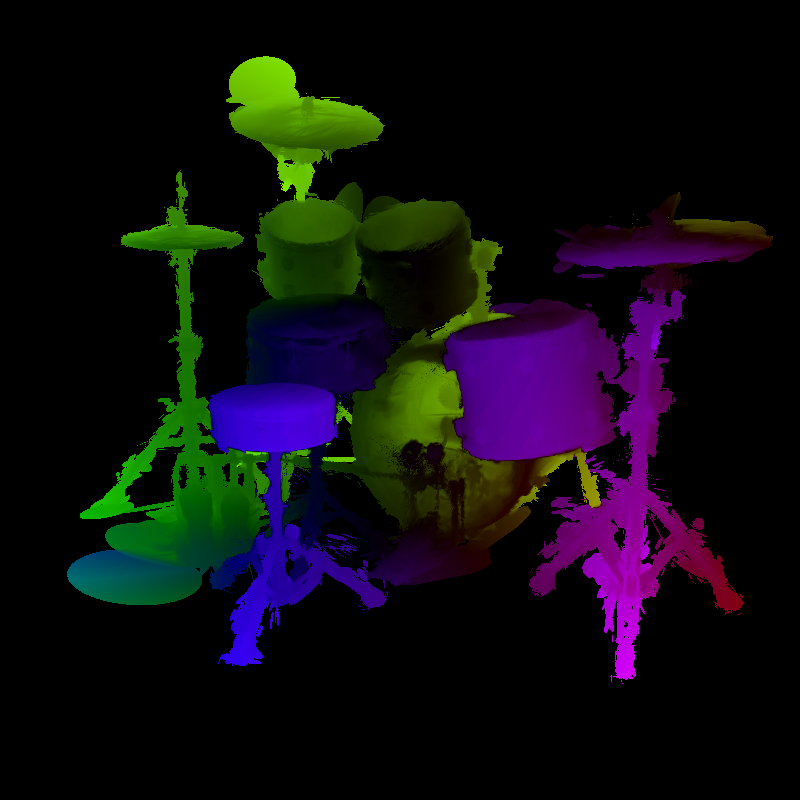}}}
        & {\gframe{\includegraphics[trim={0 0 0 2.5cm},clip,width=\linewidth]{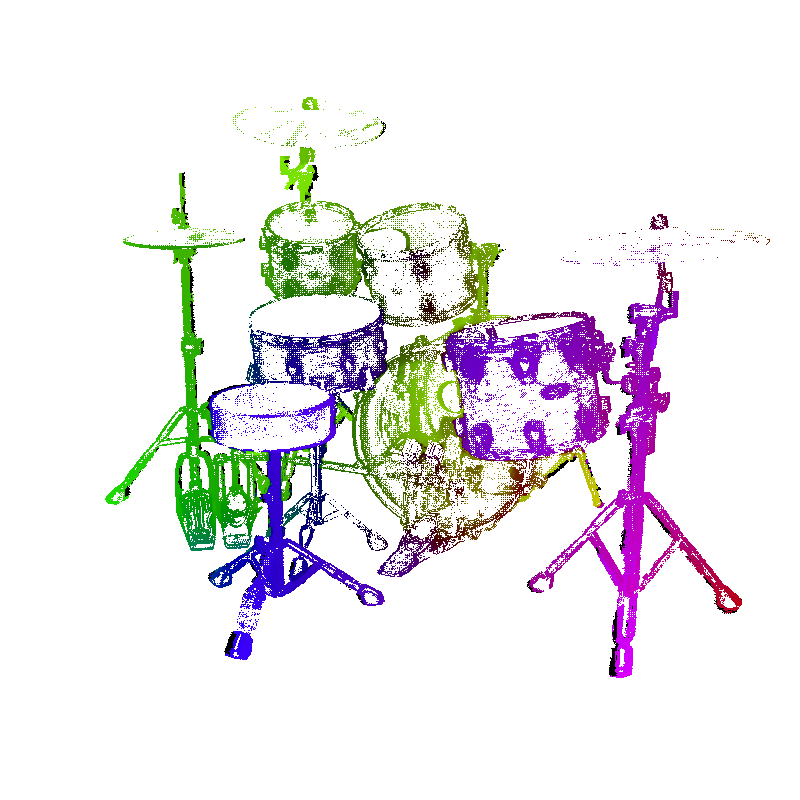}}}
        & {\gframe{\includegraphics[trim={0 0 0 2.5cm},clip,width=\linewidth]{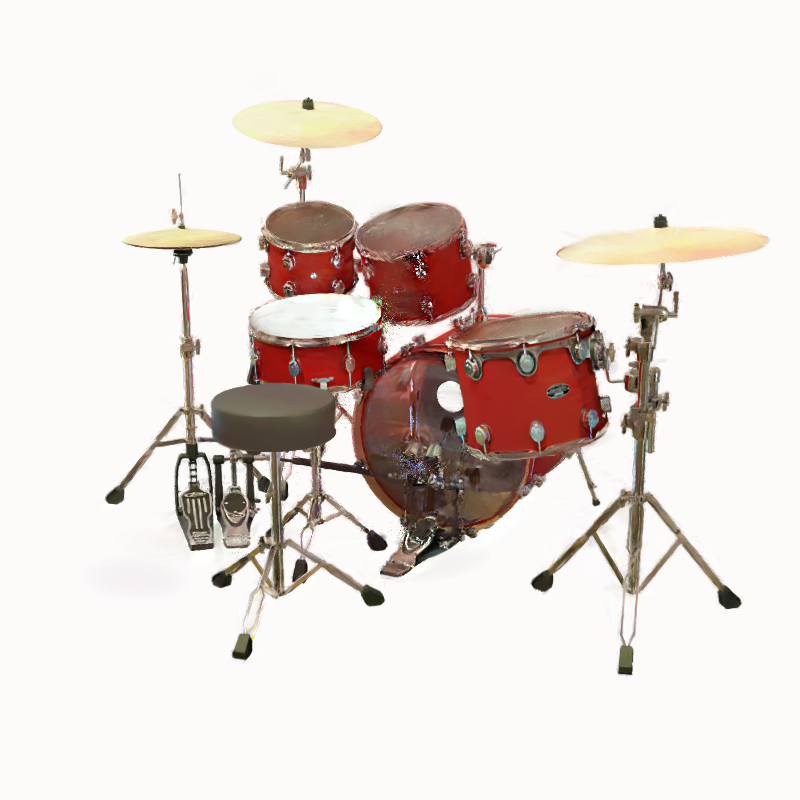}}}
        & {\gframe{\includegraphics[trim={0 0 0 2.5cm},clip,width=\linewidth]{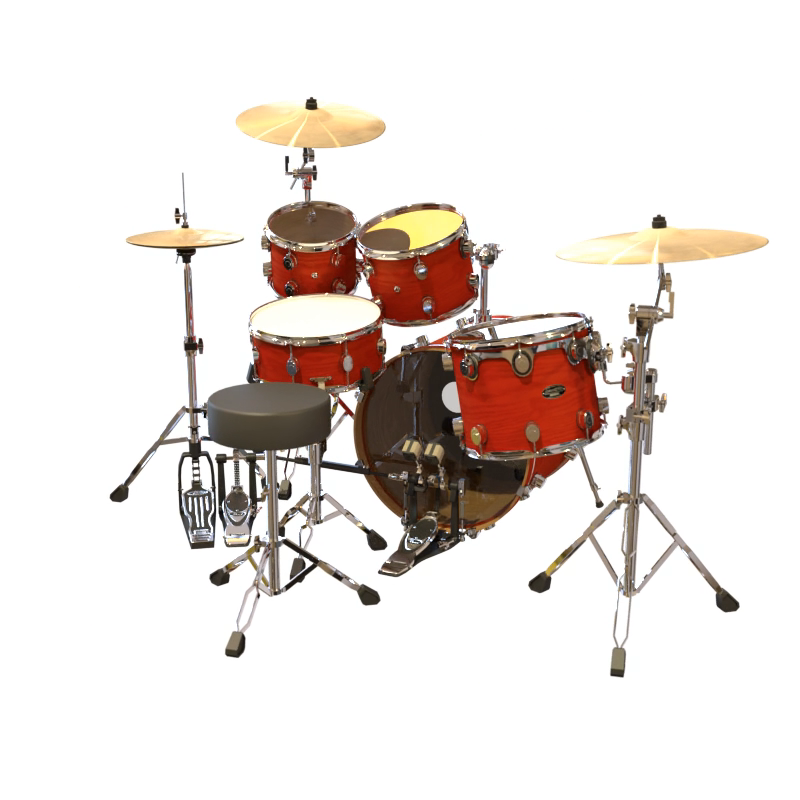}}}\\

        (a) Dense depth
		& (b) Sparse depth
		& (c) Dense flow
		& (d) Sparse flow
		& (e) Rendered intensity
		& (f) GT
		\\
	\end{tabular}
	}
    \caption{Additional depth, flow and intensity reconstruction results on EDS (rows 1 and 2), TUM-VIE (row 3) and color synthetic datasets (rows 4 and 5). 
    }
    \label{fig:suppl:moreRes}
\end{figure*}

\cleardoublepage
{
    \small
    \bibliographystyle{ieeetr_fullname} %
    \bibliography{all}
}

\else 

{
    \small
    \bibliographystyle{ieeetr_fullname} %
    \bibliography{all}
}

\cleardoublepage

\fi

\end{document}